\documentclass[preprint,12pt]{elsarticle}

\usepackage[margin=2.5cm]{geometry}

\usepackage{amssymb}
\usepackage{amsthm}
\usepackage{lineno}
\graphicspath{{plot/}}
\usepackage{subcaption}
\usepackage{tabularx}
\usepackage{array}
\usepackage{amsmath,bm}
\usepackage{multirow}
\usepackage{hyperref}
\usepackage{upgreek}
\DeclareMathOperator*{\argmin
}{argmin} 

\newcommand{\R}{\mathbb{R}}
\begin{document}
\begin{frontmatter}

\title{Use of Multifidelity Training Data and Transfer Learning for Efficient Construction of Subsurface Flow Surrogate Models}
\author[1]{Su Jiang\corref{cor1}}
\author[1]{Louis J.~Durlofsky}

\cortext[cor1]{Corresponding author}

\address[1]{Department of Energy Resources Engineering, Stanford University, Stanford, CA 94305, USA}
            
\begin{abstract}
In subsurface flow settings, data assimilation/history matching presents computational challenges because many high-fidelity models must be simulated. Various deep-learning-based surrogate modeling techniques have been developed to reduce the simulation costs associated with these applications. However, to construct data-driven surrogate models, several thousand high-fidelity simulation runs may be required to provide training samples, and these computations can make training prohibitively expensive. To address this issue, in this work we present a framework where most of the training simulations are performed on coarsened (low-fidelity) geomodels. These models are constructed using a flow-based upscaling method. The framework entails the use of a transfer-learning procedure, incorporated within an existing recurrent residual U-Net architecture, in which network training is accomplished in three steps. In the first step. where the bulk of the training is performed, only low-fidelity simulation results are used. The second and third steps, in which the output layer is trained and the overall network is fine-tuned, require a relatively small number of high-fidelity simulations. Here we use 2500 low-fidelity runs and 200 high-fidelity runs, which leads to about a 90\% reduction in training simulation costs. The method is applied for two-phase subsurface flow in 3D channelized systems, with flow driven by wells. The surrogate model trained with multifidelity data is shown to be nearly as accurate as a reference surrogate trained with only high-fidelity data in predicting dynamic pressure and saturation fields in new geomodels. Importantly, the network provides results that are significantly more accurate than the low-fidelity simulations used for most of the training. The multifidelity surrogate is also applied for history matching using an ensemble-based procedure, where accuracy relative to reference results is again demonstrated.

\end{abstract}




\end{frontmatter}


\section{Introduction}
Accurate subsurface flow modeling is required to manage the production of energy resources and the storage of CO$_2$. The key input in these frameworks is the geological model. Before the geomodel can be used for predictions or optimization, it must be calibrated such that simulation predictions match historical observations. The computations required in this history matching step (also referred to as data assimilation) can be very demanding, however, because thousands to millions of high-fidelity forward simulations must be performed. Surrogate modeling techniques can be very useful in this setting. Data-driven surrogate models represent an effective family of approaches for treating complex geological systems, though existing procedures require several thousand high-fidelity simulations to provide training samples. The computation associated with these training runs is one of the most expensive components of these procedures and may limit their application.

In recent work, a data-driven surrogate model for subsurface flow that employs residual U-Nets within a recurrent neural network (the overall model is referred to as recurrent R-U-Net) was developed and applied for history matching problems in oil production and CO$_2$ storage applications \cite{tang2020deep, tang2021deep, tang2021history, tang2021deep_co2}. In the examples considered, which included up to 128,000 finite volume cells (with two unknowns per cell), 2000-3000 high-fidelity simulations were required as training samples to achieve accurate surrogate model predictions. Our goal in this work is to use high and low fidelity (referred to as multifidelity) simulation data in combination with transfer learning to accelerate surrogate model construction. This strategy enables us to perform $\sim$90\% of the training simulations using low-fidelity models. The low-fidelity geomodels required for these simulations are constructed from high-fidelity geomodels using a flow-based upscaling procedure. A relatively small number of computationally demanding fine-scale simulations are required to fine-tune the surrogate model. 


Many approaches have been developed for the construction of surrogate flow models using deep neural networks. We classify these approaches into two general (though overlapping) categories -- physics-informed neural networks (PINNs) and data-driven deep-learning-based surrogate models. The training process for PINNs is conducted by minimizing the (physical) loss from the residual of the governing partial differential equations along with losses associated with boundary and initial conditions. The minimization process can be accomplished, in theory, without running any forward simulations, though simulation data are also commonly used. \citet{raissi2019physics} first proposed the idea of PINNs to approximate multiple different governing equations. Theory-guided neural networks (TgNNs), which are also physics-informed treatments, were developed by \citet{wang2020deep, wang2021efficient, wang2021theory} and have been used to approximate 2D single-phase flow and other problems. PINN methods incorporating physical loss and simulation data mismatch have been successfully applied by \citet{he2020physics} and \citet{tartakovsky2020physics, tartakovsky2021physics} for single-phase flow and transport problems. Extending these methods to realistic 3D multiphase flow problems, or to multiphysics settings, will require additional development. 

Data-driven deep-learning-based surrogate models are trained with labeled data. After training, these models provide a mapping from simulation input to output. These procedures may require a substantial amount of training data, though they have been applied in complicated flow settings. The spatial correlations are generally captured through use of convolutional neural networks (CNNs), and time evolution is modeled using approaches for the treatment of time series. \citet{zhu2018bayesian} introduced a convolutional encoder-decoder framework to predict single-phase flow solutions. \citet{mo2019deep, mo2019deep_auto, mo2020integration} further developed the framework through application of autoregressive strategies to predict temporal evolution and simulation output over long time frames. Applications of this framework include 2D CO$_2$ storage and 2D and 3D contaminant transport in channelized systems. 

\citet{tang2020deep} developed a recurrent residual U-Net (R-U-Net) model for 2D oil-water problems with flow driven by wells. A residual U-Net and a convolutional long short-term memory (convLSTM) recurrent network were integrated to capture pressure and saturation fields at a number of discrete time steps. \citet{tang2021deep, tang2021history} extended the method to treat 3D oil-water systems with channelized geomodels. This procedure was later modified to model 3D CO$_2$ storage problems involving coupled flow and geomechanics (multiphysics) problems~\cite{tang2021deep_co2}. \citet{wen2021towards} introduced a residual U-Net surrogate model for predicting plume migration in CO$_2$ storage settings. This model, which was developed to be predictive over a wide range of flow and engineering parameters, was trained using 20,000 simulation runs. Transfer learning, with a few hundred additional training samples, was applied to fine-tune the model when new injection settings were considered. 

From the above discussion we see that, with sufficient training, data-driven surrogate models can predict complicated subsurface flow behavior, but the large number of high-fidelity simulation runs required for training may limit their overall applicability. The use of multifidelity data (from high-fidelity and low-fidelity simulation output) for surrogate model construction can mitigate this problem. \citet{geneva2020multi} developed a conditional invertible neural network to generate the probability distribution for high-fidelity turbulent flows conditioned to low-fidelity data. 
Physically accurate turbulent flows and statistics were constructed with a generative method trained with a reduced data set. \citet{meng2020composite} utilized multifidelity data to train a composite physics-informed neural network for unsaturated flow and reactive transport. Three networks were coupled to capture the linear and nonlinear correlations between the high and low-fidelity data. High accuracy was achieved through use of a small set of high-fidelity training data. 

Multifidelity data can be combined with transfer learning to construct surrogate models that provide high-fidelity predictions. This entails first training the surrogate model with low-fidelity simulation results and then using transfer learning to fine-tune the model with a small number of high-fidelity samples. \citet{de2020transfer} applied transfer learning and developed a bi-fidelity-weighted learning for uncertainty propagation in engineering problems. For the process of bi-fidelity-weighted learning, a Gaussian process model was trained with a few high-fidelity samples. This provided synthetic data samples that were used to update the surrogate model trained with low-fidelity data. The use of a Gaussian process model may affect prediction accuracy for more complicated processes. \citet{song2021transfer} applied transfer learning to build a CNN-based surrogate model to solve two-phase flow problems in 2D Gaussian geomodels. The high-fidelity models had four times as many grid blocks as the low-fidelity models. The optimal ratio of low-fidelity to high-fidelity training runs was found to be around five.

In this work, we develop a transfer-learning-based surrogate model that builds on the existing recurrent R-U-Net framework. The target application here is two-phase (well-driven) subsurface flow in 3D channelized systems. A global single-phase transmissibility upscaling procedure is applied to generate coarse-scale models from high-fidelity geomodels. Two-phase flow simulations performed on these models provide the low-fidelity training samples. In our approach, which differs from previous treatments, the high-fidelity geomodel is always used as network input, and the decoder output is also always at high-fidelity. In the first training step, 2500 low-fidelity simulation results are used, and the high-fidelity decoder output is fed to a separate low-fidelity output layer. In subsequent training steps, a small number ($\sim$200) of high-fidelity simulation results are used to train a high-fidelity output layer, which enables the overall network to provide high-fidelity pressure and saturation fields at a specified set of time steps. The surrogate model is then applied for history matching with an ESMDA (ensemble smoother with multiple data assimilation) procedure to generate posterior geomodels. The surrogate model predictions are compared to reference fine-scale simulations to assess the performance of the overall procedure.

This paper proceeds as follows. In Section~\ref{sec:method}, we present the governing equations for the two-phase flow problem and describe the upscaling procedure, which enables the generation of multifidelity data. Then, in Section~\ref{sec:network}, we discuss the recurrent R-U-Net surrogate model and the use of transfer learning with multifidelity data for training. Surrogate model results for oil-water flow in 3D channelized systems are presented in Section~\ref{sec:surr_result}. Comparisons between a surrogate model trained with multifidelity data, a surrogate model trained with only high-fidelity data, and reference numerical simulation results are provided. In Section~\ref{sec:post_surr}, the transfer-learning-based recurrent R-U-Net is combined with ESMDA to generate posterior models and predictions. Conclusions and suggestions for future work in this area appear in Section~\ref{sec:conclusion}.
\section{Governing Equations and Upscaling Procedure}
\label{sec:method}

In this section, we present the governing equations for two-phase subsurface flow problems and describe the upscaling procedure used to provide low-fidelity geomodels. 

\subsection{Governing Equations}\label{sec:governing}
The two-phase immiscible subsurface flow equations considered in this work are applicable for oil reservoir simulation (e.g., oil production via water injection) and to environmental remediation modeling involving water and a nonaqueous-phase liquid (NAPL) contaminant. The governing equations are  
\begin{equation}\label{eq:governing}
    \nabla \cdot (\rho_j \textbf{u}_j) + q_j + \frac{\partial}{\partial t}(\phi \rho_j S_j) = 0, \ j = o, w,
\end{equation}
where $j$ denotes phase, with $o$ indicating oil phase and $w$ water phase, $\rho_j$ is the density of phase $j$, $\textbf{u}_j$ is the Darcy velocity of phase $j$, $q_j$ denotes the source/sink term, $\phi$ is porosity and $S_j$ is the phase saturation (volume fraction). Darcy velocity $\textbf{u}_j$ is given by
\begin{equation}\label{eq:velocity}
    \textbf{u}_j = - \frac{\textbf{k} k_{rj}(S_j)}{\mu_j(p_j)}(\nabla p_j - \rho_j g \nabla z), \ j = o, w,
\end{equation}
where $\textbf{k}$ represents the absolute permeability tensor, $k_{rj}$ is the relative permeability of phase $j$, $\mu_j$ is the viscosity of phase $j$, $p_j$ is the pressure of phase $j$, $g$ is gravitational acceleration, and $z$ is depth. The relative permeability $k_{rj}$ and viscosity $\mu_j$ are nonlinear functions of saturation and pressure, respectively. As is common in reservoir-scale simulations, we neglect capillary pressure in this work, so we have $p = p_w = p_o$.

In the context of oil reservoir simulation, which is the setting considered here, the governing equations \ref{eq:governing} and \ref{eq:velocity} are usually discretized using finite volume methods. Newton’s method is applied to solve the discrete nonlinear equations. In this work, we use Stanford's Automatic Differentiation General Purpose Research Simulator ADGPRS \cite{zhou2012parallel} for both high-fidelity (HF) and low-fidelity (LF) simulation.

High-fidelity geomodels are denoted by $\mathbf{m}^h \in \R^{n^h \times 1}$, where $n^{h} = n_x^h \times n_y^h \times n_z^h$ is the number of grid blocks in the HF model, with $n_x^h$, $n_y^h$ and $n_z^h$ the number of blocks in each coordinate direction (superscripts $h$ and $l$ indicate HF and LF). Note that here the geological parameters are represented in terms of a single quantity in each block, which defines (isotropic) permeability and porosity. The governing equations are solved, with $\mathbf{m}^h$ as input, to provide HF solutions for the state variables. In oil-water problems these are pressure $\mathbf{P}^h \in \R^{n^h \times n_{ts}}$ and saturation $\mathbf{S}^h \in \R^{n^h \times n_{ts}}$ in every grid block at $n_{ts}$ simulation time steps. The simulation process can be represented as 
\begin{equation}
    \mathbf{x}^h = [\mathbf{P}^h, \mathbf{S}^h] = g(\mathbf{m}^h), 
\end{equation}
where $g$ represents the forward numerical simulation process and $\mathbf{x}^h$ is the simulation output. 

Well injection/production rates are primary quantities of interest in subsurface flow problems. These rates are computed through application of  
\begin{equation} \label{eq:well}
    (q_j^w)_i = WI_i\left(\frac{k_{rj}\rho_j}{\mu_j}\right)_i(p_i - p_i^w),
\end{equation}
where $(q_j^w)_i$ denotes the source/sink mass flow rate for phase $j$ in well block $i$, $p_i$ denotes the well-block pressure, $p_i^w$ denotes the wellbore pressure evaluated at the center of well block $i$, and $WI_i$ is the well index, computed as \cite{peaceman1983interpretation}
\begin{equation} \label{eq:wi}
    WI_i = \frac{2 \pi k_i \Delta z }{\ln{\frac{r_0}{r_w}}}, 
\end{equation}
where $k_i$ is the well block permeability, $\Delta z$ is the thickness of the well block, $r_0 = 0.2 \Delta x$ for isotropic permeability and $\Delta x = \Delta y$, where $\Delta x$ and $\Delta y$ denote the dimensions of the well block. Eq.~\ref{eq:wi} is for a vertical well that fully penetrates the grid block. The wellbore pressure $p_i^w$, which varies with depth, is given by 
\begin{equation}
    p_{i+1}^w = p_i^w + (\rho_{i, i + 1} g) \Delta z_{i, i + 1}.
\end{equation}
Here $\rho_{i, i + 1}$ is the average fluid density between well blocks $i$ and $i + 1$ and $\Delta z_{i, i + 1}$ denotes the difference in depth. Bottom-hole pressure (BHP), which will be specified in our simulations, corresponds to the wellbore pressure at the uppermost perforation.

\subsection{Upscaling Procedure}
\label{sec:method_upscaling}
A flow-based, single-phase global transmissibility upscaling procedure \cite{zhang2008new, chen2008nonlinear, crain2020multifidelity} is applied to generate coarse-scale (LF) models from fine-scale (HF) geomodels. The specific implementation used in this work is that of \citet{crain2020multifidelity}. This discussion follows \cite{jiang2021treatment}, where upscaled models of the type used here were applied with error models for history matching in data space.

The first step in the upscaling process is the solution of the global single-phase steady-state pressure equation over the HF geomodel, i.e., 
\begin{equation}\label{eq:pressure_single_phase}
    \nabla \cdot (k \nabla p) = q,
\end{equation}
where the source term $q$, which corresponds to flow driven by wells, is expressed using the single-phase-flow analog of Eq.~\ref{eq:well}. From the HF pressure solution, the block-to-block volumetric flow rate $f_{\alpha}^h$ is computed via
\begin{equation} \label{eq:HF_flux}
    f_{\alpha}^h = T_{\alpha}(p_{\alpha^+}^h - p_{\alpha^-}^h),
\end{equation}
where $T_{\alpha}$ is the transmissibility, $\alpha$ denotes the interface of two fine-scale grid cells, and $p_{\alpha^{\pm}}^h$ denotes the pressures of the grid blocks on either side of interface $\alpha$. Transmissibility, which is essentially the numerical analog of permeability, is given by $T_{\alpha} = k_{\alpha} \Delta y \Delta z/\Delta x$, for cells connected in the $x$-direction, where $k_{\alpha}$ is the harmonic average of the permeabilities of the blocks on either side of interface $\alpha$. Analogous expressions define transmissibilities for cells connected in the $y$ and $z$-directions.

Given the HF pressure solution and flow rates through all HF block-to-block interfaces (Eq.~\ref{eq:HF_flux}), the coarse-scale (LF) transmissibilities and well indices can be constructed. Let $\beta$ denote the interface between two LF grid blocks. The LF block pressures $p_{\beta^{\pm}}^l$ for block $\beta^+$ and $\beta^-$ are estimated as the volume average ($\langle p \rangle_{\beta^{\pm}}$) of the fine-scale pressures $p^h$ that lie within the range of coarse blocks $\beta^+$ and $\beta^-$. The flow rate $f_{\beta}^l$ through LF interface $\beta$ is estimated to be the sum of the fine-scale flow rates $f_{\alpha}^h$ over the range of interface $\beta$. In analogy to Eq.~\ref{eq:HF_flux}, the upscaled transmissibility $T_{\beta}^*$, over the LF interface $\beta$, is then given by
\begin{equation} \label{eq:coarse_T}
    T_{\beta}^* = \frac{f_{\beta}^l}{p_{\beta^{-}}^l - p_{\beta^{+}}^l} \approx \frac{\sum_{\alpha \in \beta} f_{\alpha}^h}{\langle p \rangle_{\beta^{-}} - \langle p \rangle_{\beta^{+}}}.
\end{equation}
The above estimate of LF transmissibility, which can be viewed as a volume average of the HF flow result, has been shown to provide accurate LF transmissibilities. It can be improved by iterating on the LF model to force very close agreement between $\sum_{\alpha \in \beta} f_{\alpha}^h$ and $f_{\beta}^l$ \cite{chen2008nonlinear}. Treatments for handling anomalous $T_{\beta}^*$, which can occur when $f_{\beta}^l$ and/or $(p_{\beta^{-}}^l - p_{\beta^{+}}^l)$ are very small, are discussed by \citet{crain2020multifidelity}.

The upscaled well index $WI^*$ provides a transmissibility relating well injection/production rate to the difference in pressure between the wellbore and the well block. In analogy to Eq.~\ref{eq:coarse_T}, this is given by 
\begin{equation}
    WI_i^* = \frac{f^{w, l}_i}{p_i^l - p^w_i} \approx \frac{\sum_{k \in i} f_k^{w, h}}{\langle p \rangle_i - p^w_i},
\end{equation}
where $f^{w, l}_i$ denotes the flow rate between the well and LF well block $i$. This is simply the sum of the HF flow rates $f_k^{w, h}$ in the range of coarse well-block $i$. The LF block pressure $p_i^l$ is again the volume average of the corresponding HF block pressures. 

The upscaling procedure is applied to generate LF geomodels $\mathbf{m}^{l} \in \R^{n^l \times 4}$ from high-fidelity models $\mathbf{m}^h \in \R^{n^h \times 1}$, where $n^l = n_x^l \times n_y^l \times n_z^l$ is the number of LF grid blocks. Here $n^l \times 4$ appears in the LF model dimension because this geomodel is characterized, in general, by LF porosity and different transmissibilities in the $x$, $y$ and $z$-directions (even when the HF permeability field is isotropic). The two-phase flow simulation of the LF model can be expressed as 
\begin{equation}
    \mathbf{x}^l = [\mathbf{P}^l, \mathbf{S}^l] = g(\mathbf{m}^l),
\end{equation}
where $\mathbf{P}^l\in \R^{n^l \times n_{ts}}$, $\mathbf{S}^l \in \R^{n^l \times n_{ts}}$ denote the pressure and saturation from LF simulation and $\mathbf{x}^l$ represents LF simulation output. It is important to note that, because the LF model is constructed to capture single-phase pressures and flow rates, it is only approximate in two-phase-flow settings (i.e., for the solution of Eqs.~\ref{eq:governing} and \ref{eq:velocity}). In particular, the LF model used here does not include any upscaling of relative permeability effects, which can be important in coarse-scale models for two-phase flow. Thus we expect this representation to lead to some error relative to HF simulation results.

\section{Network Training with Transfer Learning and Multifidelity Data}
\label{sec:network}

In this section we first provide a brief description of the recurrent R-U-Net surrogate model~\cite{tang2020deep, tang2021deep}. The use of transfer learning with multifidelity data to train the surrogate model is then described.  

\subsection{3D Recurrent Residual U-Net} \label{sec:3D_RRUN}

With the existing deep-neural-network surrogate \citep{tang2021deep}, the forward model can be expressed as 
\begin{equation}
    \hat{\mathbf{x}}^h =[\hat{\mathbf{P}}^h, \hat{\mathbf{S}}^h] =  \tilde{g}(\mathbf{m}^h, \bm{\uptheta}^h), 
\end{equation}
where $\hat{\mathbf{x}}^h$ represents the HF surrogate model output, $\hat{\mathbf{P}}^h \in \R^{n^h \times n_{t}}$ and $\hat{\mathbf{S}}^h \in \R^{n^h \times n_{t}}$ are the surrogate model pressure and saturation output, $\tilde{g}$ denotes the surrogate model, and $\bm{\uptheta}$ indicates the tunable network parameters. Here $n_{t}$ is the number of time steps at which surrogate model output is constructed. This is generally somewhat less than the number of simulation time steps $n_{ts}$, e.g., in this work $n_{ts} \sim 30$--35, and $n_{t}=10$.

\begin{figure}[!htb]
\centering
\includegraphics[width = 0.95\textwidth]{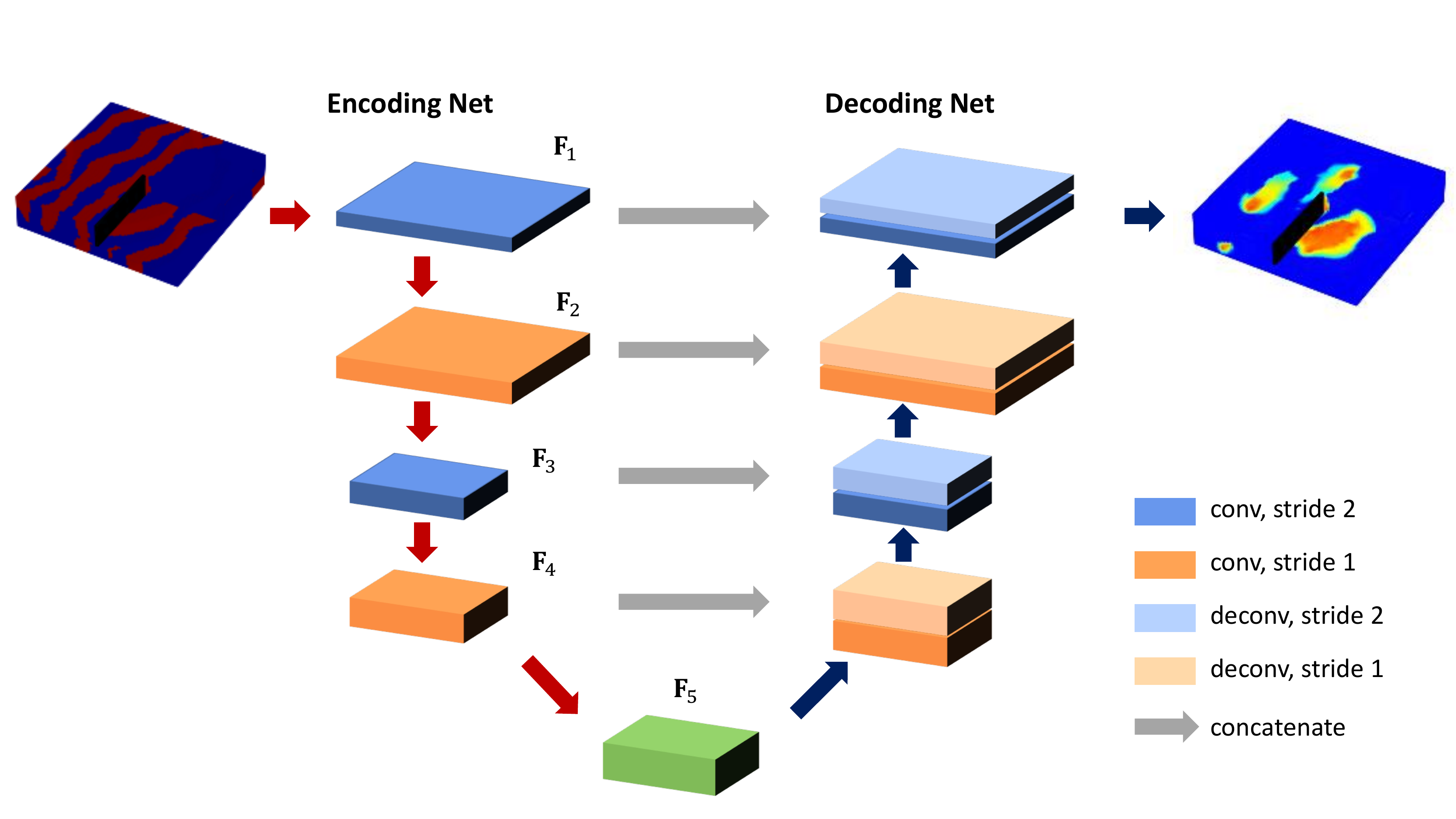}
\caption{3D residual U-Net architecture.}
\label{fig:r_u_net}
\end{figure}

The 3D recurrent R-U-Net architecture is used as the framework in this study. The residual U-Net, illustrated in Fig.~\ref{fig:r_u_net}, includes encoding and decoding networks to capture the spatial correlation in simulation input and output. The encoding network maps the input geomodel $\mathbf{m}^h$ to low-dimensional latent features $\mathbf{F}_1, \ldots, \mathbf{F}_5$. These feature maps are concatenated with the upsampled features in the decoding net, which facilitates the representation of multiscale effects in the solution. 

\begin{figure}[!htb]
\centering
\includegraphics[width = 0.95\textwidth]{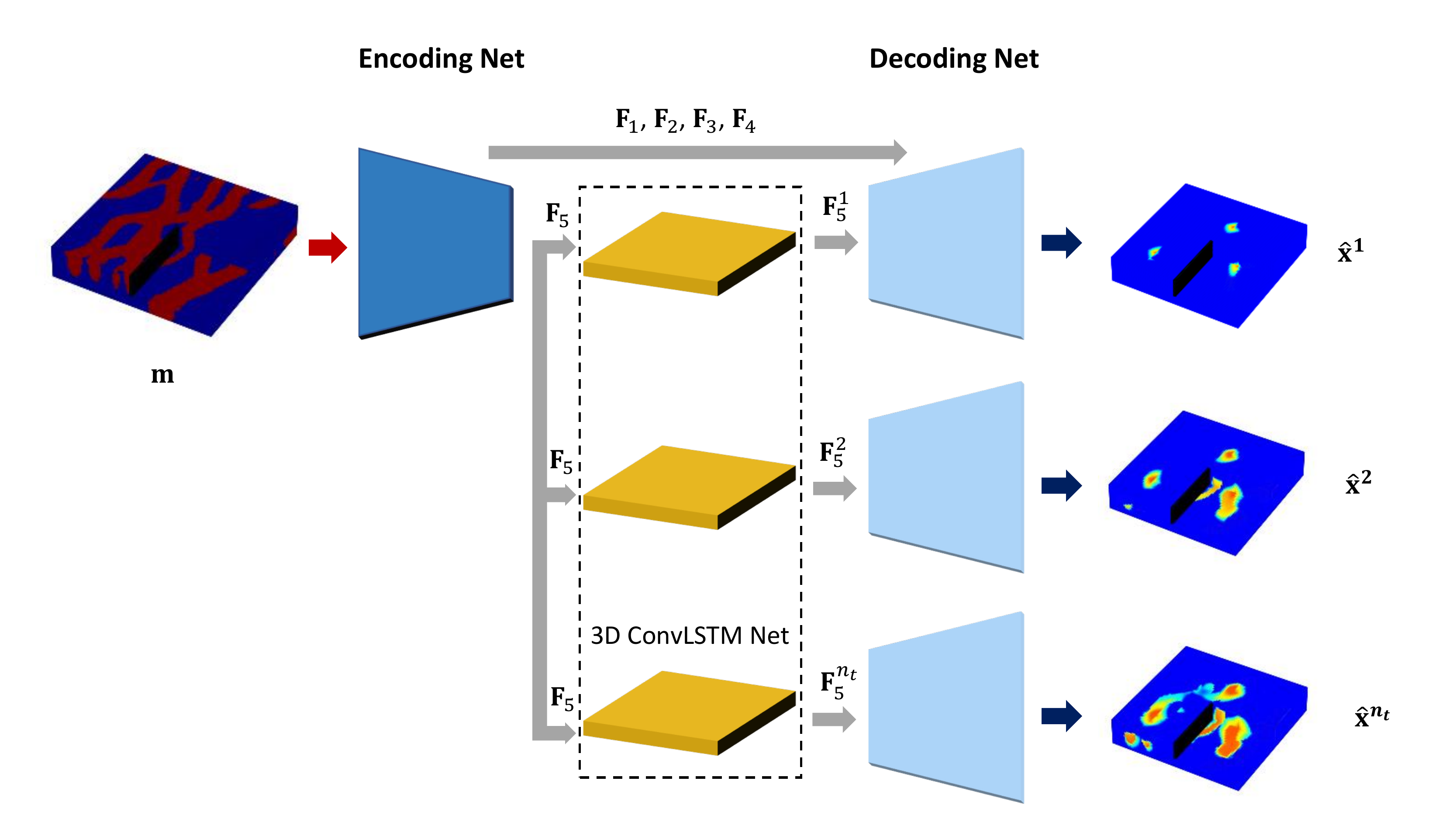}
\caption{Recurrent R-U-Net architecture~\cite{tang2021deep}.}
\label{fig:recurrent_r_u_net}
\end{figure}

The recurrent network is introduced to capture the temporal evolution of the simulation output. Figure~\ref{fig:recurrent_r_u_net} displays the recurrent component of the recurrent R-U-Net. The overall surrogate model includes the residual U-Net and a long short-term memory convolutional (convLSTM) recurrent network. The most compressed (and most global) feature $\mathbf{F}_5$ is mapped to a time series of latent features, $\mathbf{F}_5^{1}, \mathbf{F}_5^{2}, \ldots, \mathbf{F}_5^{n_t}$, by the convLSTM module. The feature $\mathbf{F}_5^{t}$ from the convLSTM (for $t=1, \dots, n_t$), along with extracted features $\mathbf{F}_1, \ldots, \mathbf{F}_4$ from the encoding net, are combined as they move through the decoding net. The decoding net then generates a prediction $\hat{\mathbf{x}}^{h, t}$ for each of the $n_t$ time steps. Please see~\cite{tang2020deep, tang2021deep} for more details on the recurrent R-U-Net architecture. 

The recurrent R-U-Net is trained by minimizing the loss between the simulated pressure and saturation fields and the corresponding surrogate model predictions. A set of forward simulations is performed with specified well locations and settings (BHPs in our case). The pressure and saturation data from $n_t$ time steps are collected. The minimization problem is given by
\begin{equation}\label{eq:obj}
\begin{split}
    \bm{\uptheta}^* =  \argmin_{\bm{\uptheta}} & \frac{1}{n_{smp}}\frac{1}{n_t}\sum_{i = 1}^{n_{smp}}\sum_{t = 1}^{n_t}||\hat{\mathbf{x}}_i^{h,t} - \mathbf{x}_i^{h,t}||_2^2  \\
    & + \lambda_w \frac{1}{n_{smp}}\frac{1}{n_t}\frac{1}{n_w} \sum_{i = 1}^{n_{smp}}\sum_{t = 1}^{n_t}\sum_{w = 1}^{n_w} ||\hat{x}_i^{h, t, w} - x_i^{h, t, w}||_2^2,
\end{split}
\end{equation}
where $n_{smp}$ is the number of training samples, $n_w$ is the number of well blocks, and $\lambda_w$ represents the weighting for well-block data. The second term in Eq.~\ref{eq:obj} enables additional weighting at well blocks, which acts to improve the accuracy of pressure and saturation at well locations. This in turn improves the accuracy of well rate predictions (see Eq.~\ref{eq:well}), which are often the key observations used in data assimilation problems. Note that, following \cite{tang2021deep}, separate networks are trained for pressure and saturation predictions.

\subsection{Transfer Learning with Multifidelity Data} 
\label{sec:TLMF}
The training of the existing recurrent R-U-Net requires a large number of high-fidelity simulation runs as training samples. For example, in \citet{tang2021history}, 2500 runs were used to train the network to predict pressure and saturation in 3D channelized models containing 128,000 grid blocks. Each high-fidelity simulation for the 3D models considered in this work (which also contain 128,000 blocks) requires 20 minutes using ADGPRS on a single CPU. Practical 3D models, which may contain millions of grid blocks, may entail hours of computation for a single forward simulation. Thus there is a practical need to reduce the computational demands associated with network training.

The idea of transfer learning is to apply a neural network trained for one task to a different but related task. With the knowledge embodied in the neural network trained for the first task, the training cost for the second task may be substantially reduced. In the approach used here, the model is first trained with 2500 LF (coarse-scale) simulation runs. The coarse-scale flow physics (i.e., simulation operator $g$) is the same as that on the fine scale, and the coarse-scale flow responses approximate (with some error) the HF simulation results. A relatively small number of HF simulations (100--400 are considered here) are then applied to fine-tune the neural networks and thus complete the HF surrogate model. 

\begin{figure}[!htb]
\centering
\begin{minipage}{.67\linewidth}\centering
\includegraphics[trim = 0 00 0 00, clip, width=\linewidth]{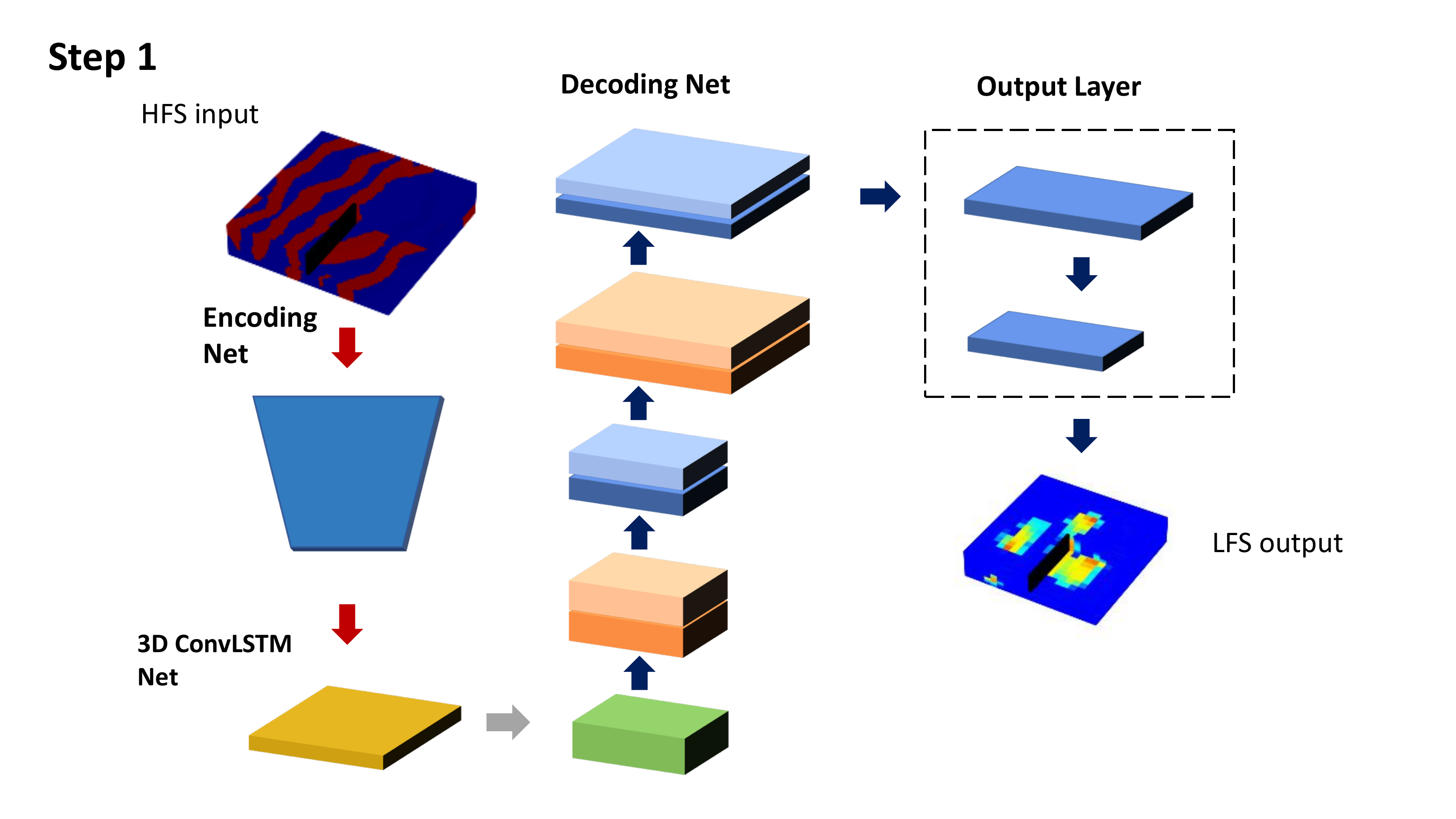}
\subcaption{Step 1: surrogate model training with LF output layer}
\end{minipage}
\begin{minipage}{.67\linewidth}\centering
\includegraphics[trim = 0 00 0 00, clip, width=\linewidth]{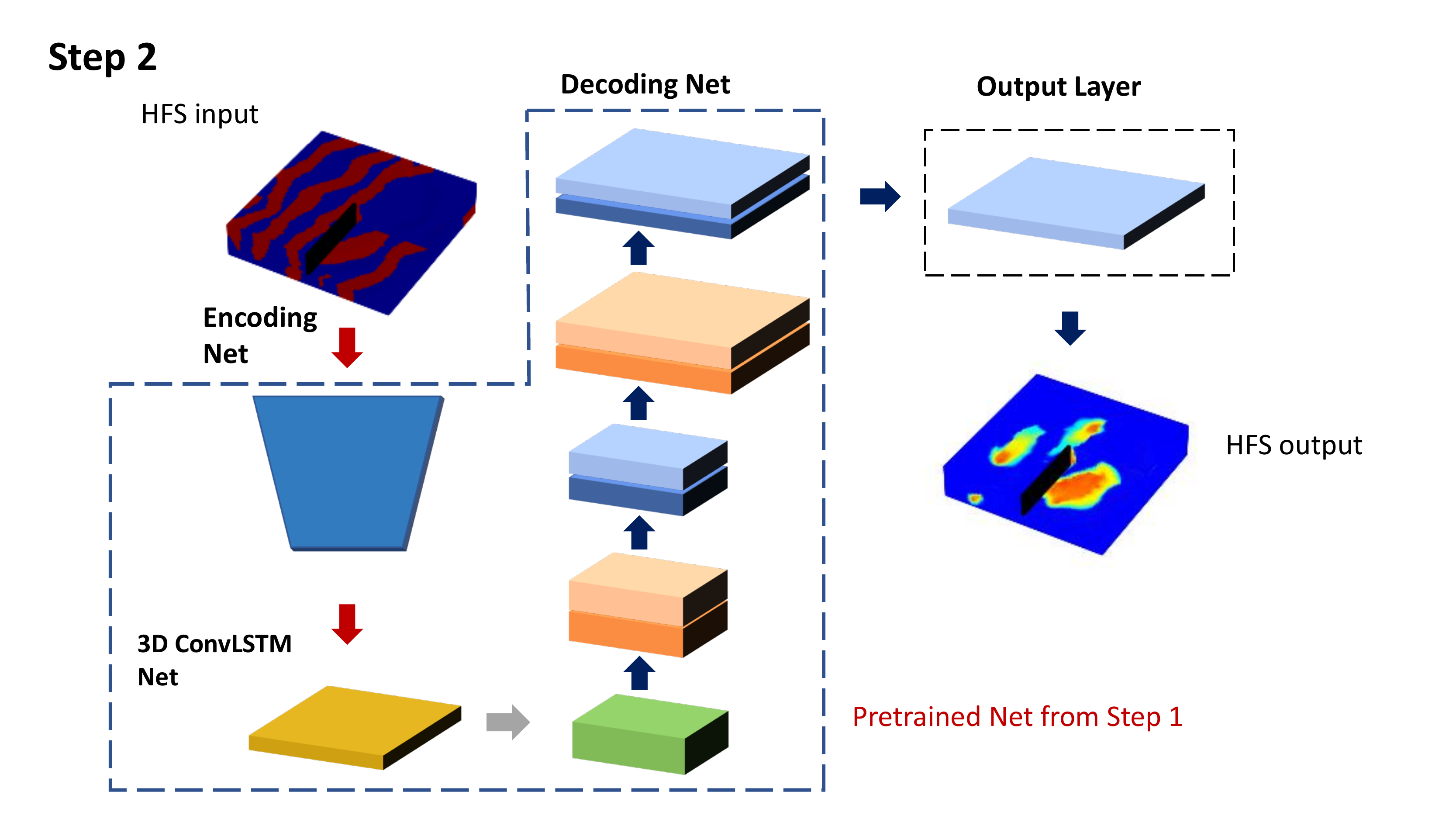}
\subcaption{Step 2: transfer learning with HF output layer}
\end{minipage}
\begin{minipage}{.67\linewidth}\centering
\includegraphics[trim = 0 0 0 0, clip, width=\linewidth]{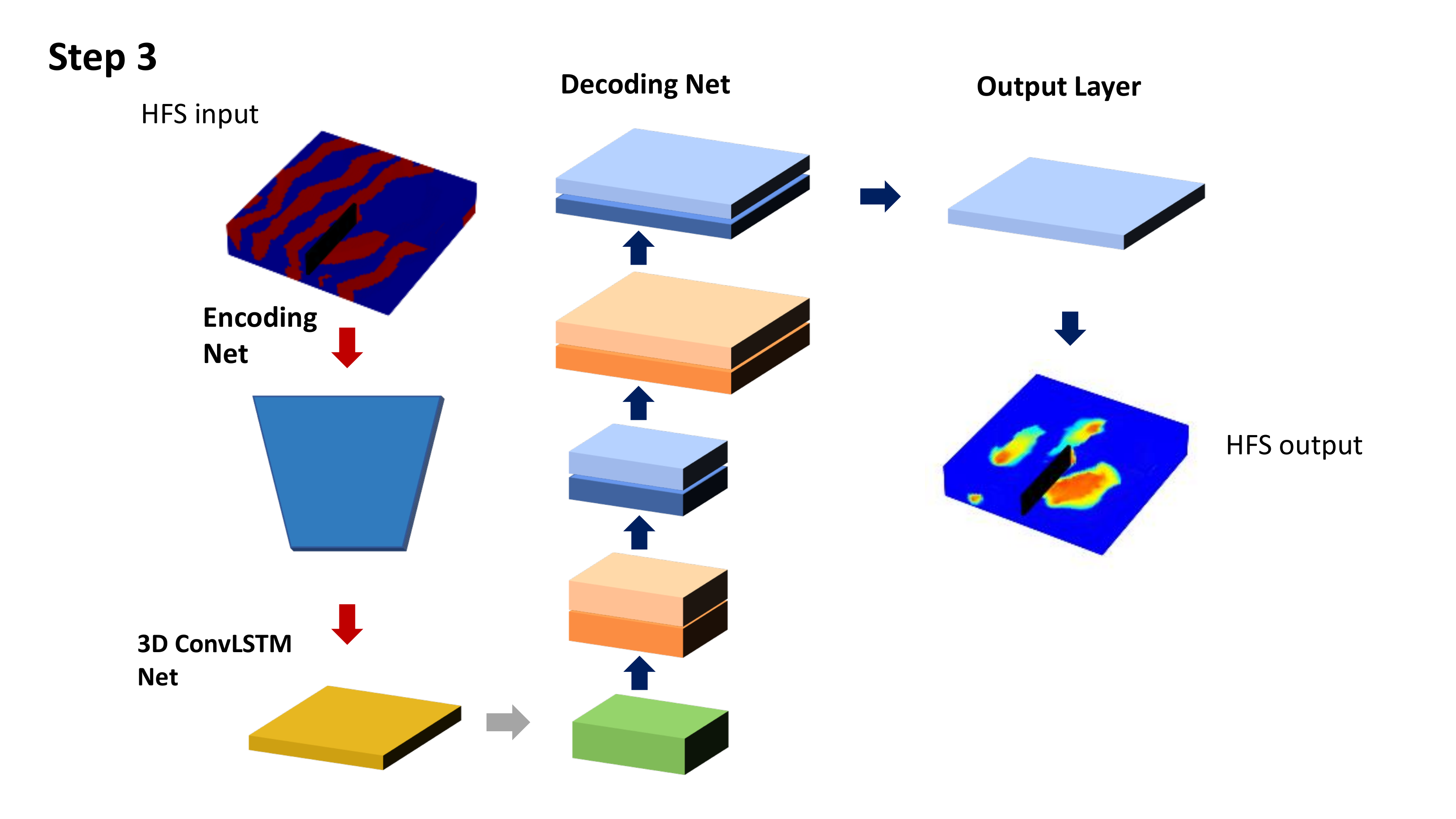}
\subcaption{Step 3: fine-tuning of full network with HF data}
\end{minipage}
\caption{Workflow for surrogate model training with multifidelity data and transfer learning.}
\label{fig:multi-fidelity}
\end{figure}

The transfer learning process involves three steps, which are shown in Fig.~\ref{fig:multi-fidelity}. First, we predict LF model output $\hat{\mathbf{x}}^l$ from the input high-fidelity geomodel $\mathbf{m}^h \in \R^{n^h \times 1}$, 
\begin{equation}
    \hat{\mathbf{x}}^l = [\hat{\mathbf{P}}^l, \hat{\mathbf{S}}^l] = \tilde{g}(\mathbf{m}^h, \bm{\uptheta}^l).
\end{equation}
Here $\bm{\uptheta}^l$ denotes the parameters for the LF neural networks. We train the surrogate model with $n_{smp}^l=2500$ LF samples. The model parameters $\bm{\uptheta}^l = [\bm{\uptheta}_{enc}^l,\ \bm{\uptheta}_{convLSTM}^l,\ \bm{\uptheta}_{dec}^l, \  \bm{\uptheta}_{output}^l]$ include parameters for the encoder ($\bm{\uptheta}_{enc}^l$), 3D convLSTM network ($\bm{\uptheta}_{convLSTM}^l$), decoder ($\bm{\uptheta}_{dec}^l$), and output layers ($\bm{\uptheta}_{output}^l$). Note that the LF geomodel $\mathbf{m}^{l}$ is not input to the network (here or in subsequent steps). Rather, the network always works with $\mathbf{m}^{h}$, even when it is trained using $\mathbf{x}^l$.

In the Step~1 training process, the surrogate model is trained to capture spatial distributions and temporal evolution with parameters $\bm{\uptheta}_{enc}^l, \  \bm{\uptheta}^l_{convLSTM}$, $\bm{\uptheta}^l_{dec}$. These parameters are associated with major portions of the recurrent R-U-Net. Step~1 training also provides $\bm{\uptheta}_{output}^l$. In Step~2, parameters $\bm{\uptheta}_{enc}^l$, $\bm{\uptheta}^l_{convLSTM}$ and $\bm{\uptheta}^l_{dec}$ are not updated, though the output-layer parameters are modified. Specifically, in this step we map the geomodel input $\mathbf{m}^h$ to HF simulation output $\hat{\mathbf{x}}^h\in \R^{n^h \times n_t}$ through 
\begin{equation}
    \hat{\mathbf{x}}^h = [\hat{\mathbf{P}}^h, \hat{\mathbf{S}}^h] =  \tilde{g}(\mathbf{m}^h, \bm{\uptheta}^h).
\end{equation}
The parameter set is now defined as $\bm{\uptheta}^h =[\bm{\uptheta}_{enc}^l, \  \bm{\uptheta}_{convLSTM}^l, \ \bm{\uptheta}_{dec}^l, \  \bm{\uptheta}_{output}^h]$. The output layers with parameters $\bm{\uptheta}_{output}^h$ are trained with $n_{smp}^h$ HF samples. In this work, we evaluate network performance over the range $n_{smp}^h = 100$--400.

The last step of the transfer learning procedure is the fine-tuning of all parameters using the $n_{smp}^h$ HF samples. The parameter set is now denoted by $\bm{\uptheta}^h =[\bm{\uptheta}_{enc}^h$, $\bm{\uptheta}_{convLSTM}^h$, $\bm{\uptheta}_{dec}^h, \  \bm{\uptheta}_{output}^h]$. The parameters $\bm{\uptheta}_{enc}^h$, $\bm{\uptheta}_{convLSTM}^h$ and $\bm{\uptheta}_{dec}^h$ are well-approximated by $\bm{\uptheta}_{enc}^l$, $\bm{\uptheta}_{convLSTM}^l$ and $\bm{\uptheta}_{dec}^l$, and an initial estimate for $\bm{\uptheta}_{output}^h$ is already available, so this training step is relatively fast.

Table~\ref{tab:recurrent_runet} presents the detailed architecture of the 3D recurrent R-U-Net with HF and LF data.  The input model is of dimensions $n_x^h \times n_y^h \times n_z^h \times 1$. The HF and LF models share the same encoder, 3D convLSTM and decoder networks (though these are updated in Step~3). Different output layers are applied to map the decoder output to LF data (of dimensions $n_x^l \times n_y^l \times n_z^l \times 1 \times n_t$) and to HF data (dimension of $n_x^h \times n_y^h \times n_z^h \times 1 \times n_t$).

The encoder is composed of four `conv' blocks and two residual blocks. Each conv block includes one 3D convolutional layer, a batch normalization layer and ReLU nonlinear activation. The residual block includes one conv block with 64 filters of size $3\times 3 \times 3$ with stride $(1, 1, 1)$, followed by one 3D convolutional layer and a batch normalization layer. The output of the residual block is the sum of the input of the residual block and the output of the second convolutional layer. The output of the encoder is fed to the 3D convLSTM net, which uses 64 filters of size $3\times 3 \times 3$ with stride $(1, 1, 1)$ in all LSTM blocks. The output size of the 3D convLSTM net is $(\frac{n_x^h}{4}, \frac{n_y^h}{4}, \frac{n_z^h}{4}, 64)$ for $n_t$ time steps. The decoder includes two residual blocks and four `deconv' blocks. The deconv block is a stack of a 3D deconvolutional (upsampling) layer, a batch normalization layer and ReLU nonlinear activation. The decoder output is of dimensions $(n_x^h, n_y^h, n_z^h, 16, n_t)$.

\begin{table}[!ht]
\footnotesize
    \caption{Architecture of 3D recurrent R-U-Net with multifidelity data}
    \begin{tabular}{l | l| l}
    \hline
         Network & Layer & Output  \\
         \hline
         \multirow{7}{*}{Encoder ($\bm{\uptheta}_{enc}$)} & Input & $(n_x^h, n_y^h, n_z^h, 1)$\\
         & conv, 16 filters of size $3\times 3 \times 3$, stride 2 & $(\frac{n_x^h}{2}, \frac{n_y^h}{2}, \frac{n_z^h}{2}, 16)$\\
         & conv, 32 filters of size $3\times 3 \times 3$, stride 1 & $(\frac{n_x^h}{2}, \frac{n_y^h}{2}, \frac{n_z^h}{2}, 32)$\\
         & conv, 32 filters of size $3\times 3 \times 3$, stride 2 & $(\frac{n_x^h}{4}, \frac{n_y^h}{4}, \frac{n_z^h}{4}, 32)$\\
         & conv, 64 filters of size $3\times 3 \times 3$, stride 1 & $(\frac{n_x^h}{4}, \frac{n_y^h}{4}, \frac{n_z^h}{4}, 64)$\\
         & residual block, 64 filters of size $3\times 3 \times 3$, stride 1 & $(\frac{n_x^h}{4}, \frac{n_y^h}{4}, \frac{n_z^h}{4}, 64)$\\
         & residual block, 64 filters of size $3\times 3 \times 3$, stride 1 & $(\frac{n_x^h}{4}, \frac{n_y^h}{4}, \frac{n_z^h}{4}, 64)$\\
         \hline
         3D ConvLSTM $(\bm{\uptheta}_{convLSTM})$ & convLSTM3D, 64 filters $3\times 3 \times 3$, stride 1 & $(\frac{n_x^h}{4}, \frac{n_y^h}{4}, \frac{n_z^h}{4}, 64, n_t)$\\
         \hline
         \multirow{6}{*}{Decoder ($\bm{\uptheta}_{dec}$)} & residual block, 64 filters of size $3\times 3 \times 3$, stride 1 & $(\frac{n_x^h}{4}, \frac{n_y^h}{4}, \frac{n_z^h}{4}, 64, n_t)$\\
         & residual block, 64 filters of size $3\times 3 \times 3$, stride 1 & $(\frac{n_x^h}{4}, \frac{n_y^h}{4}, \frac{n_z^h}{4}, 64, n_t)$\\
         & deconv, 64 filters of size $3\times 3 \times 3$, stride 1 & $(\frac{n_x^h}{4}, \frac{n_y^h}{4}, \frac{n_z^h}{4}, 64, n_t)$\\
         & deconv, 32 filters of size $3\times 3 \times 3$, stride 2 & $(\frac{n_x^h}{4}, \frac{n_y^h}{4}, \frac{n_z^h}{4}, 32, n_t)$\\
         & deconv, 32 filters of size $3\times 3 \times 3$, stride 1 & $(\frac{n_x^h}{2}, \frac{n_y^h}{2}, \frac{n_z^h}{2}, 32, n_t)$\\
         & deconv, 16 filters of size $3\times 3 \times 3$, stride 2 & $(n_x^h, n_y^h, n_z^h, 16, n_t)$\\
         \hline
         Output Layers (LF, $\bm{\uptheta}_{output}^l$) & multi conv, 1 filters of size $3\times 3 \times 3$, stride 2 & $(n_x^l, n_y^l, n_z^l, 1, n_t)$ \\
         \hline
         Output Layers (HF, $\bm{\uptheta}_{output}^h$) & conv, 1 filters of size $3\times 3 \times 3$, stride 1 & $(n_x^h, n_y^h, n_z^h, 1, n_t)$\\
      \hline   
    \end{tabular}
    \label{tab:recurrent_runet}
\end{table}

In the system considered in this work, the fine-scale (HF) geomodel is defined on a grid of dimensions $80 \times 80 \times 20$, and it is upscaled to a $20 \times 20 \times 10$ LF grid. For the LF surrogate model in Step~1, the output layers include two 3D convolutional layers to map the decoder output to LF output of dimensions $(n_x^l, n_y^l, n_z^l, 1, n_t)$. The HF output layer in Steps~2 and 3 is a 3D convolutional layer that generates output of dimensions $(n_x^h, n_y^h, n_z^h, 1, n_t)$. 

Data preprocessing is essential, and the detailed treatments impact the performance of the overall procedure. In this study the network is applied to predict pressure and saturation for binary geomodels with wells operating under BHP control. The geomodels $\mathbf{m}^h$ are defined in terms of ones (indicating the cell contains sand) and zeros (indicating shale/mud), so no preprocessing of $\mathbf{m}^h$ is required. Saturation values are in range of 0 to 1, so these can also be used directly. For pressure normalization, the BHPs of the injectors and producers are taken as the maximum and minimum values, respectively. Min-max normalization is applied for both the LF and HF pressure output. 

The minimization applied in Step~1 training, involving LF data and predictions, is expressed as 
\begin{equation}\label{eq:opt_lf}
\begin{split}
    (\bm{\uptheta}^l)^* =  \argmin_{\bm{\uptheta}^l} & \frac{1}{n_{smp}^l}\frac{1}{n_t}\sum_{i = 1}^{n_{smp}^l}\sum_{t = 1}^{n_t}||\hat{\textbf{x}}_i^{l, t} - \textbf{x}_i^{l, t}||_2^2 \\
    & + \lambda_w^{l} \frac{1}{n_{smp}^l}\frac{1}{n_t}\frac{1}{n_w^l} \sum_{i = 1}^{n_{smp}^l}\sum_{t = 1}^{n_t}\sum_{w = 1}^{n_w^l} ||\hat{{x}}_i^{l, t, w} - {x}_i^{l, t, w}||_2^2, 
\end{split}
\end{equation}
where $\lambda_w^{l}$ represents the additional weight for the $n_w^l$ well blocks in the LF model. For Step~2 and Step~3 training (HF data and predictions) the minimization problem is 
\begin{equation}\label{eq:opt_hf}
\begin{split}
    (\bm{\uptheta}^h)^* =  \argmin_{\bm{\uptheta}^h} & \frac{1}{n_{smp}^{h}}\frac{1}{n_t}\sum_{i = 1}^{n_{smp}^{h}}\sum_{t = 1}^{n_t}||\hat{\textbf{x}}_i^{h, t} - \textbf{x}_i^{h, t}||_2^2 \\
    & + \lambda_w^{h} \frac{1}{n_{smp}^{h}}\frac{1}{n_t}\frac{1}{n_w^h} \sum_{i = 1}^{n_{smp}^{h}}\sum_{t = 1}^{n_t}\sum_{w = 1}^{n_w^h} ||\hat{{x}}_i^{h, t, w} - {x}_i^{h, t, w}||_2^2, 
\end{split}
\end{equation}
where $\lambda_w^{h}$ is the well-block weighting for the $n_w^h$ well blocks in the HF model. Pressure quantities in Eqs.~\ref{eq:opt_lf} and \ref{eq:opt_hf} are normalized as described earlier. Following training, the surrogate model provides flow predictions for a new geomodel in about 0.05~seconds on a single GPU.

\section{Surrogate Model Evaluation} \label{sec:surr_result}

In this section, we describe the problem setup and then present detailed results using the transfer-learning-based surrogate model. Comparisons with flow results from HF simulation, LF simulation, and a reference deep-learning surrogate procedure are provided.

\subsection{Problem Setup}

The 3D channelized system considered in this work corresponds to that used in~\cite{tang2021history}, though the specific realizations here are all newly generated. The well types and completion intervals are also different from those in \cite{tang2021history}. The models are binary and include channel sand and background mud/shale. The realizations are randomly generated through application of a convolutional neural network -- principal component analysis (CNN-PCA) parameterization procedure~\cite{liu20213d}. The CNN-PCA representation itself is trained using reference realizations generated through application of geostatistical software (see \cite{liu20213d} for details on the parameterization procedure).

\begin{figure}[!htb]
\centering
\begin{minipage}{.32\linewidth}\centering
\includegraphics[trim = 60 100 60 85, clip, width=\linewidth]{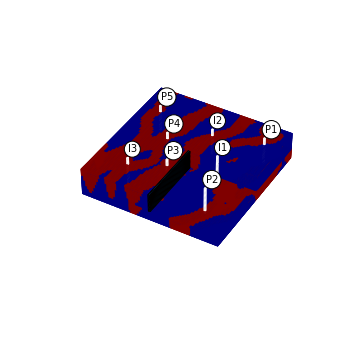}
\subcaption{CNN-PCA realization~1}
\end{minipage}
\begin{minipage}{.32\linewidth}\centering
\includegraphics[trim = 60 120 60 50, clip, width=\linewidth]{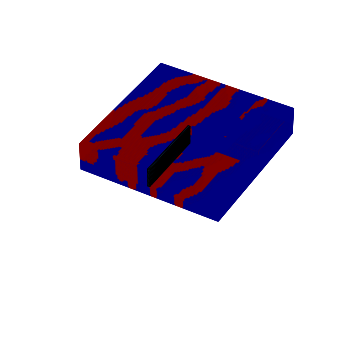}
\subcaption{CNN-PCA realization~2}
\end{minipage}
\begin{minipage}{.32\linewidth}\centering
\includegraphics[trim = 60 120 60 50, clip, width=\linewidth]{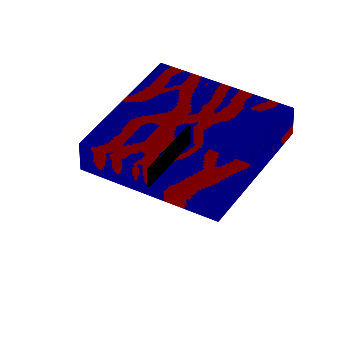}
\subcaption{CNN-PCA realization~3}
\end{minipage}
\caption{CNN-PCA geomodel realizations conditioned to hard data at eight well locations. Well locations are shown in (a). Realization in (c) corresponds to `true' model used for history matching. Cutaway view used for better 3D visualization.} \label{fig:perm}
\end{figure}

The 3D CNN-PCA model in~\cite{tang2021history} is applied to generate the geological realizations used for training and testing. The fine-scale (HF) geomodels are defined on an $80 \times 80 \times 20$ grid, with blocks of size 20~m $\times$ 20~m $\times$ 2~m. Figure~\ref{fig:perm} presents three geomodel realizations. Well locations are shown in Fig.~\ref{fig:perm}(a). Porosity is set to 0.25 in sand facies and 0.1 in mud/shale facies. Permeability is prescribed as 2000~md for sand and 20~md for mud/shale. The realization in Fig.~\ref{fig:perm}(c) is used for some of the results shown later in this section, and it also represents the `true' model applied for history matching in Section~\ref{sec:post_surr}.  

The geological realizations are conditioned to facies type at well locations. In contrast to the setup in~\cite{tang2021history}, here there are three injection wells (denoted I1 to I3) and five production wells (denoted P1 to P5), as shown in Fig.~\ref{fig:perm}(a). The initial reservoir pressure is 325~bar. The injectors and producers operate at fixed bottom-hole pressures (BHPs) of 330~bar and 310~bar, respectively. Wells I2, I3, P3, P4 and P5 are completed (open to flow) in the top eight layers of the model (layers~1-8), while wells I1, P1 and P2 are completed in layers~13-20.

\begin{figure}[!htb]
\centering
\includegraphics[width = 0.5\textwidth]{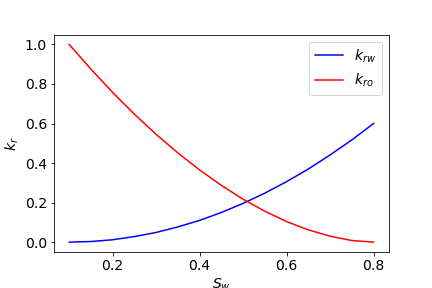}
\caption{Oil--water relative permeability curves.}
\label{fig:relaPerm}
\end{figure}

The flow problem involves two immiscible fluids (oil and water). Figure~\ref{fig:relaPerm} shows the oil-water relative permeability curves used for both the HF and LF flow simulations. The initial water and oil saturations are 0.1 and 0.9. Water viscosity is constant at 0.31~cp; oil viscosity varies with pressure and is 1.14~cp at 325~bar. All flow simulations are performed using ADGPRS~\cite{zhou2012parallel}. The simulation time frame is 1000~days. The global transmissibility upscaling procedure described in Section~\ref{sec:method_upscaling} is applied to generate the LF models. These models are defined on grids of dimensions $20 \times 20 \times 10$, which corresponds to an upscaling factor of 32.  

\subsection{Surrogate Model Training and Overall Performance}

The process for generating the multifidelity data used for training is as follows. We generate 2500 HF binary geomodel realizations with CNN-PCA. These realizations are upscaled to LF models using the procedure described in Section~\ref{sec:method_upscaling}. A total of $n_{smp}^l = 2500$ LF simulation runs are performed with ADGPRS. These represent the samples used in Step~1 of the training procedure. The pressure and saturation fields for these LF simulations are collected at $n_t = 10$ time steps (specifically at 50, 100, 150, 300, 400, 500, 600, 700, 850 and 1000~days). Separate surrogate models are trained for pressure and saturation with these 2500 LF simulation results. The hyperparameters affecting performance include the learning rate, the weight on data at well blocks, the number of epochs, and the batch size. For Step~1 of the training, we set $\lambda_w^l=20$ and the batch size to 4. The networks are trained for 150~epochs using the ADAM optimizer~\cite{kingma2014adam} with an initial learning rate of 0.003. 

The number of HF samples used for transfer learning affects the computational cost and performance of the surrogate model. It is useful to conduct numerical experiments to determine an optimal number of HF samples. We consider $n_{smp}^h = 100$, 200, 300 and 400 (representative) realizations selected from the full set of 2500 HF geomodels. These HF models are then simulated, and Steps~2 and 3 of the training framework are applied. It is important to note that even 400 HF training samples is not nearly enough to train the recurrent R-U-Net surrogate model ($O(2500)$ samples are required). 

The HF models to be simulated are selected through use of the clustering strategy suggested in~\cite{shirangi2016general}. Specifically, a K-means procedure is first applied on the normalized flow response data constructed from the full set of LF simulation results. A total of $n_{smp}^h$ clusters are generated, and a k-medoids method is then used to select the `centers' of the $n_{smp}^h$ clusters. These (cluster center) realizations comprise the representative set. HF simulation is then performed on all $n_{smp}^h$ realizations to provide the HF training data. 

The HF simulation results are used in Steps~2 and 3 of the training. Here $\lambda_w^h$ is set to 1000. This increased weighting ($\lambda_w^l = 20$ in Step~1) is applied because we now have a much lower fraction of well blocks with pressure and saturation data ($64/128,000=0.0005$ for HF in contrast to $32/4000=0.008$ for LF), and because HF well data are especially important to capture. The initial learning rates are set to 0.003 (Step~2) and 0.001 (Step~3). We train the output layers for 200 epochs in Step~2 and fine-tune the model for 100~epochs in Step~3. Because a small number of HF samples are used in Steps~2 and 3, the training times for these steps are much smaller than that for Step~1. The detailed training times for the framework will be provided later.

For comparison purposes, we also train a reference recurrent R-U-Net surrogate model using only high-fidelity samples. The same original set of 2500 HF geomodels is used for this training. The HF networks (one for pressure and one for saturation) are trained for 150~epochs with an initial learning rate of 0.003. 

A new set of $n_e = 400$ HF geological realizations is then generated and simulated. These represent the test samples used to evaluate the performance of the transfer-learning-based surrogate model. The performance of the surrogate model is compared to HF simulation results. The relative error for pressure for test sample $i$, denoted $\delta_p^i$, for $i = 1, \dots, n_e$, is given by 
\begin{equation} \label{eq:p_error}
    \delta_p^i = \frac{1}{n^h n_t}\sum_{j=1}^{n^h}\sum_{t=1}^{n_t} \frac{|\hat{p}^t_{i,j} - p^t_{i,j}|}{p_{max} - p_{min}}.
\end{equation}
Here $p^t_{i,j}$ is the HF pressure from simulation output in grid block $j$ at time $t$ for test sample $i$, $\hat{p}^t_{i,j}$ is the corresponding pressure output for the surrogate model, and $p_{max}$ and $p_{min}$ are the injector and producer BHPs (330~bar and 310~bar). The relative error for saturation is given by
\begin{equation} \label{eq:S_error}
    \delta_S^i = \frac{1}{n^h n_t}\sum_{j=1}^{n^h}\sum_{t=1}^{n_t} \frac{|\hat{S}^t_{i,j} - S^t_{i,j}|}{S_{i, j}^t},
\end{equation}
for $i = 1, \dots, n_e$, where $S^t_{i,j}$ and $\hat{S}^t_{i,j}$ represent saturation from HF simulation and the surrogate model. The initial (and minimum) water saturation is 0.1, so the denominator in Eq.~\ref{eq:S_error} does not approach zero.

\begin{figure}[!htb]
\centering
\includegraphics[width = 0.7\textwidth]{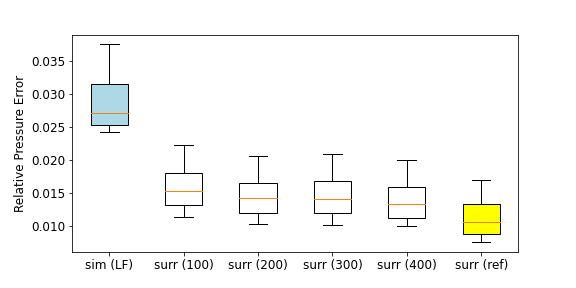}
\caption{Relative pressure errors for 400 test samples (compared to HF simulation results) for LF (coarse) simulation, surrogate model trained with 100, 200, 300 and 400 HF samples, and reference surrogate model trained with 2500 HF simulations.}\label{fig:err_multi_p}
\end{figure}

\begin{figure}[!htb]
\centering
\includegraphics[width = 0.7\textwidth]{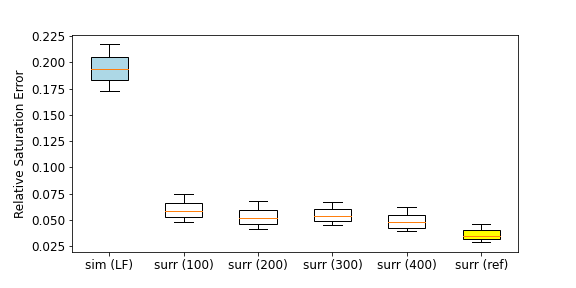}
\caption{Relative saturation errors for 400 test samples (compared to HF simulation results) for LF (coarse) simulation, surrogate model trained with 100, 200, 300 and 400 HF samples, and reference surrogate model trained with 2500 HF simulations.}\label{fig:err_multi_sat}
\end{figure}

Relative errors for pressure and saturation for the 400 test samples, computed using Eqs.~\ref{eq:p_error} and \ref{eq:S_error}, are shown in Figs.~\ref{fig:err_multi_p} and~\ref{fig:err_multi_sat}. These errors are presented as box plots. The top and bottom of each box show the P$_{75}$ and P$_{25}$ (75th and 25th percentile) relative errors, and the solid red line provides the P$_{50}$ result. The maximum and minimum relative errors are indicated by the lines extending beyond the boxes. In the figures, the blue boxes display the relative errors associated with LF (coarse-scale) simulation results. The yellow boxes indicate the relative errors for the reference surrogate model (average errors of 1.18\% for pressure and 3.70\% for saturation) -- these represent the best performance that could be expected from the transfer-learning-based surrogate. The white boxes show the relative errors for the transfer-learning-based surrogate model trained with different numbers of HF samples.  

The relative errors for the LF simulation results (blue boxes) are computed by projecting the LF pressure or saturation solution to the 80 $\times$ 80 $\times$ 20 grid, which gives pressure or saturation fields that are constant over 4 $\times$ 4 $\times$ 2 regions. Equations analogous to Eqs.~\ref{eq:p_error} and \ref{eq:S_error} are then applied to compute $\delta_p^i$ and $\delta_S^i$, for $i = 1, \dots, n_e$. Particularly in the case of saturation, the LF error is large. This is mainly due to errors around fluid fronts in the LF simulations. 

A key observation from Figs.~\ref{fig:err_multi_p} and~\ref{fig:err_multi_sat} is that the errors from the transfer-learning-based surrogate are considerably less than those from LF simulation. This accuracy is achieved even though the significant majority of the training is accomplished using LF models. Though there is a slight trend of decreasing relative error with increasing $n^h_{smp}$, the use of more than 100 HF samples has only a minor impact on results. For both pressure and saturation, the errors for the reference surrogate (yellow boxes) are less than those with the transfer-learning-based surrogate, as would be expected. The differences are relatively small, however.

In many subsurface flow problems, the flow rates associated with wells represent key quantities of interest. Here we calculate an aggregate relative well rate error $\delta_r^i$ as
\begin{equation} \label{eq:flow_error}
\begin{split}
    \delta_r^i & = \frac{1}{N_{\text{inj}}}\sum_{j=1}^{N_{\text{inj}}} \frac{\int_0^T |\hat{q}_{j,i}^{w, inj}(t) - q_{j,i}^{w, inj}(t)| dt}{\int_0^T |(q_{j,i}^{w, inj}(t)| dt} \\
    & + \frac{1}{N_{\text{prod}}}\sum_{j=1}^{N_{\text{prod}}} \frac{\int_0^T |\hat{q}_{j,i}^{w, prod}(t) - q_{j,i}^{w, prod}(t)| dt}{\int_0^T |(q_{j,i}^{w, prod}(t)| dt} \\
    & + \frac{1}{N_{\text{prod}}}\sum_{j=1}^{N_{\text{prod}}} \frac{\int_0^T |\hat{q}_{j,i}^{o, prod}(t) - q_{j,i}^{o, prod}(t)| dt}{\int_0^T |(q_{j,i}^{o, prod}(t)| dt} ,
\end{split}
\end{equation}
for $i = 1, \dots, n_e$, where $N_{\text{inj}}$ and $N_{\text{prod}}$ are the number of injectors and producers and $T$ is the total simulation time. Here $q_{j,i}^{w, inj}(t)$ denotes the injection rate of injector $j$ at time $t$ for test sample $i$ from HF simulation, and $\hat{q}_{j,i}^{w, inj}(t)$ is the corresponding injection rate for the surrogate model. Analogously, the quantities $q_{j,i}^{w, prod}(t)$ and $q_{j,i}^{o, prod}(t)$ are the HF water and oil production rates, and $\hat{q}_{j,i}^{w, prod}(t)$ and $\hat{q}_{j,i}^{o, prod}(t)$ are the corresponding production rates for the surrogate model. 

\begin{figure}[!htb]
\centering
\includegraphics[width = 0.7\textwidth]{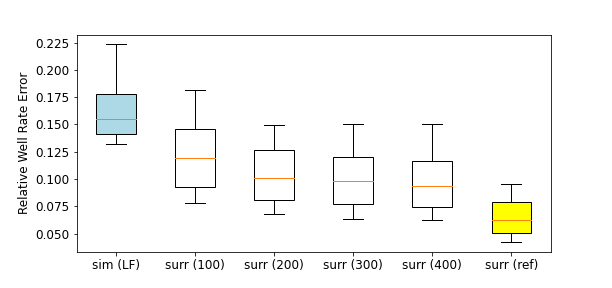}
\caption{Aggregate relative well rate errors for 400 test samples (compared to HF simulation results) for LF (coarse) simulation, surrogate model trained with 100, 200, 300 and 400 HF samples, and reference surrogate model trained with 2500 HF simulations.} \label{fig:err_multi_rate}
\end{figure}

Figure~\ref{fig:err_multi_rate} displays the aggregate relative well rate errors. These errors are larger than the pressure and saturation errors in Figs.~\ref{fig:err_multi_p} and~\ref{fig:err_multi_sat}, likely because well rates depend nonlinearly on predicted pressure and saturation values in just a few particular grid blocks (see Eq.~\ref{eq:well}). There is again a slight decrease in relative error with increasing $n^h_{smp}$. The maximum relative error, for example, decreases from 0.18 with $n_{smp}^h = 100$ to 0.15 for $n_{smp}^h = 200$, and then stays nearly constant with increasing $n_{smp}^h$. We thus conclude that 200 HF samples is a reasonable choice for this case. In future work, it may be worthwhile to investigate the detailed selection strategy and approaches for determining the optimal value of $n_{smp}^h$. In the remainder of this paper, we use $n_{smp}^h = 200$ in all transfer-learning-based surrogate results.

\subsection{Computational Costs of Surrogate Model Construction}

The computational cost of the transfer learning workflow includes the generation of LF and HF simulation data and the time for surrogate model training. Flow simulations and the upscaling computations are conducted with ADGPRS. As noted earlier, each HF simulation takes around 20~minutes on a single CPU. The overall cost for simulating 2500 HF samples is thus around 833~hours. In our assessment here, we evaluate simulation requirements in units of HF simulation runs. Thus, for the reference surrogate model, where all 2500 training runs are performed with HF simulation, the computational cost is $C_{\text{ref}} = 2500$. 

Linear solutions dominate the computational cost of the two-phase flow simulations performed in this work. Linear solvers ideally require $O(N)$ computation, where $N$ is the number of equations/unknowns in the linear system (super-linear scaling occurs in many cases). In HF simulations, $N=2n^h=256,000$, while in LF simulations, $N=2n^l=8000$. Both HF and LF simulations require $\sim$30--35 time steps with an average of $\sim$6 Newton iterations per time step (this relatively large number of Newton iterations results from taking large time steps). Thus $\sim$200 linear solutions are required for each (HF or LF) two-phase flow simulation run. The cost of simulating 2500 LF models is therefore about $C_{\text{LF}} = 2500  \times \frac{1}{32} \approx 78$. Upscaling computations involve HF models, but only one linear solution with one equation/unknown is required. Thus the cost of upscaling 2500 models is about $C_{\text{upscale}} = 2500 \times \frac{1}{200} \times \frac{1}{2} \approx 6$.

The total simulation cost of generating multifidelity data, with $n^l_{smp}=2500$ and $n^h_{smp}=200$, can be estimated as $C_{\text{multi}} = 6+78+200=284$. Compared to $C_{\text{ref}} = 2500$, this represents a computational savings of nearly 90\%. The estimated computational time for generating the multifidelity data is about 95~hours, compared to 833~hours with HF data. Note that super-linear scaling in the linear solver will lead to larger savings, while significant amounts of simulation overhead/startup time will lead to smaller savings.

The training time depends on the complexity of the surrogate model, the number of training samples, and the number of epochs required for convergence. We train two separate surrogate models for pressure and saturation in parallel on two Nvidia Tesla V100 GPUs. The training for the transfer learning framework includes training with LF samples, transfer learning training, and fine-tuning with HF samples. To achieve convergence, the Step~1 training requires 150~epochs. This can be reduced, however, by stopping this training early.

Table~\ref{tab:cost} presents cost and accuracy results for HF training and two multifidelity training strategies. The training process for the reference HF surrogate model converges in 15~hours (150~epochs, 360~seconds per epoch). For the surrogate model trained with multifidelity data with Step~1 iterated until convergence (MF training), the Step~1 training cost is about the same as the full HF training cost. This is because these trainings involve networks of similar complexity and the same number of training samples and epochs. As indicated in the table, the total training time for Steps~2 and 3 is 1.66~hours, which gives an overall training time of 16.66~hours. This is slightly larger than the reference HF training time. 

It is not necessary, however, to continue Step~1 training until convergence (150~epochs) since additional training with HF data will be performed. Accelerated training (right-most column of Table~\ref{tab:cost}) entails stopping Step~1 training at 70~epochs, at which point training loss is acceptable, although the training has not fully converged. Overall training time is now reduced to 8.66~hours. Relative errors with this approach are incrementally greater, but the computational savings are substantial. In subsequent results, in all cases the surrogate model is trained using this accelerated process. 

In summary, we see that the use of multifidelity data and transfer learning can lead to a nearly 90\% reduction in the computations associated with the training simulations, and a savings of about 42\% in training time. It is possible these speedups could be further improved with additional tuning, though this was not attempted here.

\begin{table}[!htb]
\caption{Computational costs of training reference and transfer-learning-based surrogate with multifidelity (MF) data (Step~1: training with LF data, Step~2: transfer learning with HF data, Step~3: fine-tuning with HF data). The relative errors are averages over 400 test samples}
\begin{center}
\footnotesize
\begin{tabular}{m{2.5cm}|m{3.8cm}|m{3.8cm}|m{3.8cm}}
\hline
  & HF training & MF training & Accl MF training\\
\hline
Step 1 & N/A
 & 150 ep$\times$360 s/ep
= 15 h & 70 ep$\times$360 s/ep
= 7 h
  \\ 
\hline
Step 2 & N/A & 200 ep$\times$15 s/ep
= 0.83 h & 200 ep$\times$15 s/ep
= 0.83 h \\

\hline
Step 3 & 150 ep$\times$360 s/ep = 15 h &  100 ep$\times$30 s/ep
= 0.83 h &  100 ep$\times$30 s/ep
= 0.83 h\\
\hline
Time & 15 h & 16.66 h & 8.66 h \\
\hline
$p$ error (${\bar \delta}_p$) & 1.18\% & 1.52\% & 1.58\% \\
\hline
$S$ error (${\bar \delta}_S$) & 3.70\% & 5.38\% & 5.57\% \\
\hline
\end{tabular}
\end{center}
\label{tab:cost}
\end{table}

\subsection{Pressure and Saturation Predictions}
\begin{figure}[!htb]
\centering
\begin{minipage}{.3\linewidth}\centering
\includegraphics[trim = 60 120 60 50, clip, width=\linewidth]{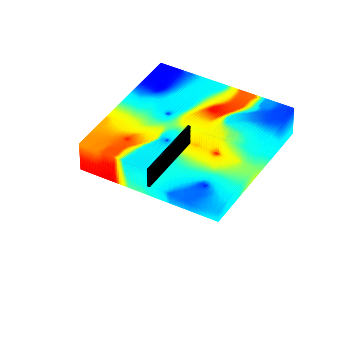}
\subcaption{Realization 1 (sim/HF)}
\end{minipage}
\begin{minipage}{.3\linewidth}\centering
\includegraphics[trim = 60 120 60 50, clip, width=\linewidth]{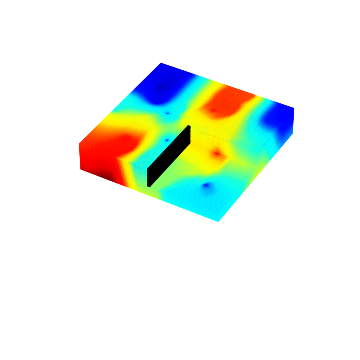}
\subcaption{Realization 2 (sim/HF)}
\end{minipage}
\begin{minipage}{.3\linewidth}\centering
\includegraphics[trim = 60 120 60 50, clip, width=\linewidth]{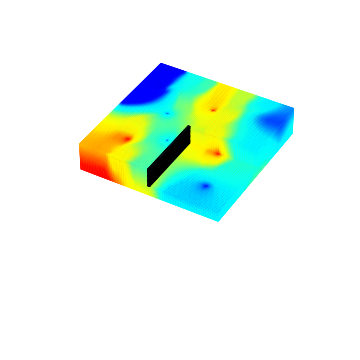}
\subcaption{Realization 3 (sim/HF)}
\end{minipage}
\begin{minipage}{.07\linewidth}\centering
\includegraphics[trim = 330 0 20 20, clip, width=\linewidth]{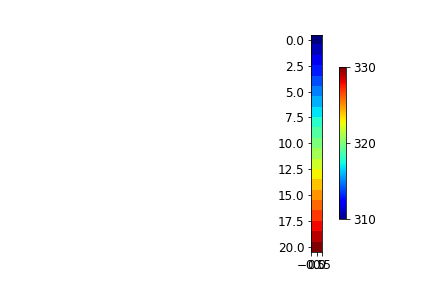}
\end{minipage}

\begin{minipage}{.3\linewidth}\centering
\includegraphics[trim = 60 120 60 50, clip, width=\linewidth]{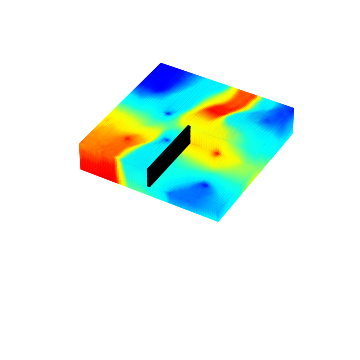}
\subcaption{Realization 1 (surr/ref)}
\end{minipage}
\begin{minipage}{.3\linewidth}\centering
\includegraphics[trim = 60 120 60 50, clip, width=\linewidth]{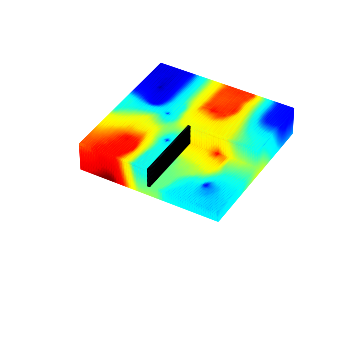}
\subcaption{Realization 2 (surr/ref)}
\end{minipage}
\begin{minipage}{.3\linewidth}\centering
\includegraphics[trim = 60 120 60 50, clip, width=\linewidth]{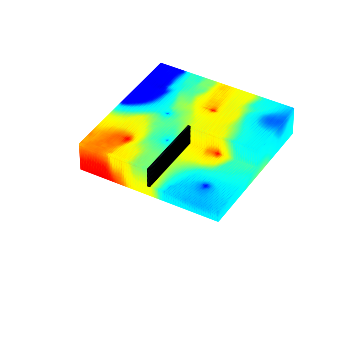}
\subcaption{Realization 3 (surr/ref)}
\end{minipage}
\begin{minipage}{.07\linewidth}\centering
\includegraphics[trim = 330 0 20 20, clip, width=\linewidth]{p_colorbar.png}
\end{minipage}

\begin{minipage}{.3\linewidth}\centering
\includegraphics[trim = 60 120 60 50, clip, width=\linewidth]{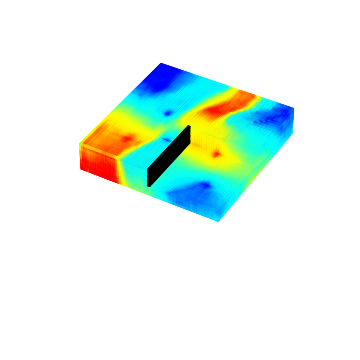}
\subcaption{Realization 1 (surr/multi)}
\end{minipage}
\begin{minipage}{.3\linewidth}\centering
\includegraphics[trim = 60 120 60 50, clip, width=\linewidth]{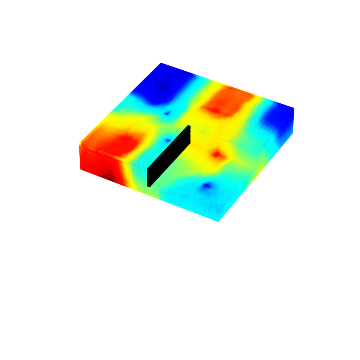}
\subcaption{Realization 2 (surr/multi)}
\end{minipage}
\begin{minipage}{.3\linewidth}\centering
\includegraphics[trim = 60 120 60 50, clip, width=\linewidth]{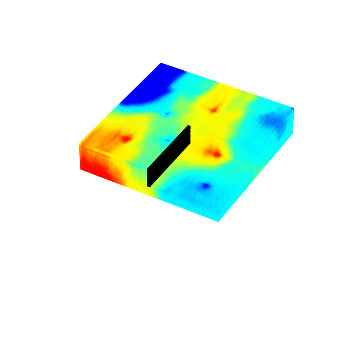}
\subcaption{Realization 3 (surr/multi)}
\end{minipage}
\begin{minipage}{.07\linewidth}\centering
\includegraphics[trim = 330 0 20 20, clip, width=\linewidth]{p_colorbar.png}
\end{minipage}
\caption{Pressure fields from HF simulation (top row), surrogate model trained with HF data (middle row), and transfer-learning-based surrogate model trained with multifidelity data (bottom row). All results are at 400~days. Cutaway views highlight 3D effects.}\label{fig:prior_p}
\end{figure}

To assess the performance of the transfer-learning-based surrogate model trained with multifidelity data, we now compare pressure and saturation predictions to results from HF simulation and the reference surrogate model (trained with 2500 HF simulation runs). The three realizations in Fig.~\ref{fig:perm}, which correspond to test-case realizations with relative pressure and saturation errors slightly larger than the median, are considered. Figure~\ref{fig:prior_p} shows the pressure fields for these three test cases at 400~days. The upper row displays the HF simulation results, the middle row shows results from the reference surrogate model, and the bottom row shows results from the transfer-learning-based surrogate model. There are noticeable differences in the pressure distributions between the three realizations, indicating a reasonable degree of variability over the test set. The reference surrogate model gives results that are visually close to the simulation results. The transfer-learning-based surrogate model also provides results that are in close visual agreement. Some discrepancies are evident, e.g., in the blue feature in the back right corner of the models in Fig.~\ref{fig:prior_p}(i) and (c), though these are relatively minor.  

\begin{figure}[!htb]
\centering
\begin{minipage}{.3\linewidth}\centering
\includegraphics[trim = 60 120 60 50, clip, width=\linewidth]{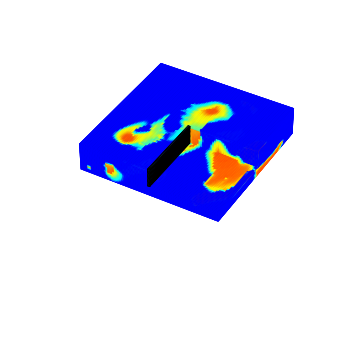}
\subcaption{Realization 1 (sim/HF)}
\end{minipage}
\begin{minipage}{.3\linewidth}\centering
\includegraphics[trim = 60 120 60 50, clip, width=\linewidth]{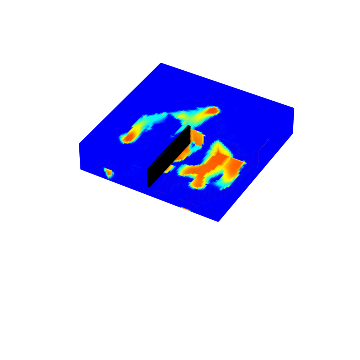}
\subcaption{Realization 2 (sim/HF)}
\end{minipage}
\begin{minipage}{.3\linewidth}\centering
\includegraphics[trim = 60 120 60 50, clip, width=\linewidth]{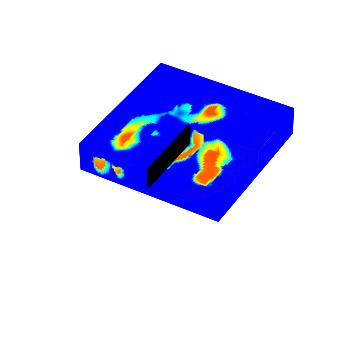}
\subcaption{Realization 3 (sim/HF)}
\end{minipage}
\begin{minipage}{.07\linewidth}\centering
\includegraphics[trim = 330 0 20 20, clip, width=\linewidth]{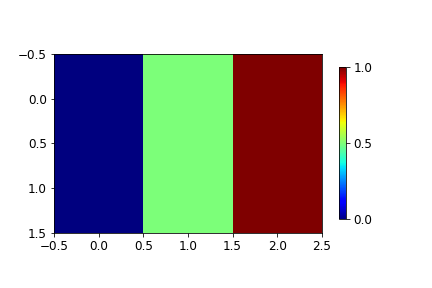}
\end{minipage}

\begin{minipage}{.3\linewidth}\centering
\includegraphics[trim = 60 120 60 50, clip, width=\linewidth]{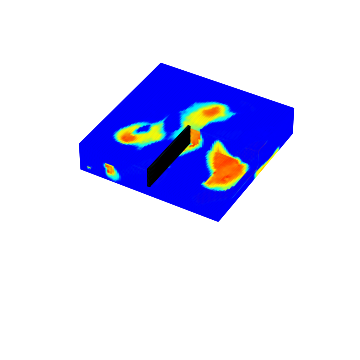}
\subcaption{Realization 1 (surr/ref)}
\end{minipage}
\begin{minipage}{.3\linewidth}\centering
\includegraphics[trim = 60 120 60 50, clip, width=\linewidth]{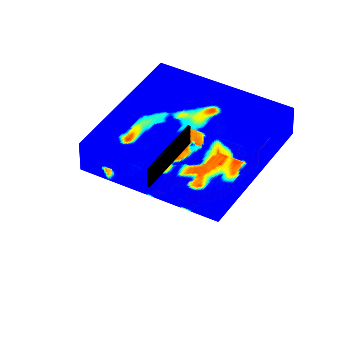}
\subcaption{Realization 2 (surr/ref)}
\end{minipage}
\begin{minipage}{.3\linewidth}\centering
\includegraphics[trim = 60 120 60 50, clip, width=\linewidth]{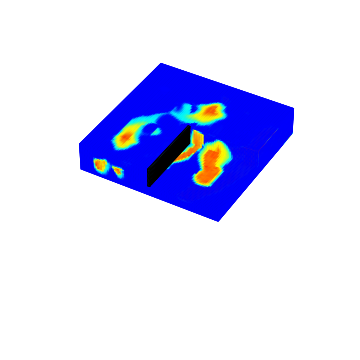}
\subcaption{Realization 3 (surr/ref)}
\end{minipage}
\begin{minipage}{.07\linewidth}\centering
\includegraphics[trim = 330 0 20 20, clip, width=\linewidth]{s_colorbar.png}
\end{minipage}

\begin{minipage}{.3\linewidth}\centering
\includegraphics[trim = 60 120 60 50, clip, width=\linewidth]{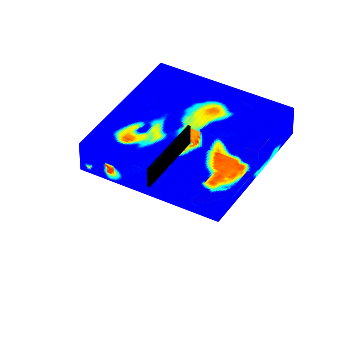}
\subcaption{Realization 1 (surr/multi)}
\end{minipage}
\begin{minipage}{.3\linewidth}\centering
\includegraphics[trim = 60 120 60 50, clip, width=\linewidth]{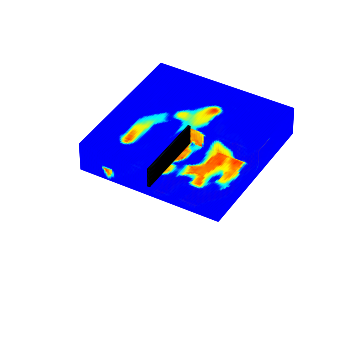}
\subcaption{Realization 2 (surr/multi)}
\end{minipage}
\begin{minipage}{.3\linewidth}\centering
\includegraphics[trim = 60 120 60 50, clip, width=\linewidth]{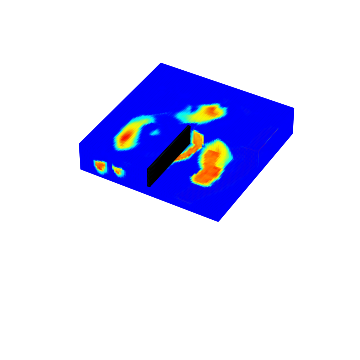}
\subcaption{Realization 3 (surr/multi)}
\end{minipage}
\begin{minipage}{.07\linewidth}\centering
\includegraphics[trim = 330 0 20 20, clip, width=\linewidth]{s_colorbar.png}
\end{minipage}

\caption{Water saturation fields from HF simulation (top row), surrogate model trained with HF data (middle row), and transfer-learning-based surrogate model trained with multifidelity data (bottom row). All results are at 1000~days.}\label{fig:prior_sat}
\end{figure}

Water saturation results for the same three realizations, at 1000~days (the end of the simulation), are presented in Fig.~\ref{fig:prior_sat}. We again observe variability between realizations and high accuracy in both the reference surrogate model and the transfer-learning-based surrogate model. The latter model is seen to provide sharp water saturation fronts, even though most of its training involves LF simulation results where fronts are diffused.

\begin{figure}[!htb]
\centering
\begin{minipage}{.45\linewidth}\centering
\includegraphics[trim = 60 0 0 0, clip, width=\linewidth]{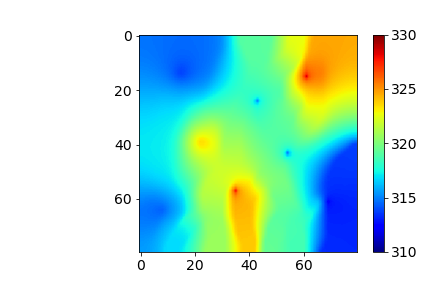}
\subcaption{Sim (HF)}
\end{minipage}
\begin{minipage}{.45\linewidth}\centering
\includegraphics[trim = 60 0 0 0, clip, width=\linewidth]{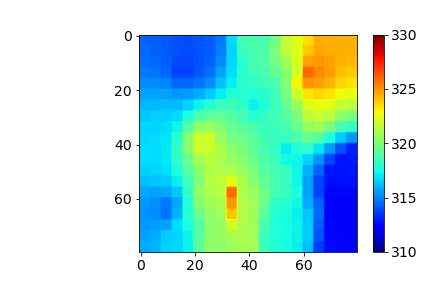}
\subcaption{Sim (LF)}
\end{minipage}
\begin{minipage}{.45\linewidth}\centering
\includegraphics[trim = 60 0 0 0, clip, width=\linewidth]{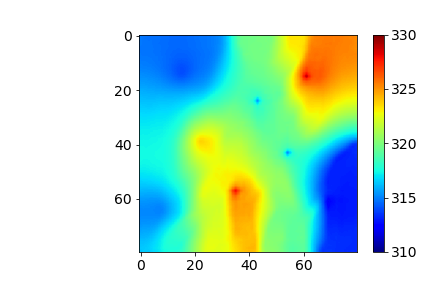}
\subcaption{Surr (ref)}
\end{minipage}
\begin{minipage}{.45\linewidth}\centering
\includegraphics[trim = 60 0 0 0, clip, width=\linewidth]{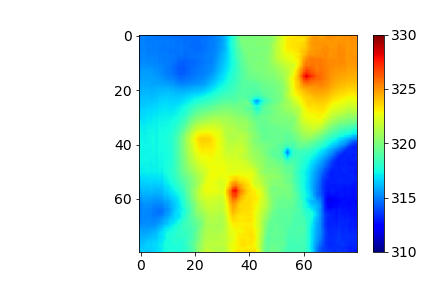}
\subcaption{Surr (multi)}
\end{minipage}
\caption{Pressure maps from HF and LF simulation, reference surrogate, and transfer-learning-based surrogate trained with multifidelity data. Results are for layer~8 from realization in Fig.~\ref{fig:perm}(c) at 1000~days.}\label{fig:2d_prior_p}
\end{figure}

We next show pressure and saturation maps at 1000~days for one particular layer (layer~8) in the realization in Fig.~\ref{fig:perm}(c). Figure~\ref{fig:2d_prior_p}(a) and (b) display the pressure fields from HF and LF simulation. In Fig.~\ref{fig:2d_prior_p}(b), the 20 $\times$ 20 LF result is mapped onto an 80 $\times$ 80 grid. Figure~\ref{fig:2d_prior_p}(c) and (d) show the pressure predictions from the two surrogate models. The reference surrogate results are in close agreement with the HF simulation result. Some differences are evident in the transfer-learning-based surrogate, though it is evident that some features that are not resolved in the LF simulation result in Fig.~\ref{fig:2d_prior_p}(b) are captured in the result in Fig.~\ref{fig:2d_prior_p}(d). Such details include the large yellow feature (midway in $x$, middle to bottom in $y$) and the light blue feature along the left edge (around the middle in $y$).

\begin{figure}[!htb]
\centering
\begin{minipage}{.45\linewidth}\centering
\includegraphics[trim = 60 0 0 0, clip, width=\linewidth]{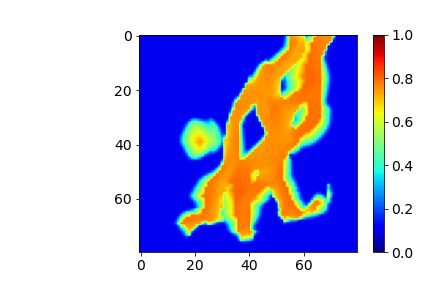}
\subcaption{Sim (HF)}
\end{minipage}
\begin{minipage}{.45\linewidth}\centering
\includegraphics[trim = 60 0 0 0, clip, width=\linewidth]{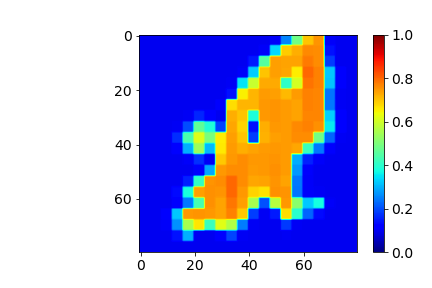}
\subcaption{Sim (LF)}
\end{minipage}
\begin{minipage}{.45\linewidth}\centering
\includegraphics[trim = 60 0 0 0, clip, width=\linewidth]{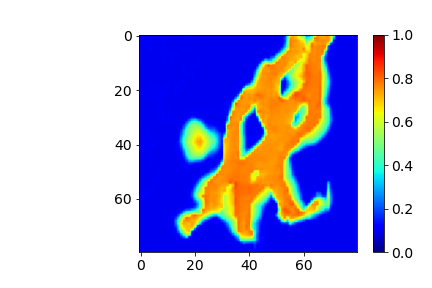}
\subcaption{Surr (ref)}
\end{minipage}
\begin{minipage}{.45\linewidth}\centering
\includegraphics[trim = 60 0 0 0, clip, width=\linewidth]{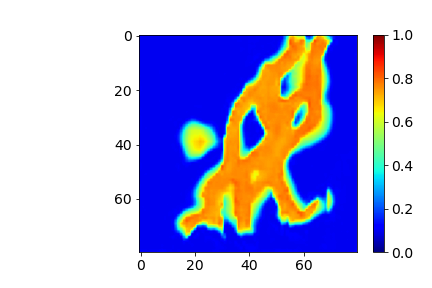}
\subcaption{Surr (multi)}
\end{minipage}
\caption{Saturation maps from HF and LF simulation, reference surrogate, and transfer-learning-based surrogate trained with multifidelity data. Results are for layer~8 from realization in Fig.~\ref{fig:perm}(c) at 1000~days.}\label{fig:2d_prior_sat}
\end{figure}

Analogous results for saturation appear in Fig.~\ref{fig:2d_prior_sat}. The LF simulation result for this quantity, in Fig.~\ref{fig:2d_prior_sat}(b), shows a high degree of smearing. The transfer-learning-based surrogate result in Fig.~\ref{fig:2d_prior_sat}(d) is in close visual agreement with the HF simulation result, and it shows much better front resolution than the LF simulation result. This illustrates the ability of the multifidelity procedure to provide high-quality HF surrogate predictions with training that is largely based on LF simulation data.

\begin{figure}[!htb]
\centering
\begin{minipage}{.6\linewidth}\centering
\includegraphics[trim = 0 0 0 0, clip, width=\linewidth]{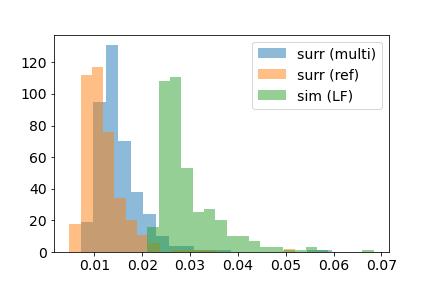}
\subcaption{Pressure}
\end{minipage}
\begin{minipage}{.6\linewidth}\centering
\includegraphics[trim = 0 0 0 0, clip, width=\linewidth]{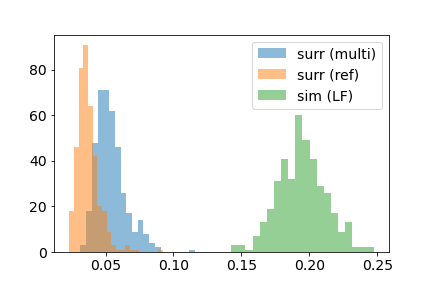}
\subcaption{Saturation}
\end{minipage}

\caption{Histograms of relative errors for pressure and saturation for the 400 test samples.}\label{fig:error_p_sat}
\end{figure}

We now present relative pressure and saturation error results for the 400 test samples. Figure~\ref{fig:error_p_sat} displays histograms of these errors, for pressure and saturation, for LF simulation and both surrogate model results. Errors are all relative to HF simulation results. The orange histograms show errors for the reference surrogate model predictions. These results are the most accurate, as expected, and display median pressure and saturation errors of 1.1\% and 3.5\%. The green histograms show relative errors for the LF simulation results. The median errors here are 2.7\% and 19.4\%. The errors for the surrogate model trained with multifidelity data (blue histograms) fall between the reference surrogate model errors and the LF simulation errors. Importantly, these errors (with medians 1.5\% and 5.5\%) are much closer to those of the reference surrogate model than to those from LF simulation. This again highlights the capabilities of the transfer-learning procedure.

\subsection{Well Rate Statistics}
\begin{figure}[!htb]
\centering
\begin{minipage}{.32\linewidth}\centering
\includegraphics[trim = 0 0 0 0, clip, width=\linewidth]{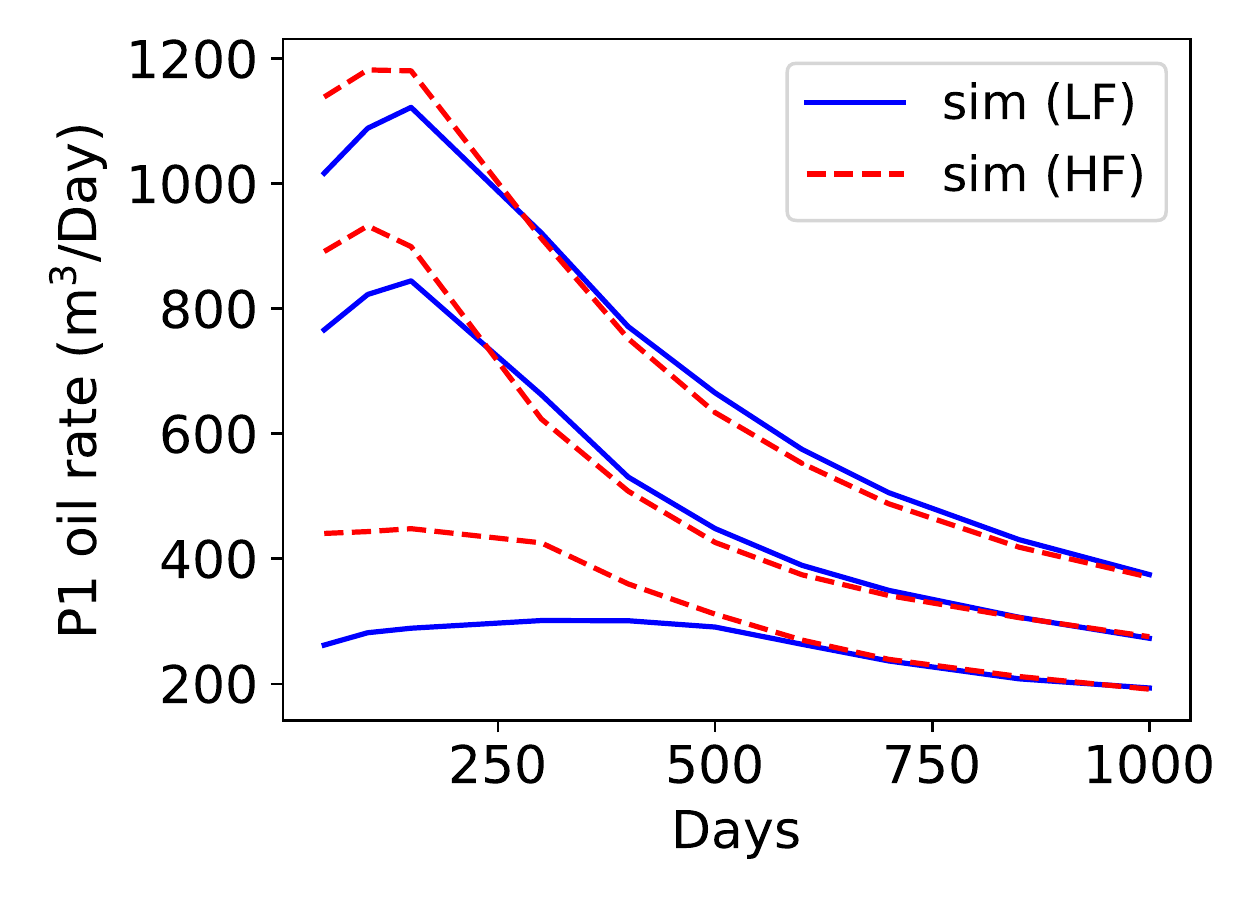}
\subcaption{P1 OPR -- HF and LF}
\end{minipage}
\begin{minipage}{.32\linewidth}\centering
\includegraphics[trim = 0 0 0 0, clip, width=\linewidth]{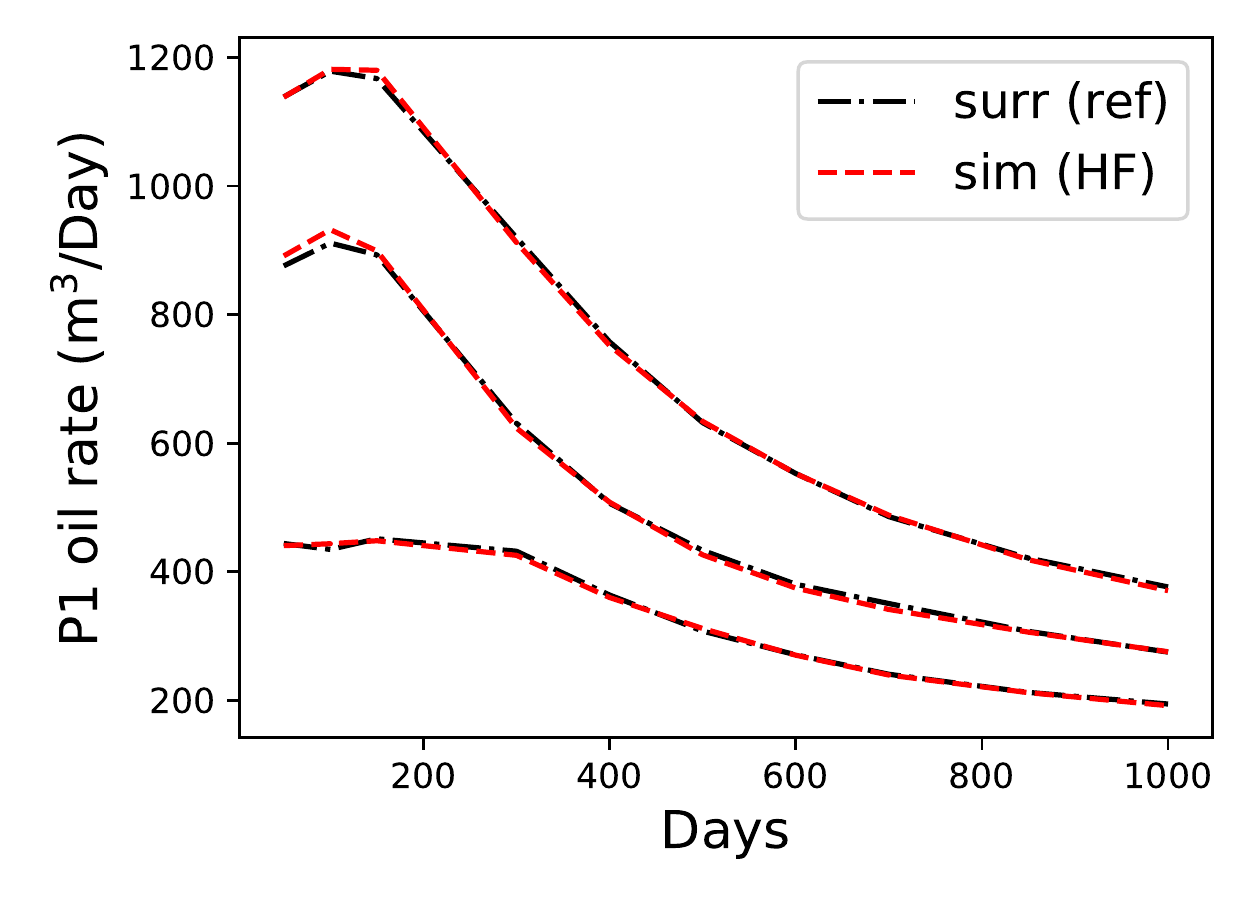}
\subcaption{P1 OPR -- HF and surr (ref)}
\end{minipage}
\begin{minipage}{.32\linewidth}\centering
\includegraphics[trim = 0 0 0 0, clip, width=\linewidth]{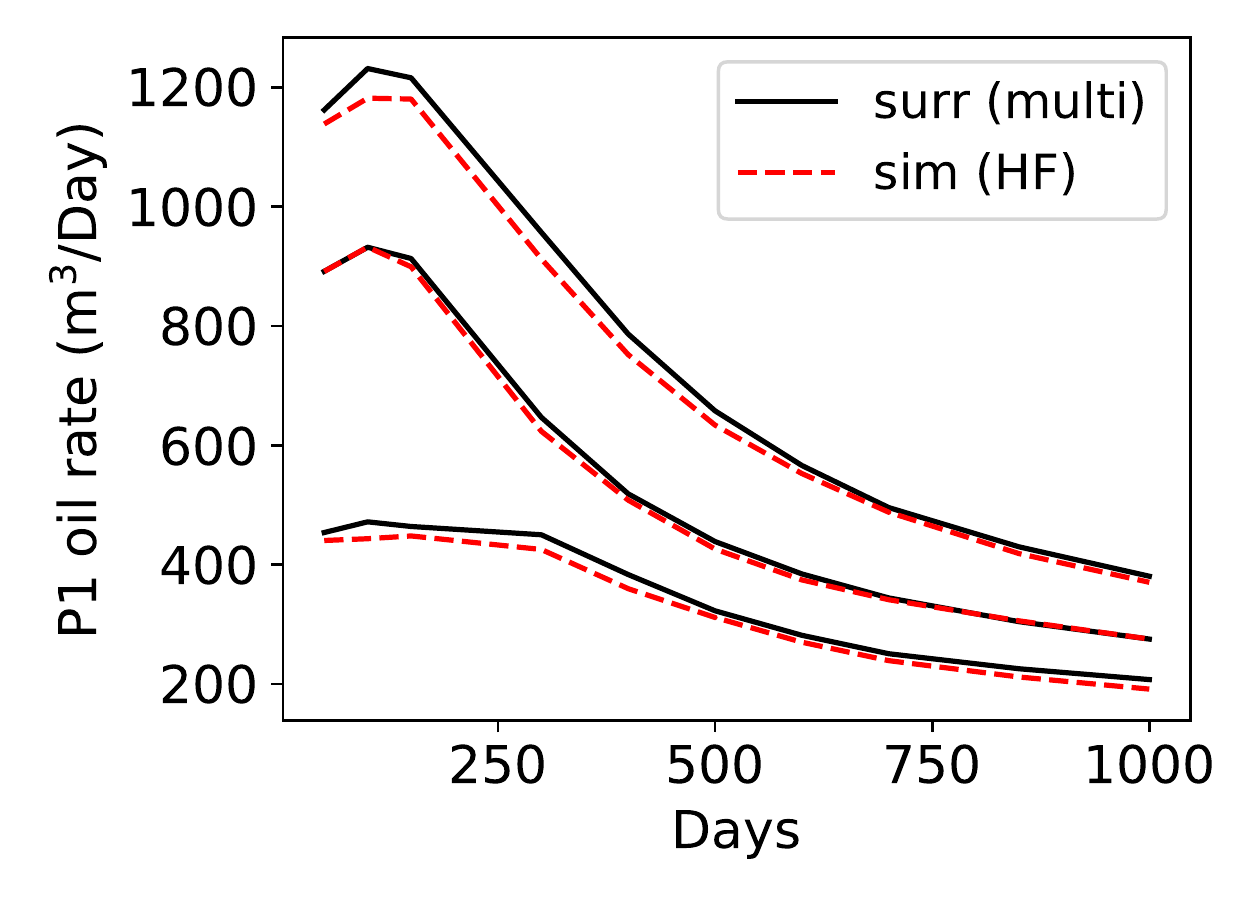}
\subcaption{P1 OPR -- HF and surr (multi)}
\end{minipage}

\begin{minipage}{.32\linewidth}\centering
\includegraphics[trim = 0 0 0 0, clip, width=\linewidth]{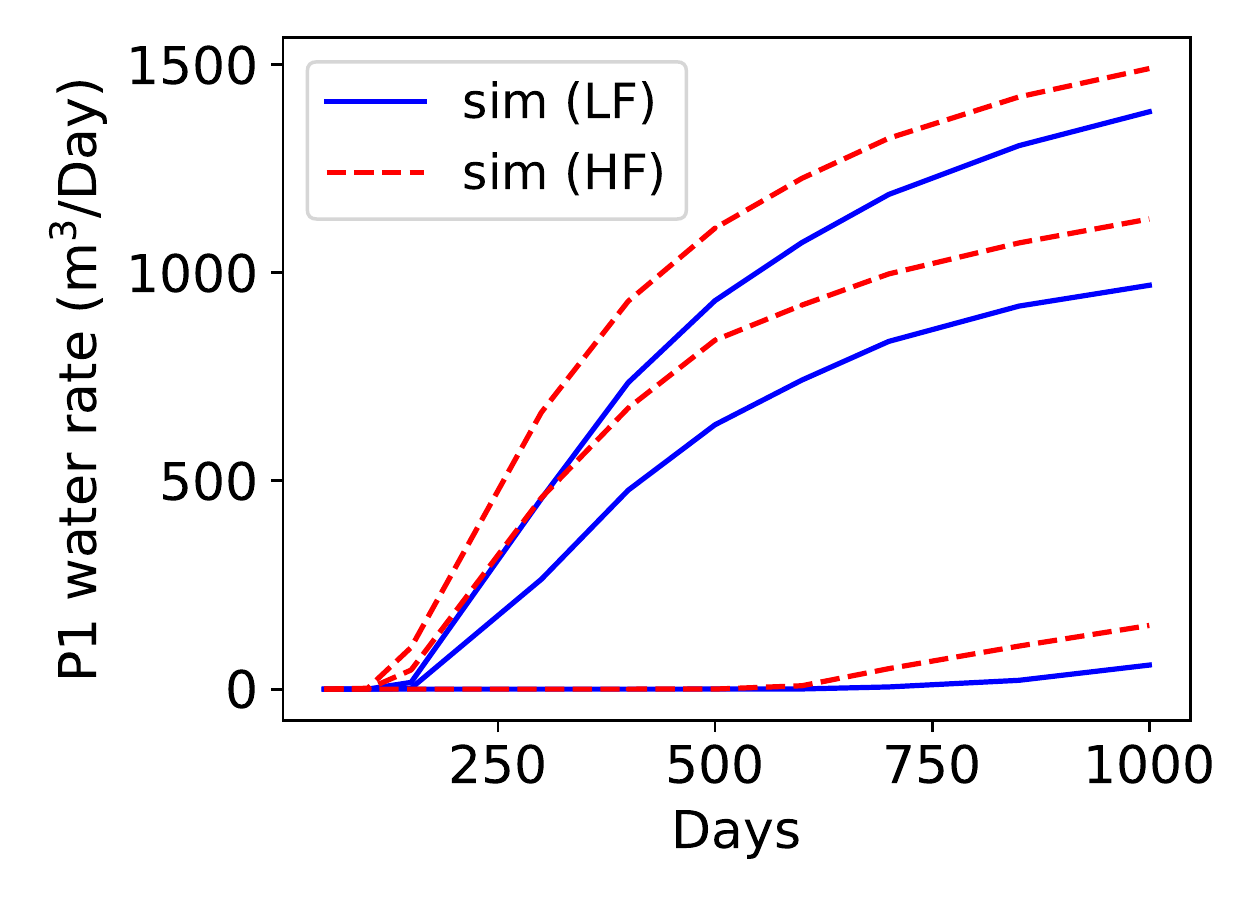}
\subcaption{P1 WPR -- HF and LF}
\end{minipage}
\begin{minipage}{.32\linewidth}\centering
\includegraphics[trim = 0 0 0 0, clip, width=\linewidth]{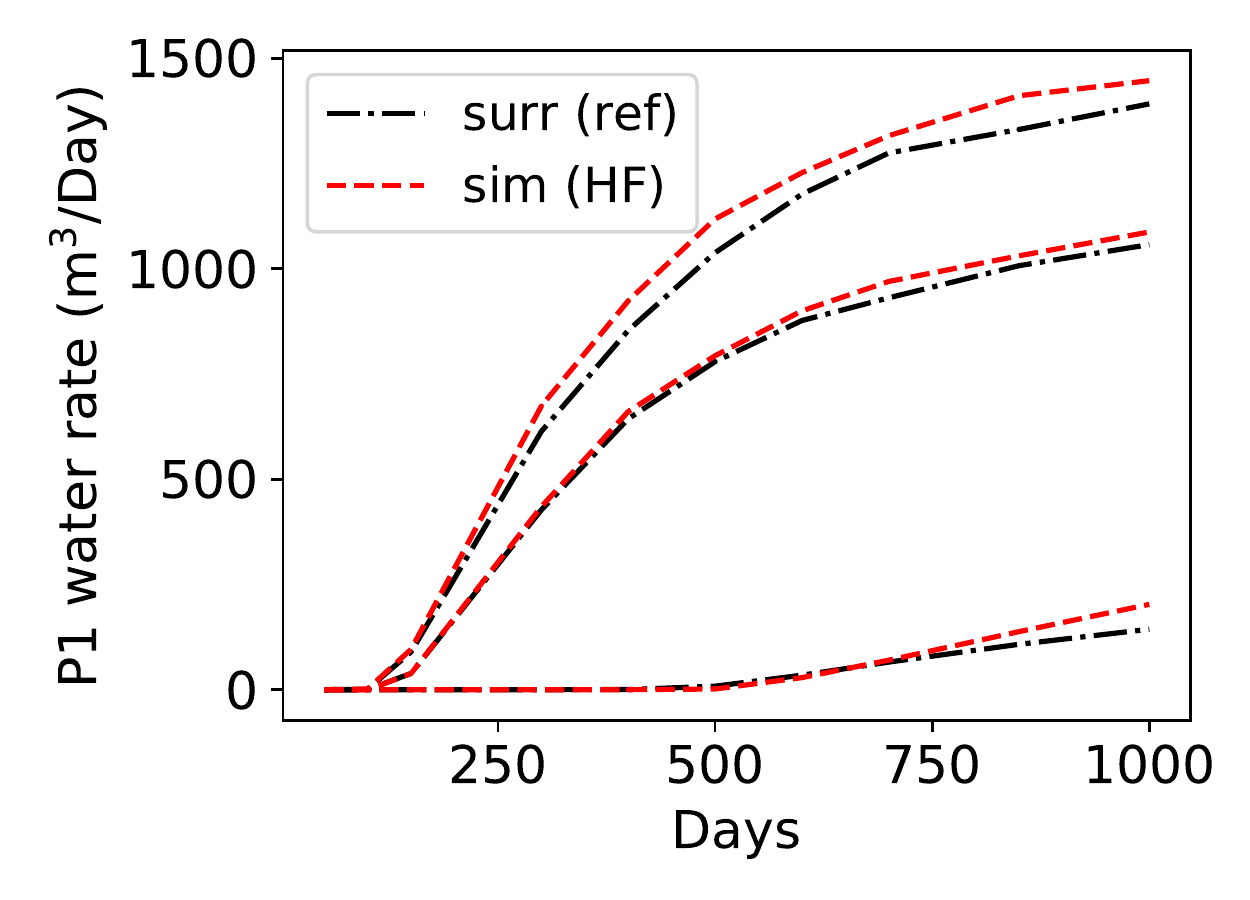}
\subcaption{P1 WPR -- HF and surr (ref)}
\end{minipage}
\begin{minipage}{.32\linewidth}\centering
\includegraphics[trim = 0 0 0 0, clip, width=\linewidth]{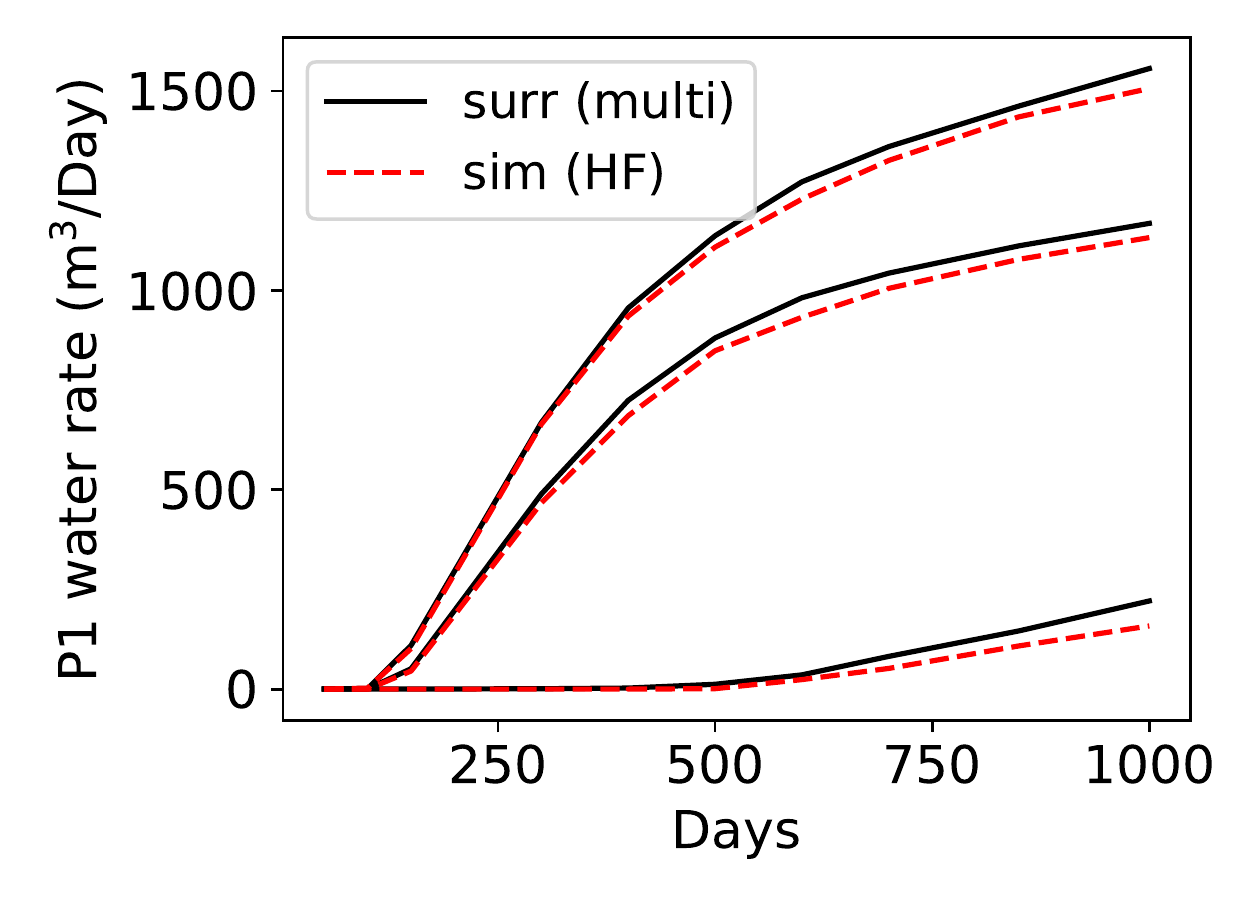}
\subcaption{P1 WPR -- HF and surr (multi)}
\end{minipage}

\begin{minipage}{.32\linewidth}\centering
\includegraphics[trim = 0 0 0 0, clip, width=\linewidth]{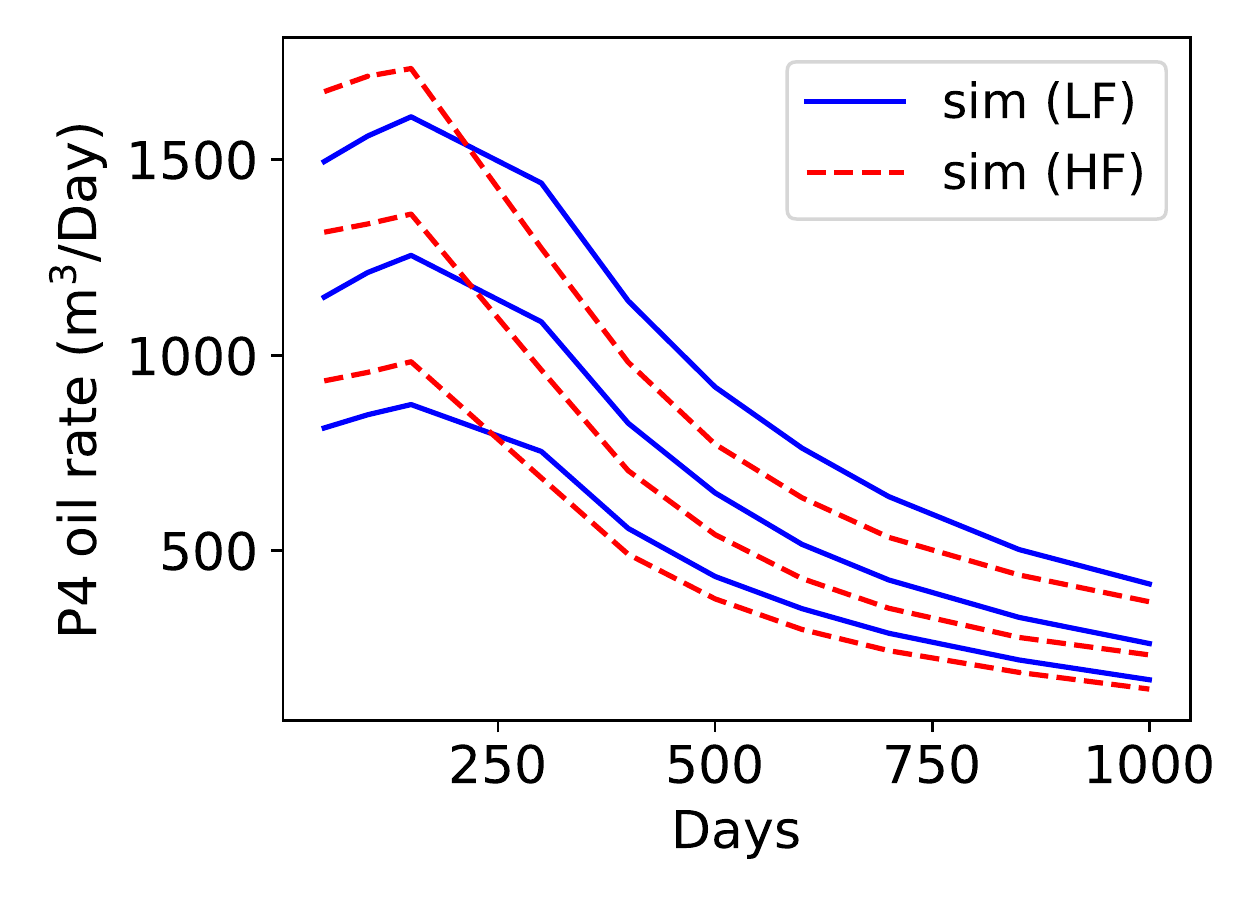}
\subcaption{P4 OPR -- HF and LF}
\end{minipage}
\begin{minipage}{.32\linewidth}\centering
\includegraphics[trim = 0 0 0 0, clip, width=\linewidth]{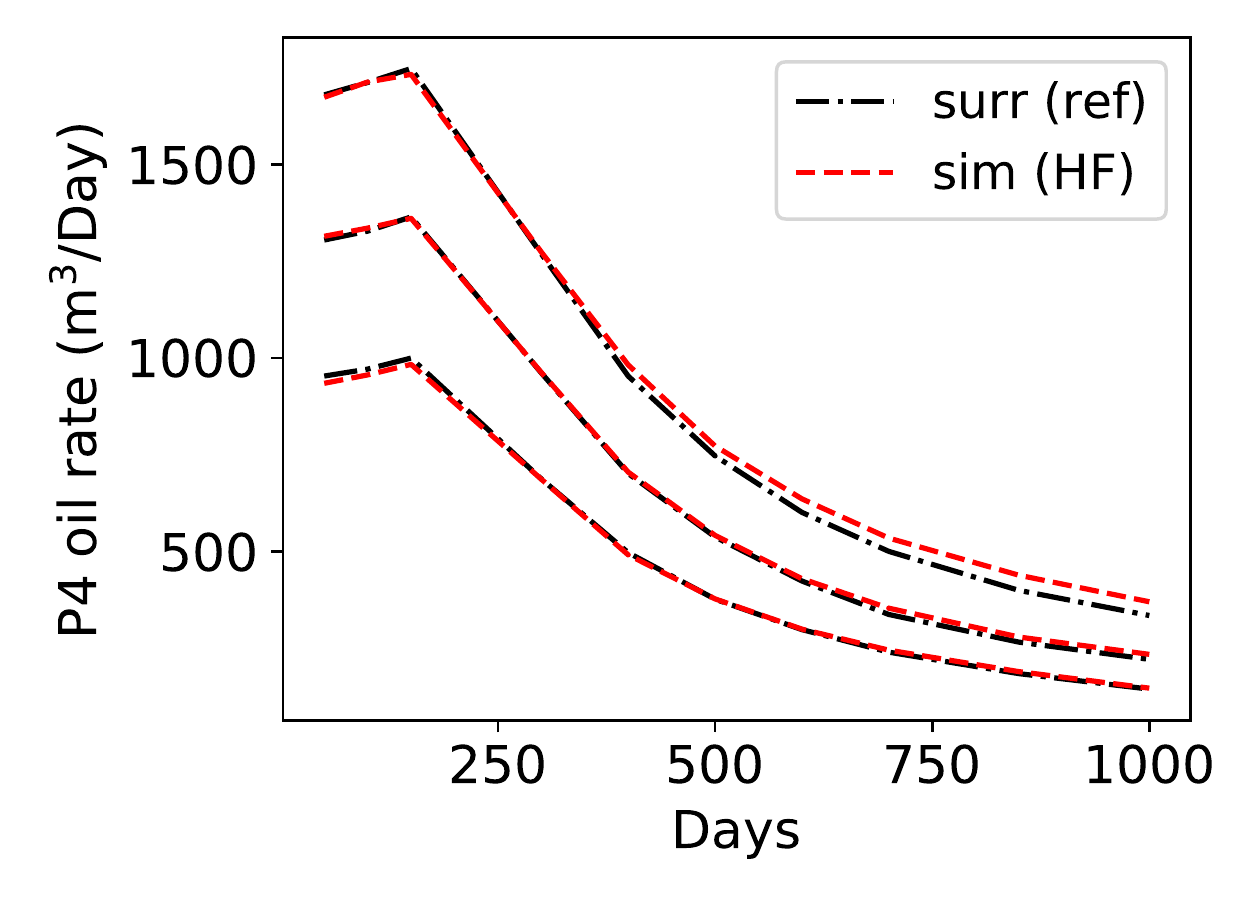}
\subcaption{P4 OPR -- HF and surr (ref)}
\end{minipage}
\begin{minipage}{.32\linewidth}\centering
\includegraphics[trim = 0 0 0 0, clip, width=\linewidth]{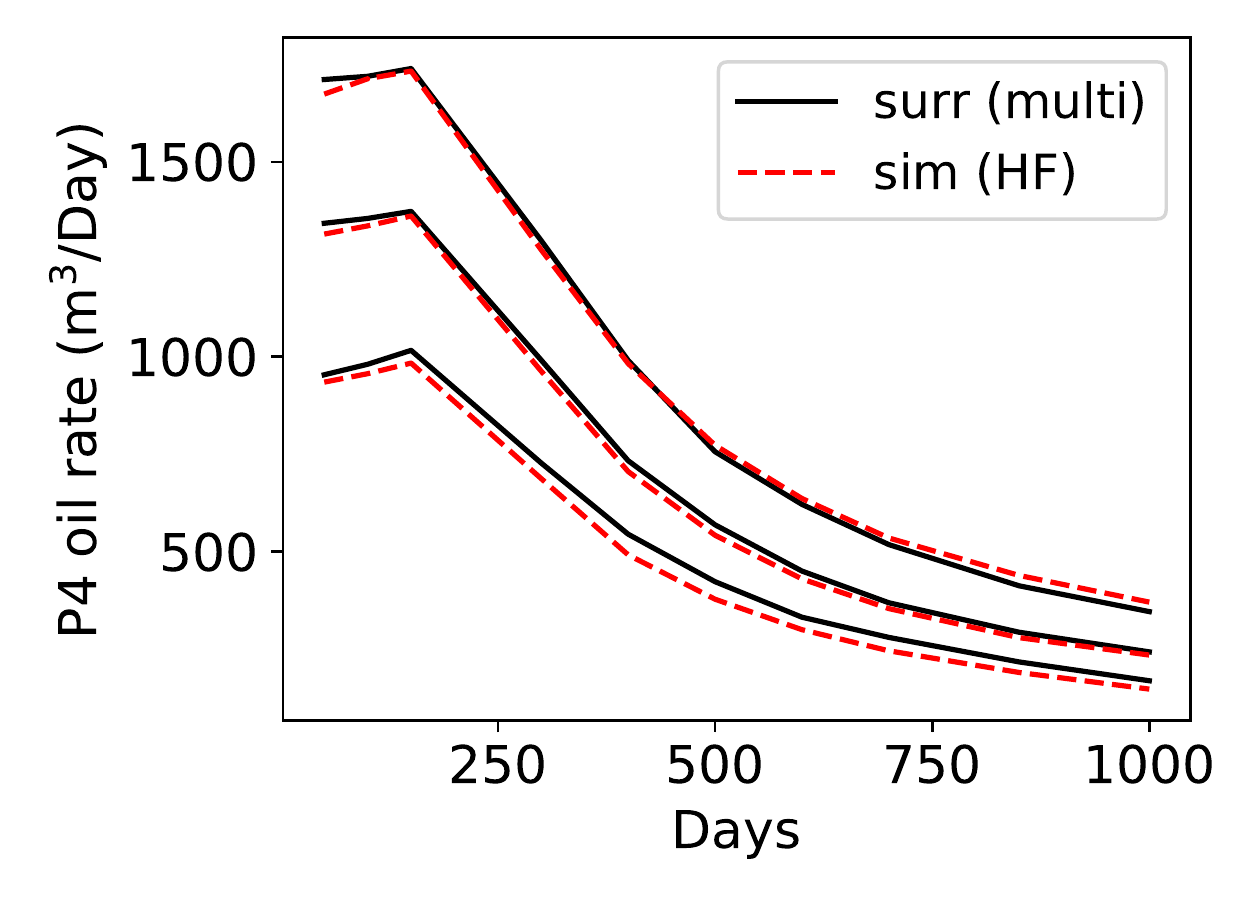}
\subcaption{P4 OPR -- HF and surr (multi)}
\end{minipage}

\begin{minipage}{.32\linewidth}\centering
\includegraphics[trim = 0 0 0 0, clip, width=\linewidth]{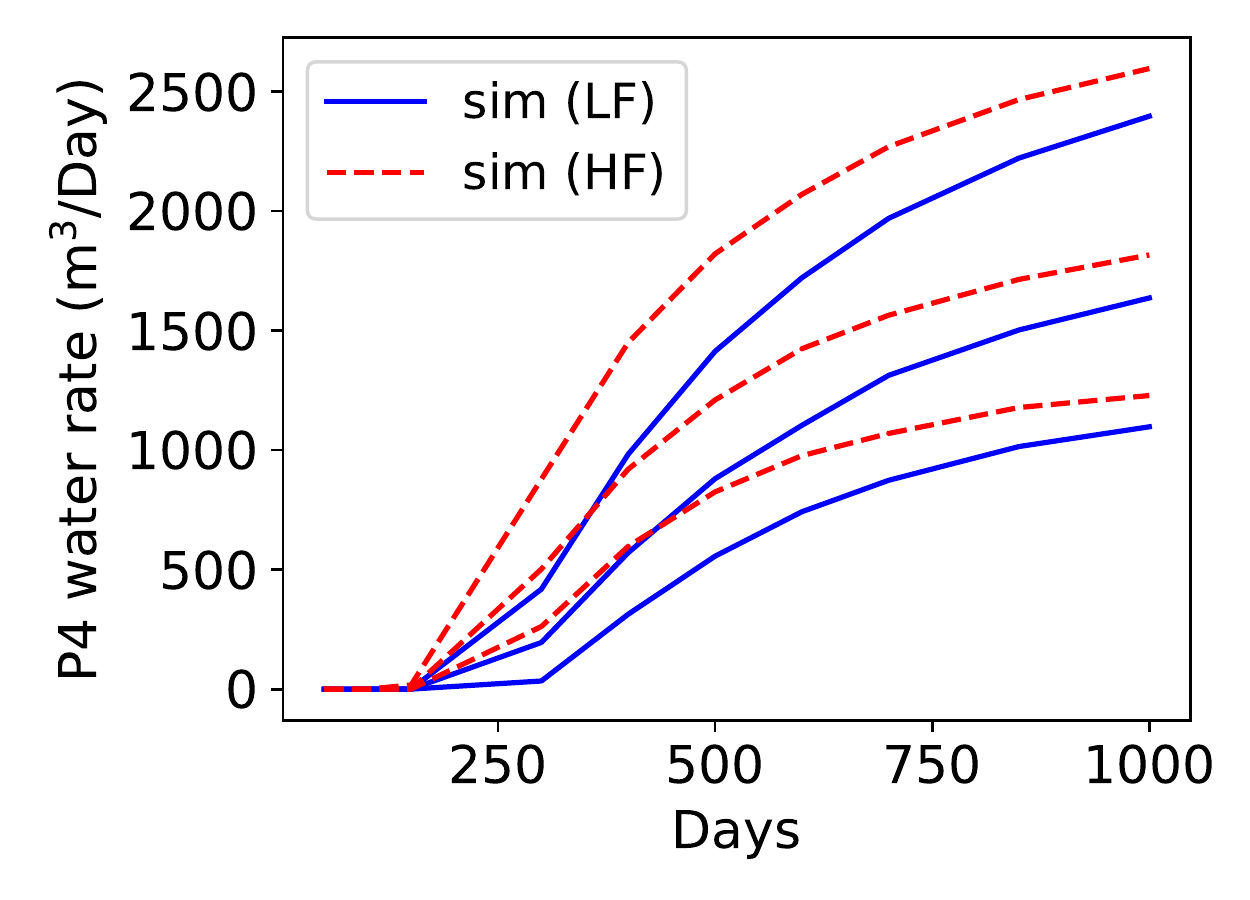}
\subcaption{P4 WPR -- HF and LF}
\end{minipage}
\begin{minipage}{.32\linewidth}\centering
\includegraphics[trim = 0 0 0 0, clip, width=\linewidth]{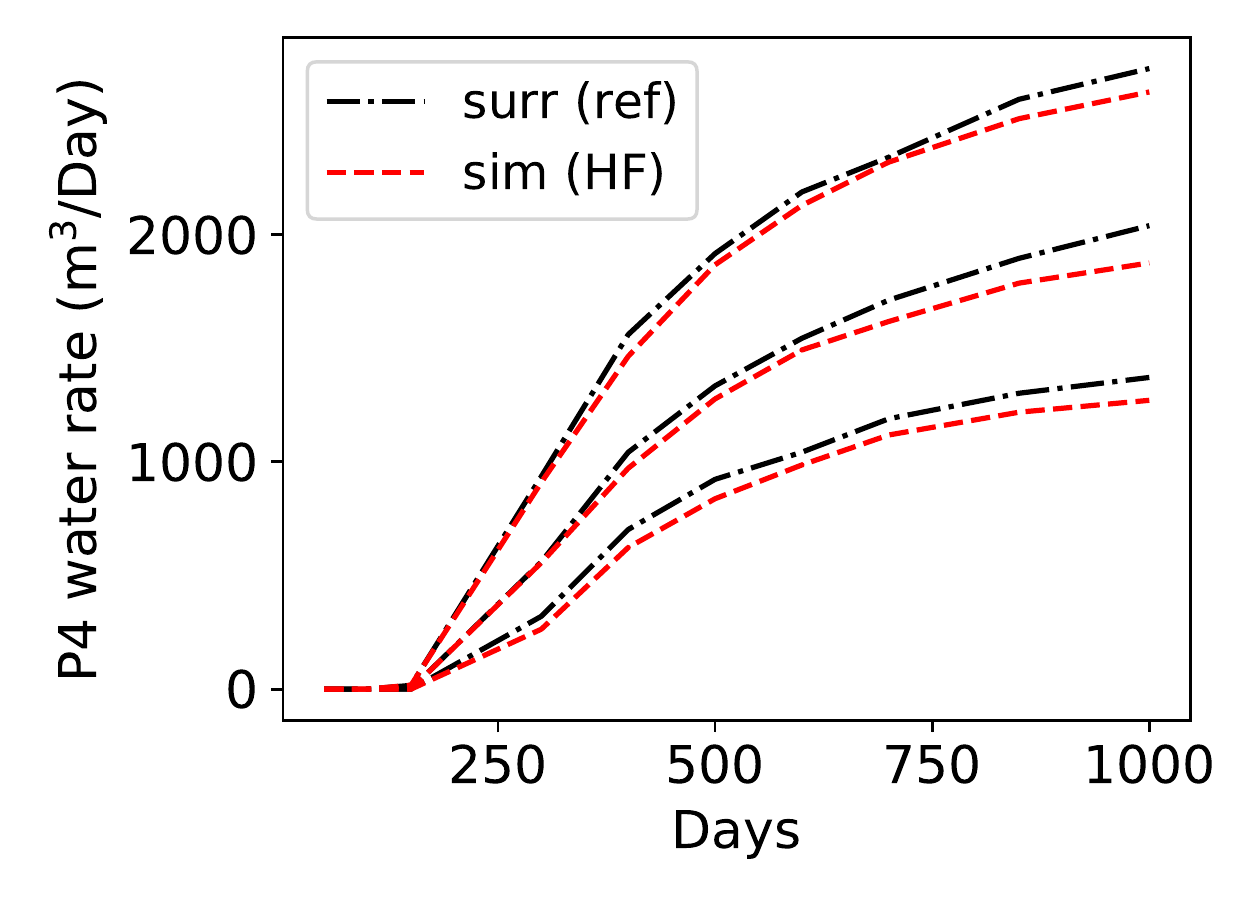}
\subcaption{P4 WPR -- HF and surr (ref)}
\end{minipage}
\begin{minipage}{.32\linewidth}\centering
\includegraphics[trim = 0 0 0 0, clip, width=\linewidth]{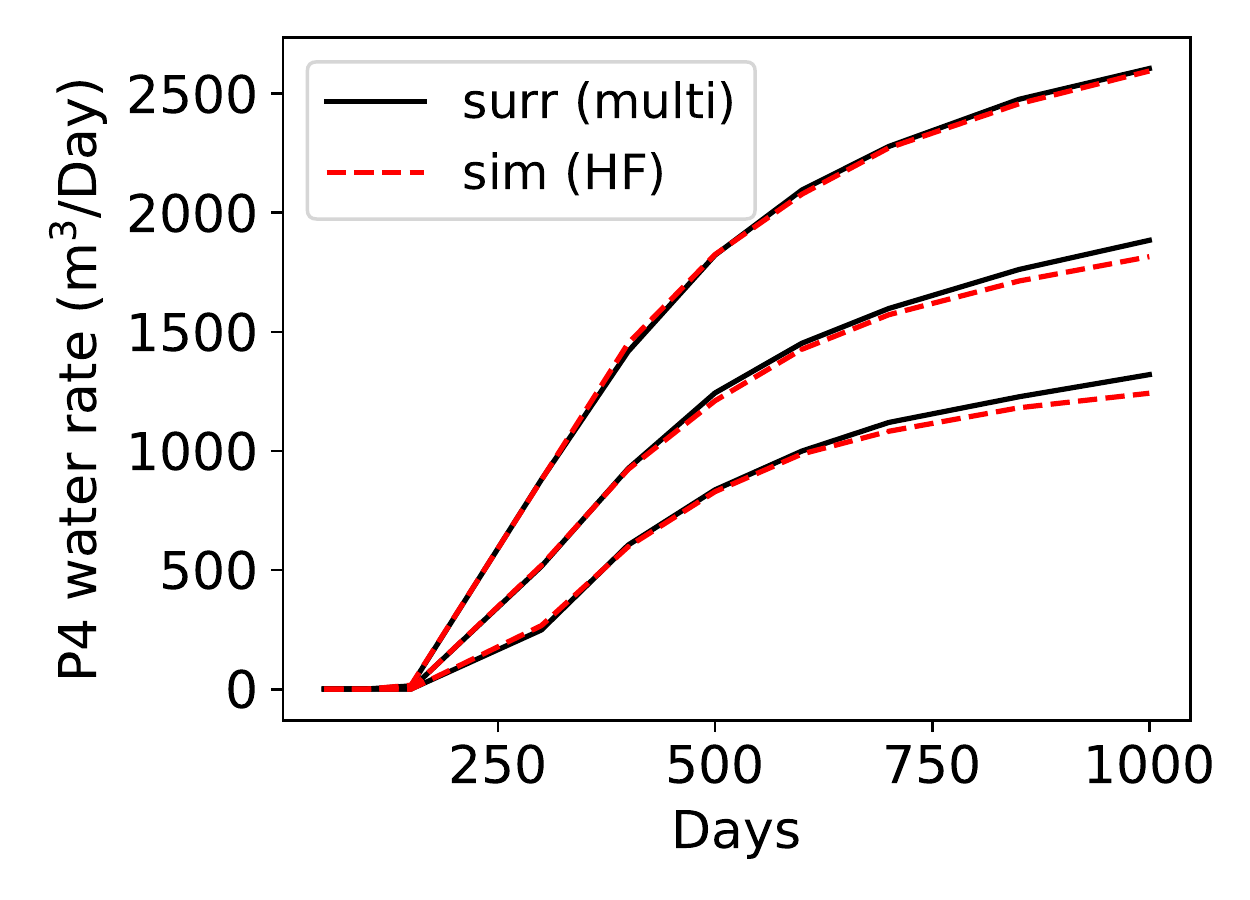}
\subcaption{P4 WPR -- HF and surr (multi)}
\end{minipage}

\caption{Flow rate statistics for HF and LF simulation (left column), reference surrogate model (middle column), and surrogate model trained with multifidelity data (right column) for P1 and P4 oil and water production rates. Results correspond to P$_{10}$, P$_{50}$ and P$_{90}$ responses over 400 test cases.} \label{fig:prior_stats_rate}
\end{figure}

Water injection rates and water and oil production rates are calculated from well-block pressure and saturation values and the specified BHP using Eqs.~\ref{eq:well} and~\ref{eq:wi}. We now present flow rate statistics evaluated over the 400 test samples. Figure~\ref{fig:prior_stats_rate} presents flow statistics for oil and water production rates (OPR and WPR) of producers P1 and P4. Results for the P$_{10}$, P$_{50}$ and P$_{90}$ (10th, 50th and 90th percentile) rates are presented at $n_t = 10$ time steps. The red dashed lines in all subplots in Fig.~\ref{fig:prior_stats_rate} represent the P$_{10}$, P$_{50}$, P$_{90}$ HF simulation results. The blue curves in the left column subplots show LF simulation results, while the black dash-dotted curves in the middle column present the reference surrogate results. The black solid lines in the right column display results for the transfer-learning-based surrogate trained with multifidelity data. 

From the left-column subplots, it is clear that the LF simulation results display some error relative to the HF results. Water production rates are systematically under-predicted in Fig.~\ref{fig:prior_stats_rate}(d) and (j), and oil rates show discrepancies in Fig.~\ref{fig:prior_stats_rate}(a) and (g). The reference surrogate model (trained with HF simulation data) provides accurate flow statistics for all quantities considered. The results for the surrogate trained with multifidelity data (right column) are quite close to the reference surrogate results shown in the middle column. This demonstrates the ability of the transfer learning framework to provide high-quality predictions for key well rate quantities.

\section{Data Assimilation with Surrogate Model} \label{sec:post_surr}

In this section, we first describe the history matching framework, which involves the use of an ensemble smoother with multiple data assimilation (ESMDA)~\cite{emerick2013ensemble}, with geomodels parameterized using CNN-PCA and flow responses determined using the transfer-learning-based surrogate model. We then present history matching results using this framework for a synthetic example.

\subsection{ESMDA Procedure}
The ESMDA-based history matching procedure in this work is the same as in~\cite{tang2021deep, tang2021history}, except here we use the transfer-learning-based surrogate instead of the (reference) recurrent R-U-Net model. CNN-PCA enables us to generate HF geological realizations $\mathbf{m}_{\text{cnnpca}} \in \R^{n^h}$ from low-dimensional latent variables $\bm{\upxi} \in \R^{N_l}$, where $N_l$ denotes the dimension of the latent variables. For the case consider in this work, $N_l = 400$, which is much less than the total number of HF grid blocks ($n^h = 128,000$). A geological realization in the history matching process is represented as $\mathbf{m}_{\text{cnnpca}}(\bm{\upxi})$. The resulting flow predictions $\mathbf{d} \in \R^{N_{\text{hm}}}$ for the historical period, generated using the surrogate model, are denoted by $\mathbf{d} = \hat{f}(\mathbf{m}_{\text{cnnpca}}(\bm{\upxi}))$. Here $N_{\text{hm}}$ denotes the number of observations in the history matching period. 

In the data assimilation procedure, we first sample latent variables from their prior distribution $N(\bm{\upmu}_{\upxi}, C_{\upxi})$, where $\bm{\upmu}_{\upxi}$ and $C_{\upxi}$ denote the prior mean and covariance of $\bm{\upxi}$. ESMDA generates posterior distributions by assimilating data and updating the model parameters multiple times. The ESMDA update equation is 
\begin{equation}
   \bm{\upxi}_i^{k+1} = \bm{\upxi}_i^{k} + C_{\upxi, \text{d}}^k (C_{\text{d}}^k + \alpha_k C_{\text{D}})^{-1}(\mathbf{d}_{\text{obs}} + \sqrt{\alpha_k} \mathbf{e}_i^k - \mathbf{d}_i^k), 
\end{equation}
for $i = 1, \dots, N_r$ and $k = 1, \dots, N_a$, where $N_r$ is the number of realizations considered and $N_a$ is the number of data assimilation steps. Here $\alpha_k$, $k = 1, \dots, N_a$, are the inflation coefficients, subject to $\sum_{k = 1}^{N_a} \alpha_k^{-1} = 1$. The choice of $N_a$ and $\alpha_k$ can affect the performance of the method. In this work, we specify $N_a = 10$ and use $\alpha_k$ values of 57.017, 35.0, 25.0, 20.0, 18.0, 15.0, 12.0, 8.0, 5.0, 3.0, as suggested in~\cite{emerick2013investigation}. 

At each iteration $k$, we perturb the observation data $\mathbf{d}_{\text{obs}} \in \R^{N_{\text{hm}}}$ with $\mathbf{d}_{\text{obs}} + \sqrt{\alpha_k} \mathbf{e}_i^k$. The vector $\mathbf{e}$ represents random noise sampled from $N(\mathbf{0}, C_{\text{D}})$, where $C_{\text{D}}$ is the covariance of the measurement error. The data in the historical period are generated through application of $\mathbf{d}_i^k = \hat{f}(\mathbf{m}_{\text{cnnpca}}(\bm{\upxi}_i^k))$. The covariance matrix $C_{\upxi, \text{d}}^k$ and auto-covariance $C_{\text{d}}^k$ are constructed from the data ensemble $\mathbf{d}^k_i$ and model parameter ensemble $\bm{\upxi}^k_i$. After $N_a$ iterations are completed, the posterior predictions $\mathbf{d}_{\text{post}}$ are constructed via  $\mathbf{d}_{\text{post}} = \hat{g}(\mathbf{m}_{\text{cnnpca}}(\bm{\upxi}_{\text{post}}))$.

\subsection{History Matching Results}
We now present history matching results using the recurrent R-U-Net surrogate model trained with transfer learning and multifidelity data. These predictions are compared to results using the reference surrogate model for the determination of $\mathbf{d}_i^k = \hat{f}(\mathbf{m}_{\text{cnnpca}}(\bm{\upxi}_i^k))$. The geomodel shown in Fig.~\ref{fig:perm}(c) is taken to be the `true' model. The observed data $\mathbf{d}_{\text{obs}}$ are generated by performing high-fidelity simulation of this model, with noise added to represent measurement error. This noise is sampled from $N(\mathbf{0}, C_{\text{D}})$, with the standard deviation specified to be 5\% of the simulated value. The observed data include the water and oil production rates for all five production wells at 150 and 300~days. We thus have $N_{\text{hm}}=20$. We set $N_r=400$.

\begin{figure}[!hbt]
\centering
\begin{minipage}{.45\linewidth}\centering
\includegraphics[trim = 0 0 0 0, clip, width=\linewidth]{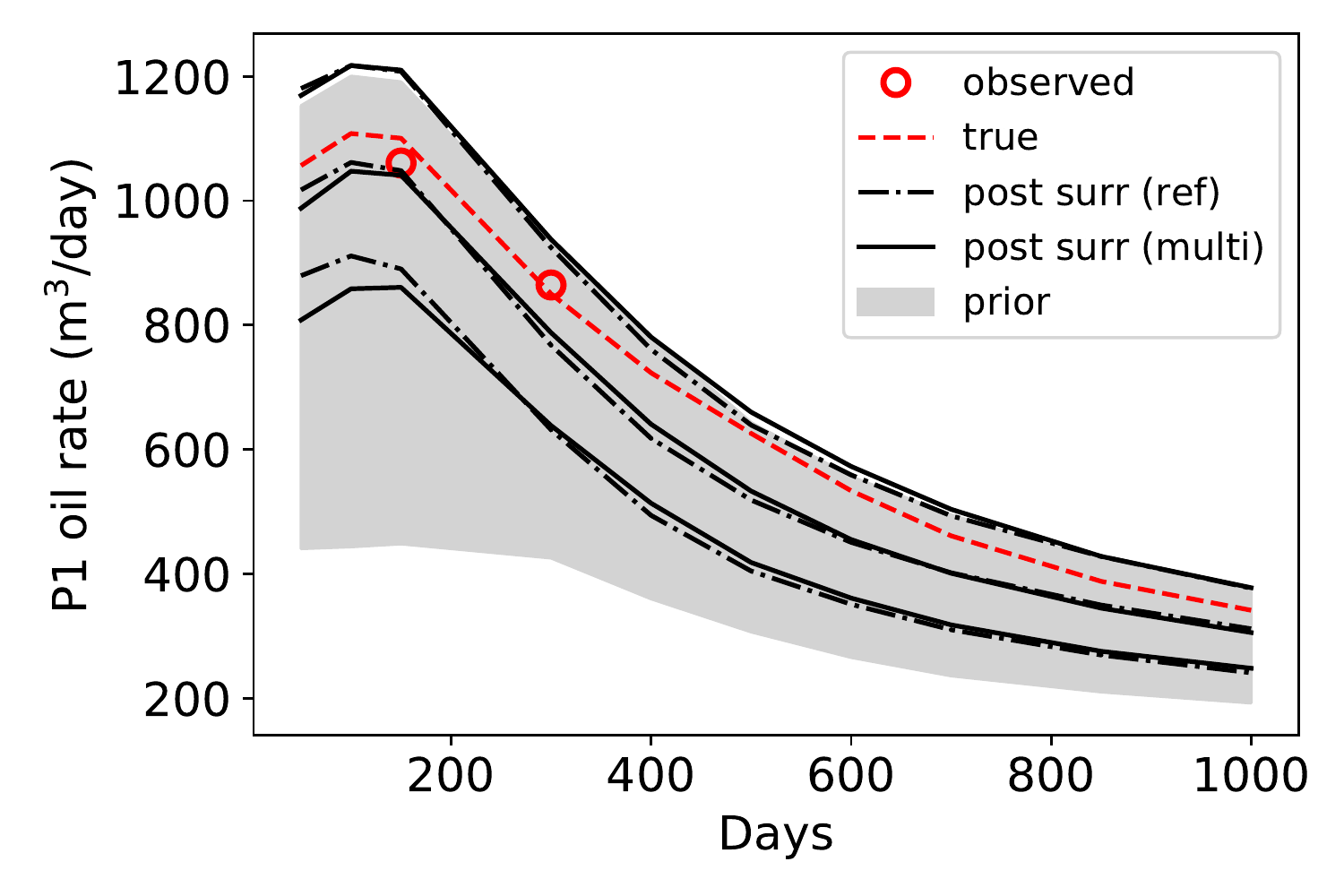}
\subcaption{P1 oil rate}
\end{minipage}
\begin{minipage}{.45\linewidth}\centering
\includegraphics[trim = 0 0 0 0, clip, width=\linewidth]{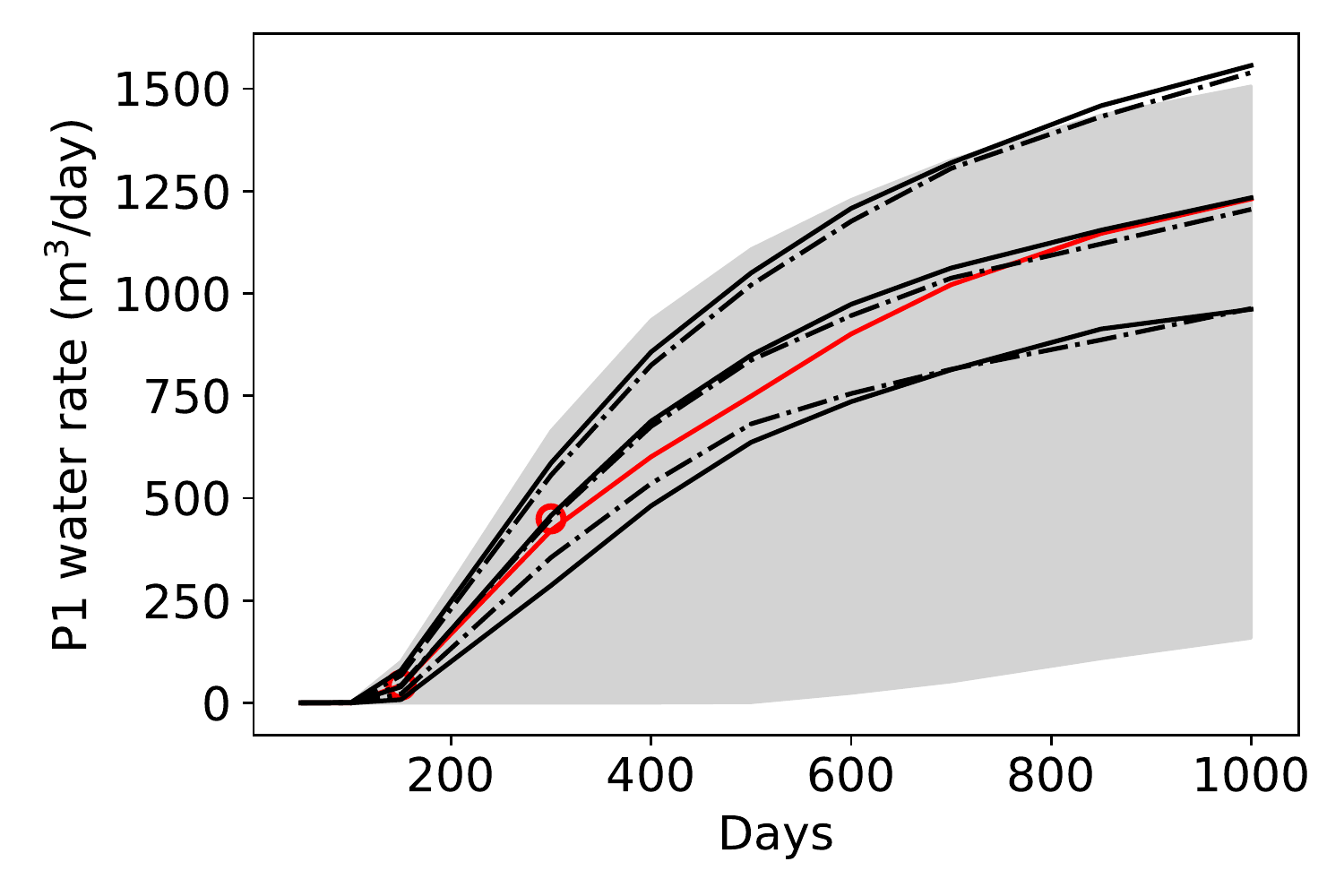}
\subcaption{P1 water rate}
\end{minipage}

\begin{minipage}{.45\linewidth}\centering
\includegraphics[trim = 0 0 0 0, clip, width=\linewidth]{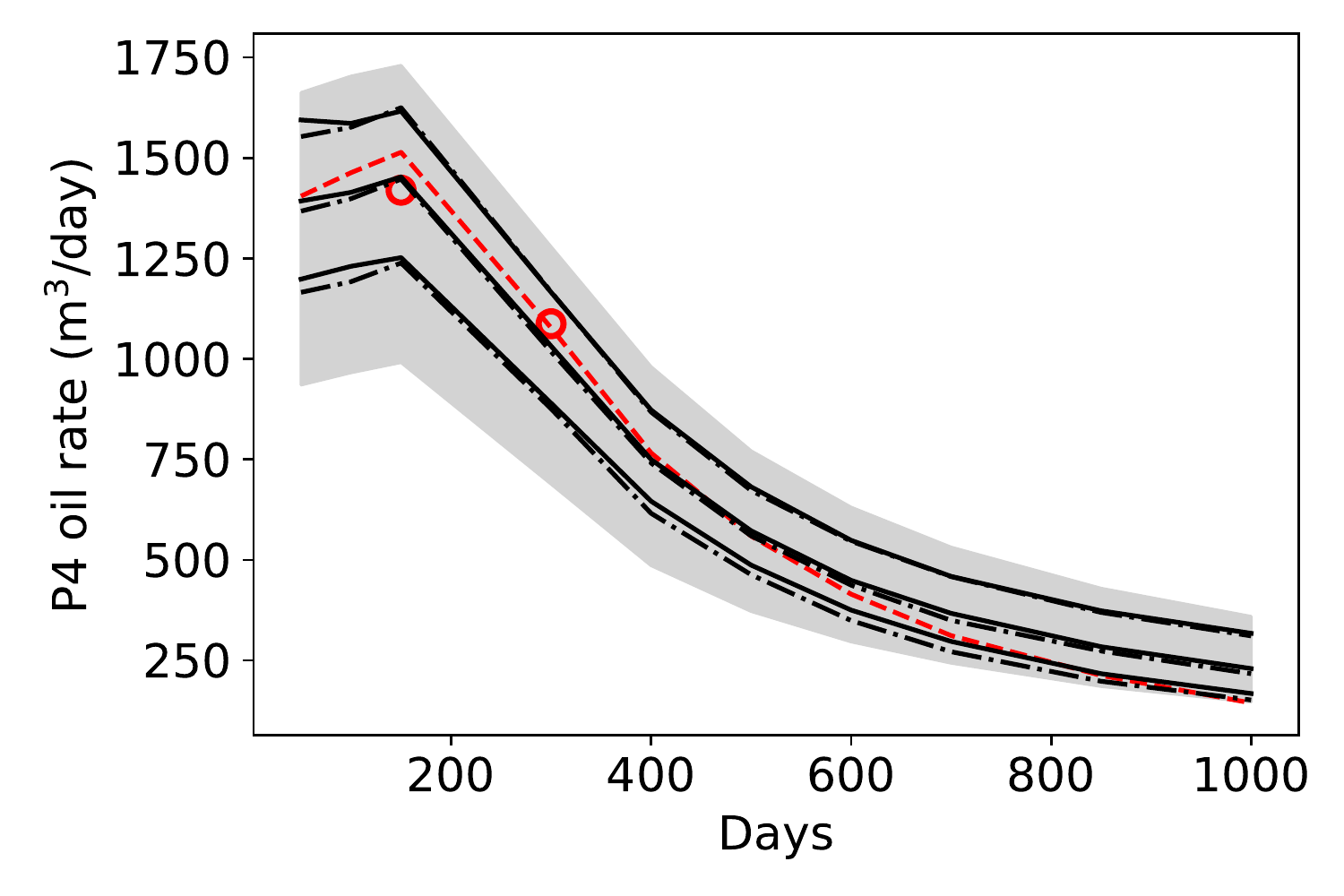}
\subcaption{P4 oil rate}
\end{minipage}
\begin{minipage}{.45\linewidth}\centering
\includegraphics[trim = 0 0 0 0, clip, width=\linewidth]{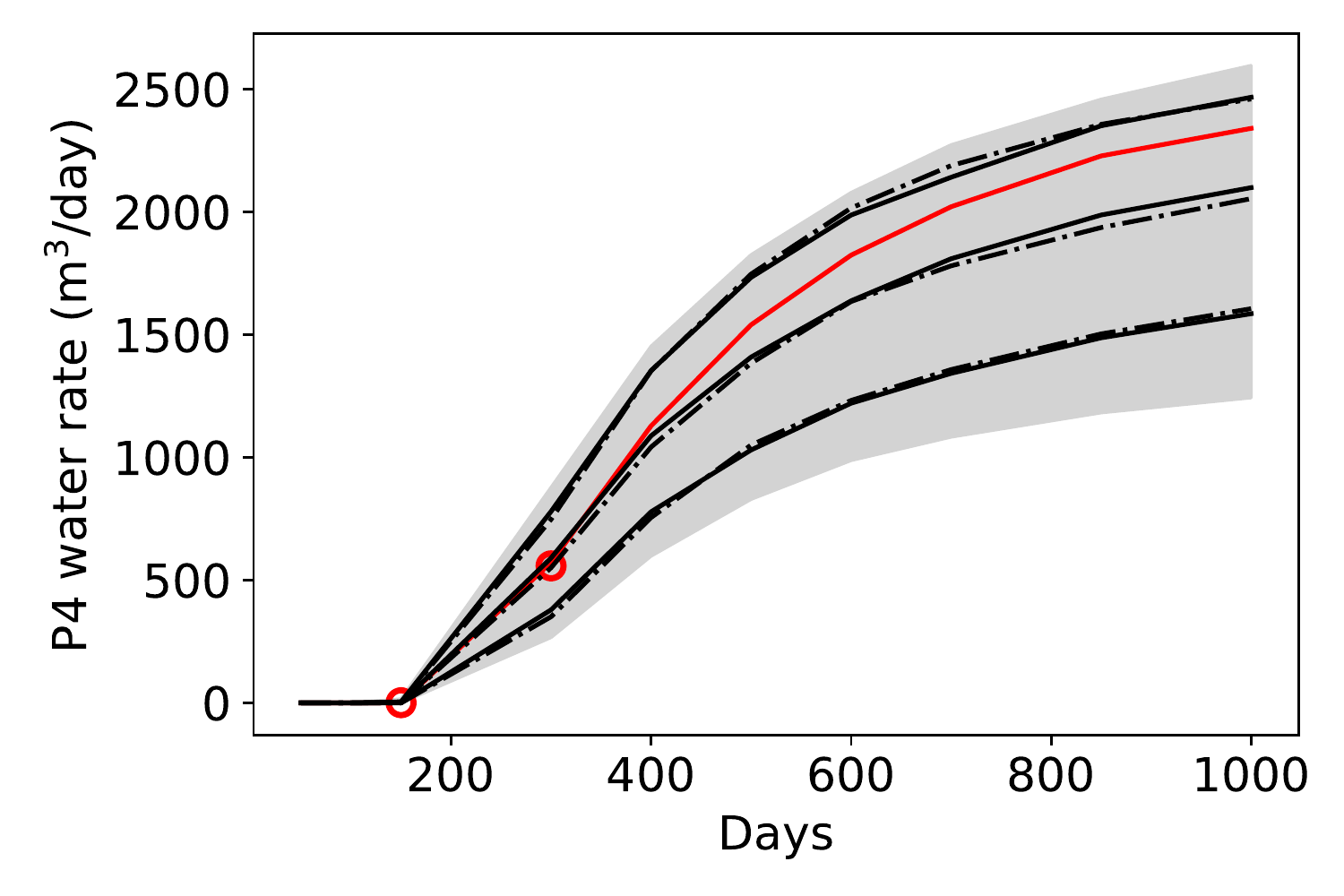}
\subcaption{P4 water rate}
\end{minipage}

\begin{minipage}{.45\linewidth}\centering
\includegraphics[trim = 0 0 0 0, clip, width=\linewidth]{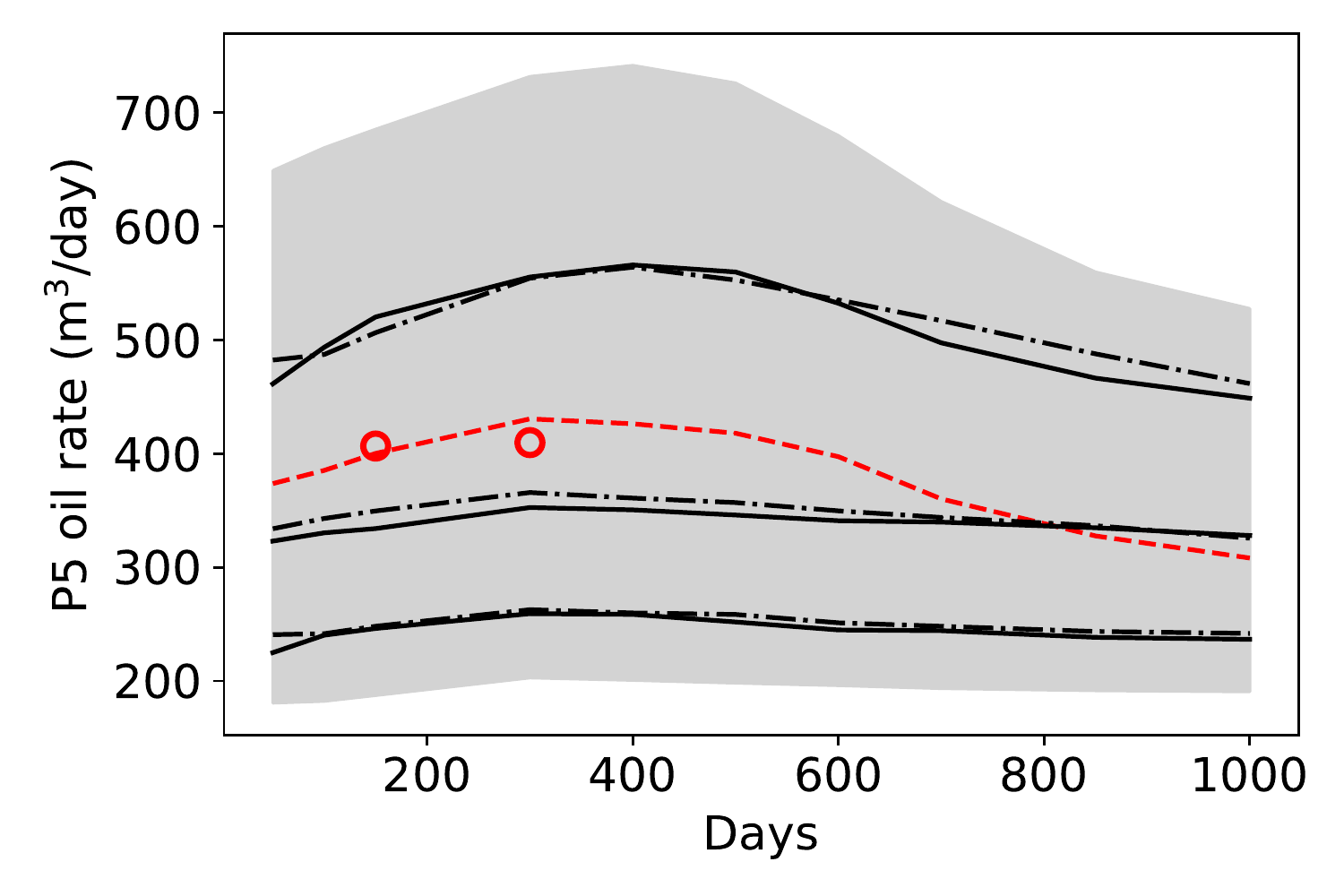}
\subcaption{P5 oil rate}
\end{minipage}
\begin{minipage}{.45\linewidth}\centering
\includegraphics[trim = 0 0 0 0, clip, width=\linewidth]{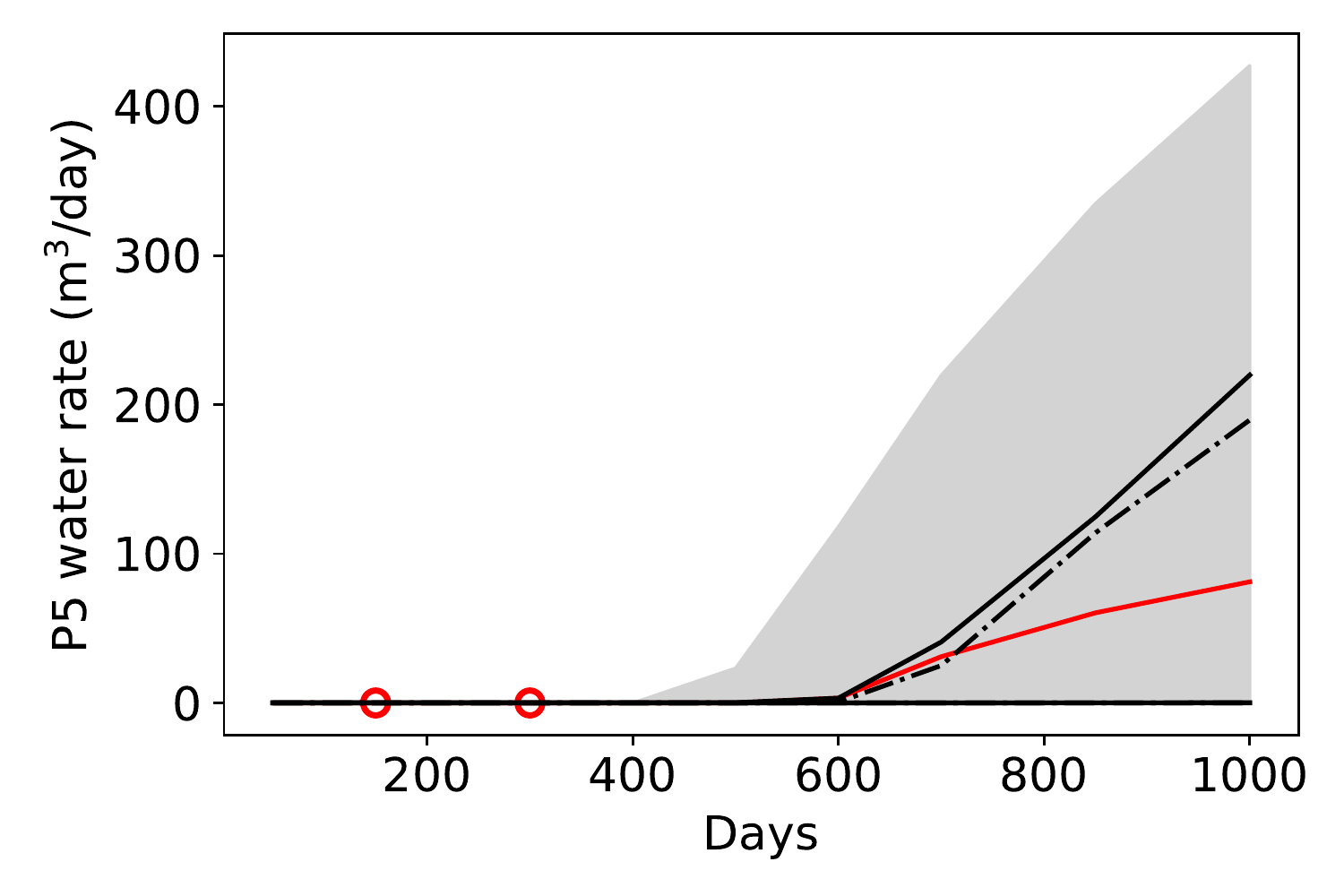}
\subcaption{P5 water rate}
\end{minipage}

\caption{Posterior results from reference surrogate model (black dash-dotted curves) and surrogate model trained with multifidelity data (black solid curves). Lower, middle and upper curves are P$_{10}$, P$_{50}$ and P$_{90}$ responses. Gray regions show the prior P$_{10}$–-P$_{90}$ range, and red circles and curves present observed and true data. Legend in (a) applies to all subplots.}\label{fig:post_prior_flow}
\end{figure}

Figure~\ref{fig:post_prior_flow} presents statistical results (over the 400 posterior samples) for water and oil production rates of wells P1, P4, and P5. These wells are representative of the overall system (the P2 flow rates resemble those of P1, and the P3 rates resemble those of P4). The prior P$_{10}$--P$_{90}$ range for the reference high-fidelity simulation results is shown in each subplot as the gray-shaded region. The red dashed lines show the simulated flow rates of the `true' model. The red circles present the observed data (which deviate from the red lines due to measurement error). The black dash-dotted lines indicate the P$_{10}$, P$_{50}$, P$_{90}$ posterior results generated from the reference surrogate model and ESMDA. The black solid lines show the P$_{10}$, P$_{50}$, P$_{90}$ posterior predictions from the surrogate model trained with multifidelity data. 

In Fig.~\ref{fig:post_prior_flow}, we observe substantial uncertainty reduction in posterior flow rate predictions for some quantities (such as P1 water rate) and very little uncertainty reduction in other quantities (P4 water rate). The true-model well rates (red curves) are essentially within the posterior P$_{10}$--P$_{90}$ ranges, as would be expected. Most importantly for current purposes, the posterior P$_{10}$, P$_{50}$, P$_{90}$ results from the surrogate model trained with multifidelity data agree closely with the reference surrogate model predictions. There are some minor discrepancies between these two sets of results (e.g., at early time in the P$_{10}$ prediction in Fig.~\ref{fig:post_prior_flow}(a)), but the differences in general are small, especially compared to the amount of uncertainty reduction achieved.

\begin{figure}[!hbt]
\centering
\begin{minipage}{.45\linewidth}\centering
\includegraphics[trim = 0 0 0 0, clip, width=\linewidth]{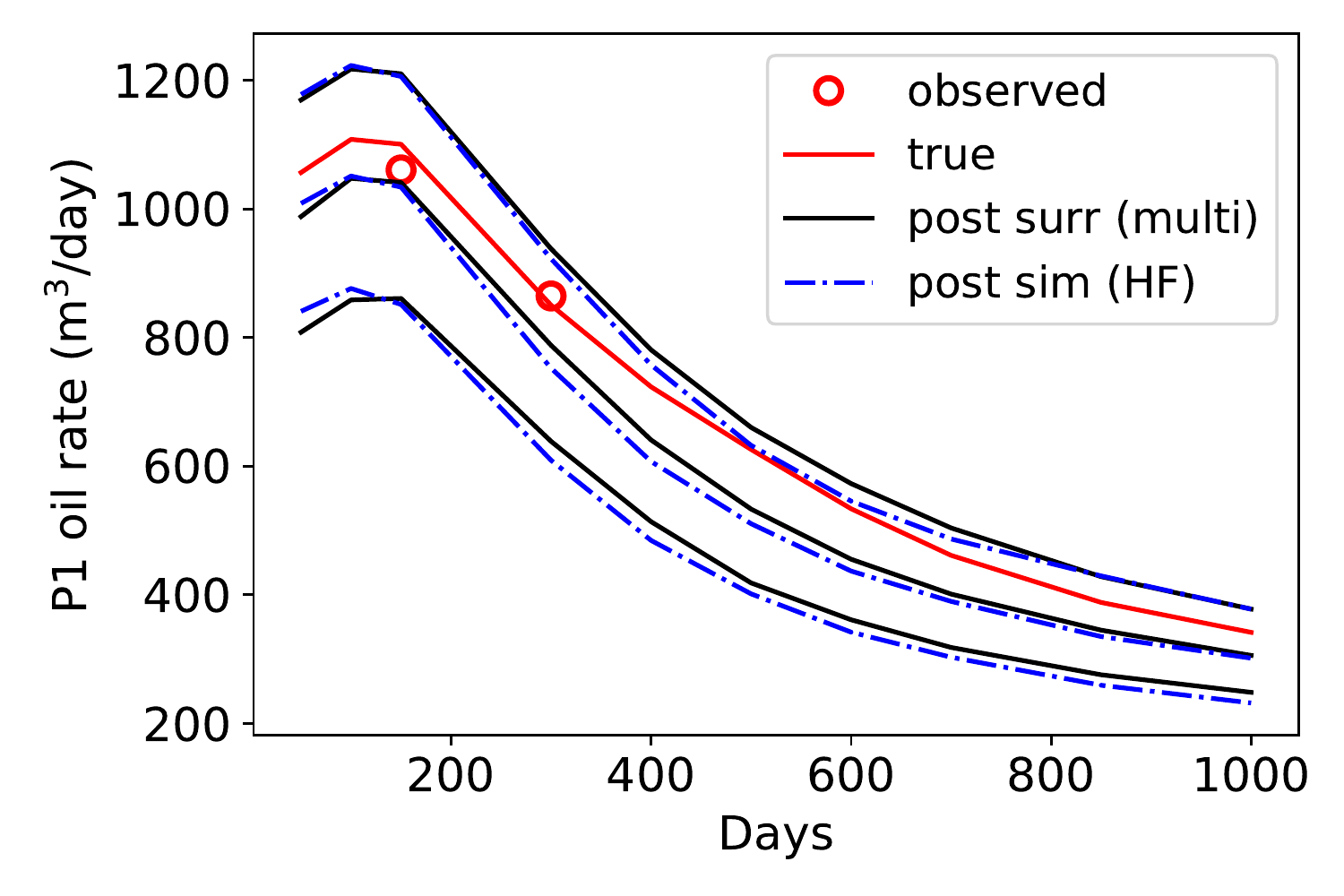}
\subcaption{P1 oil rate}
\end{minipage}
\begin{minipage}{.45\linewidth}\centering
\includegraphics[trim = 0 0 0 0, clip, width=\linewidth]{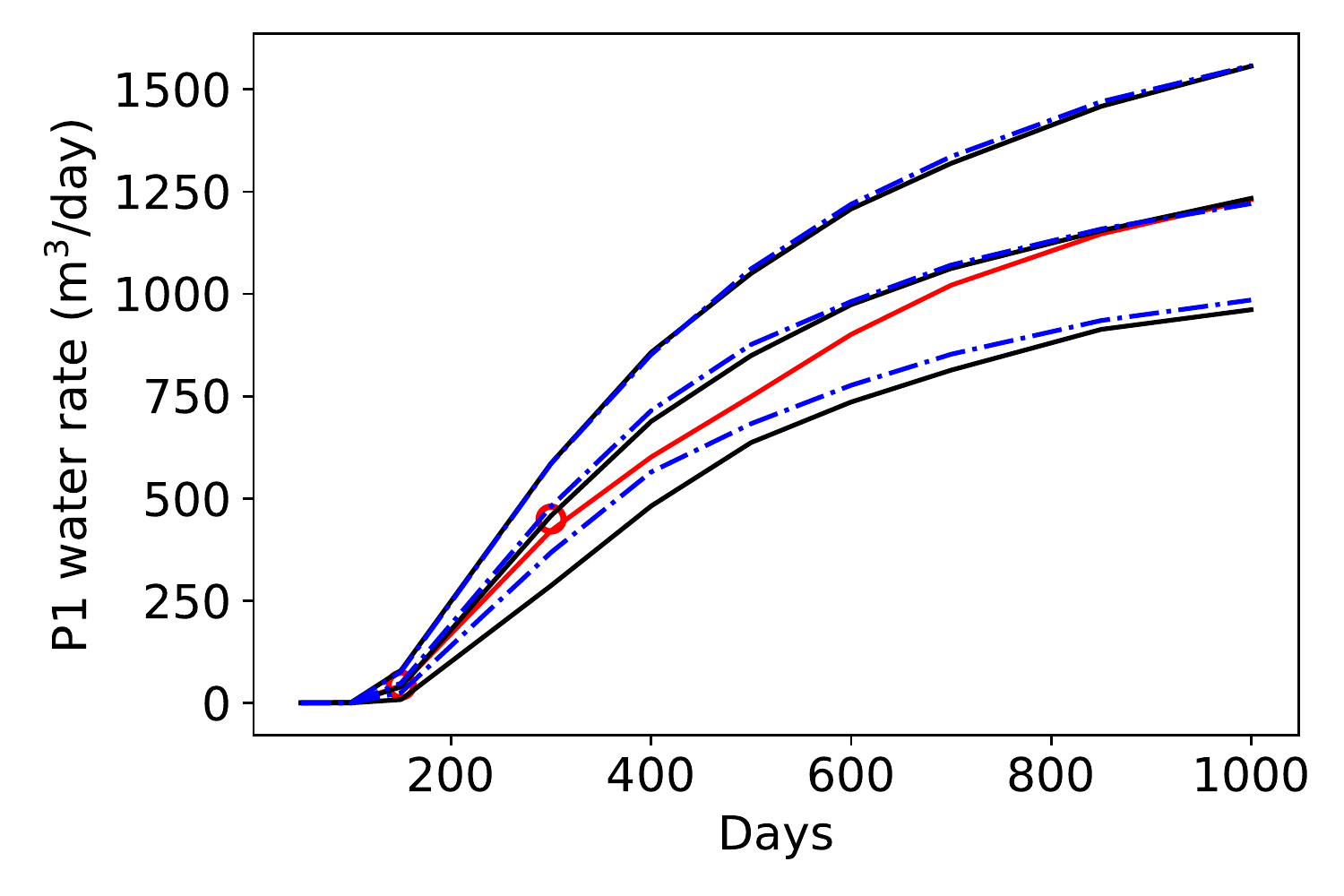}
\subcaption{P1 water rate}
\end{minipage}

\begin{minipage}{.45\linewidth}\centering
\includegraphics[trim = 0 0 0 0, clip, width=\linewidth]{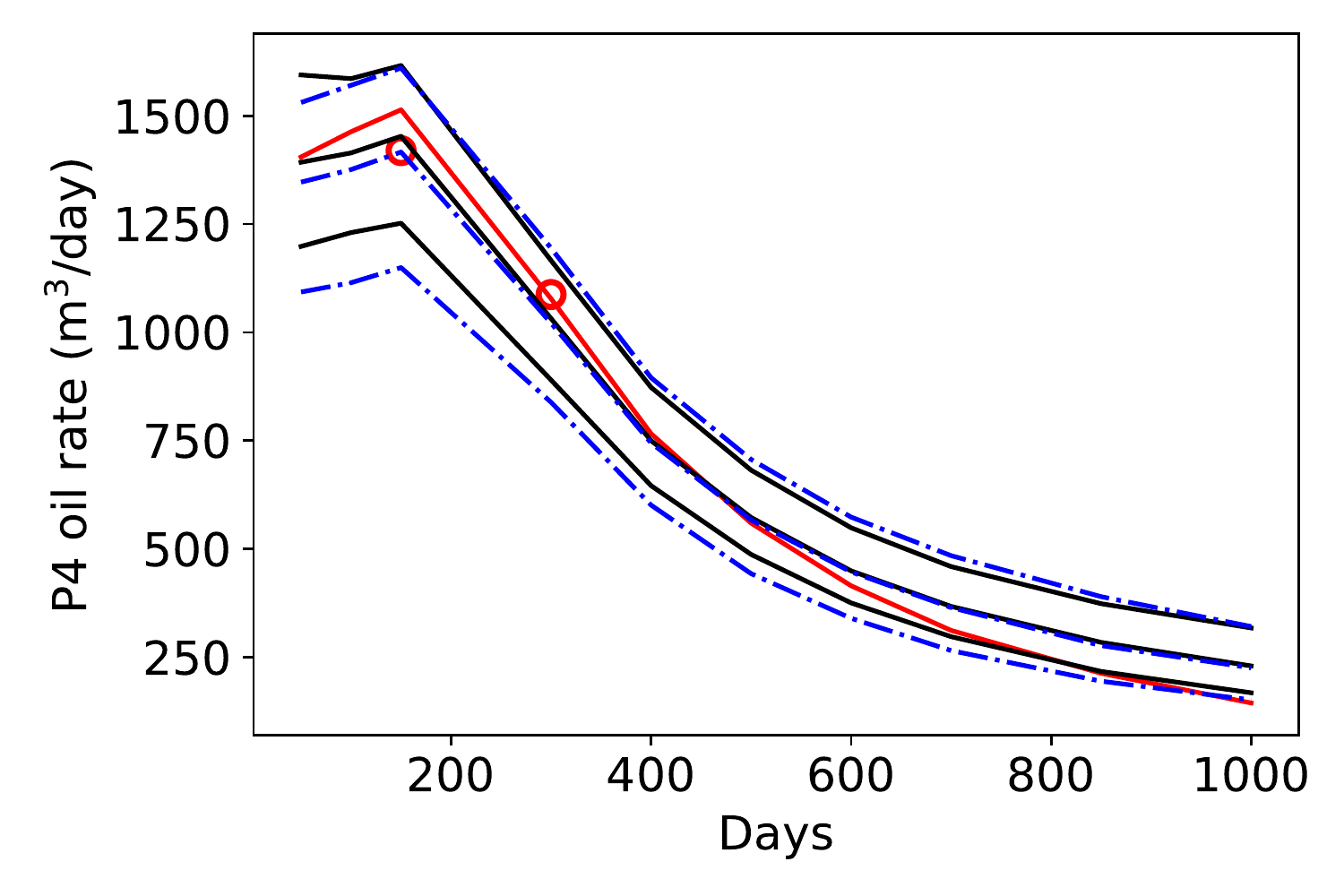}
\subcaption{P4 oil rate}
\end{minipage}
\begin{minipage}{.45\linewidth}\centering
\includegraphics[trim = 0 0 0 0, clip, width=\linewidth]{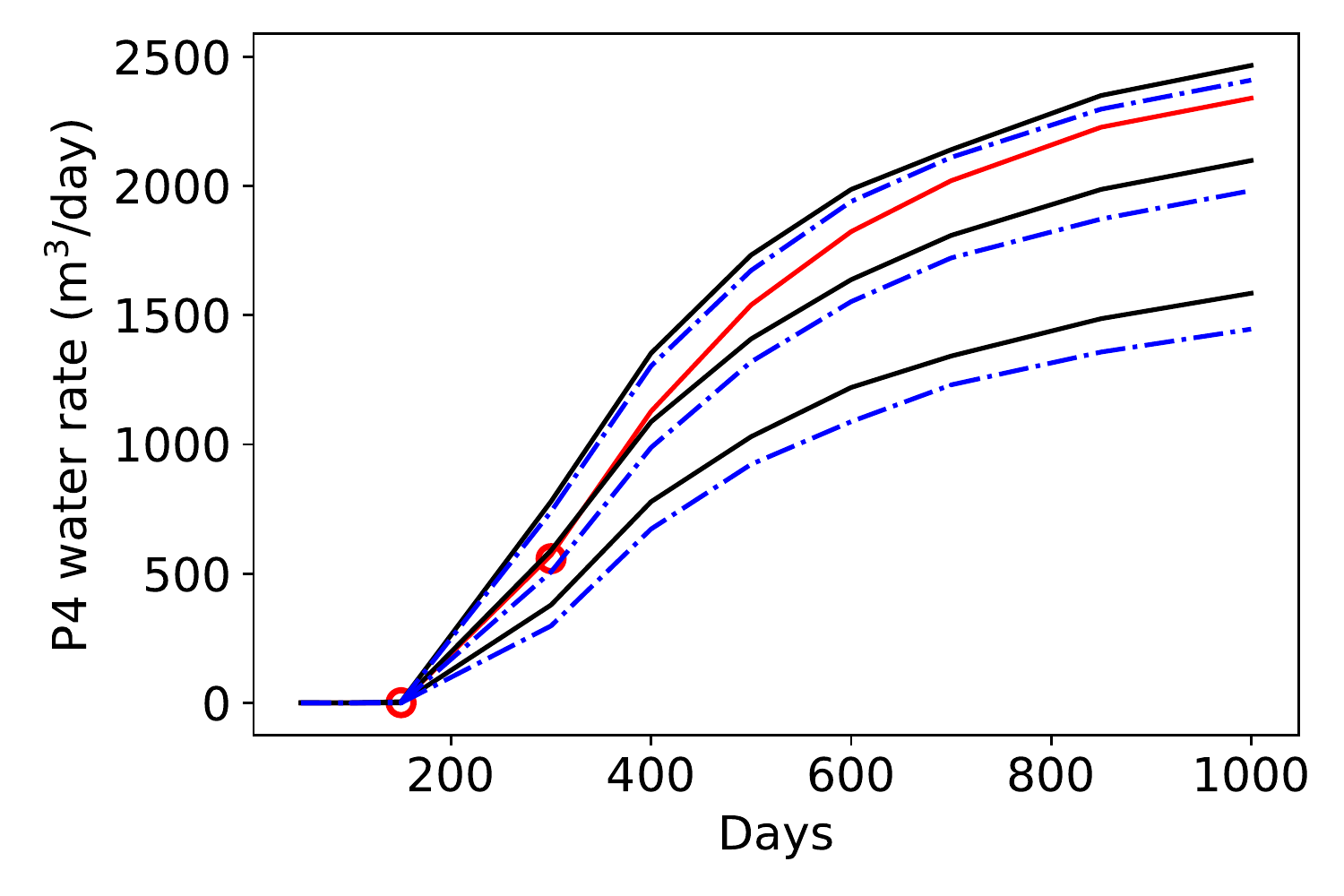}
\subcaption{P4 water rate}
\end{minipage}

\begin{minipage}{.45\linewidth}\centering
\includegraphics[trim = 0 0 0 0, clip, width=\linewidth]{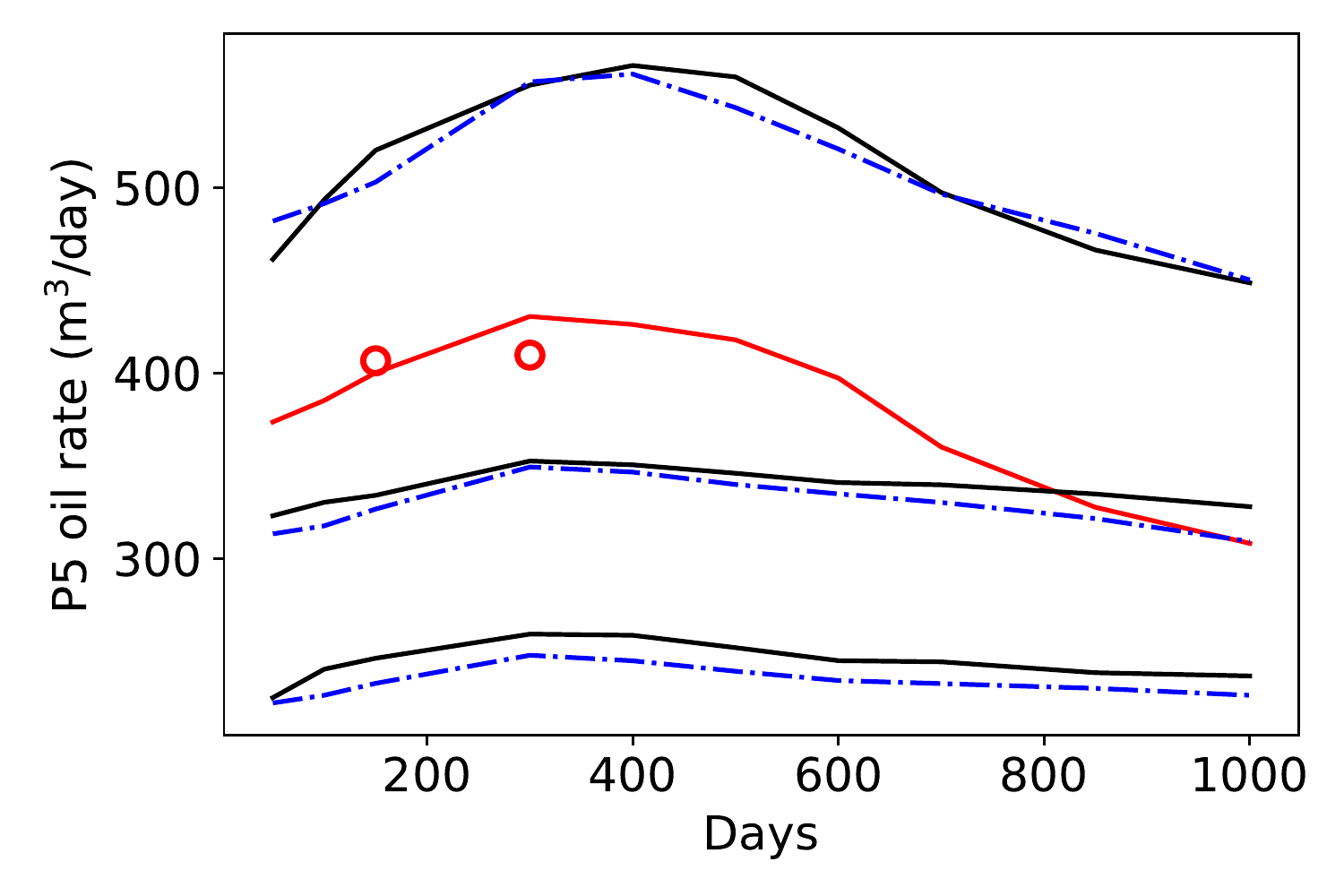}
\subcaption{P5 oil rate}
\end{minipage}
\begin{minipage}{.4\linewidth}\centering
\includegraphics[trim = 0 0 0 0, clip, width=\linewidth]{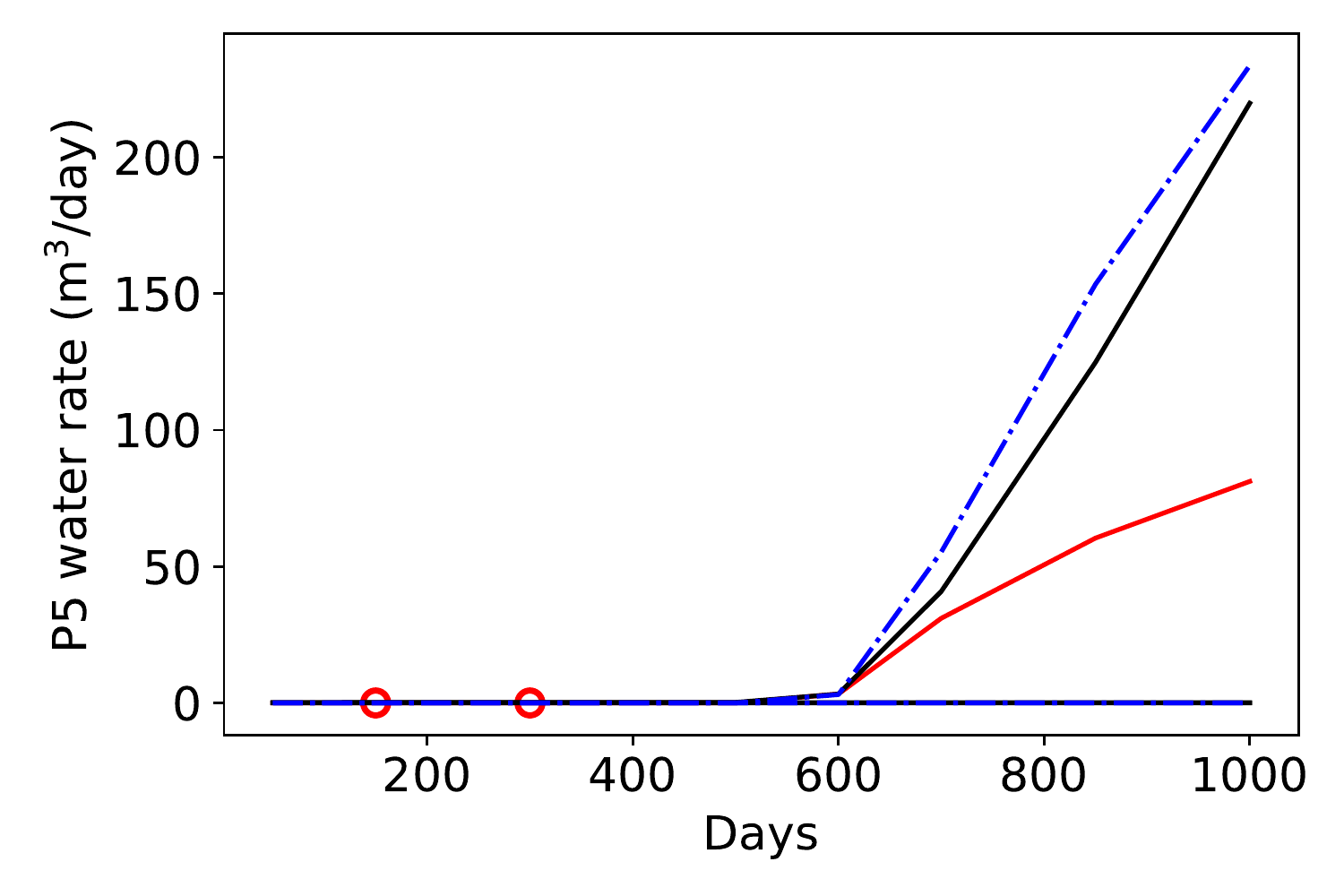}
\subcaption{P5 water rate}
\end{minipage}

\caption{Posterior results from surrogate model trained with multifidelity data (black solid curves) and corresponding fine-scale simulations (blue dash-dotted curves). Lower, middle and upper curves are P$_{10}$, P$_{50}$ and P$_{90}$ responses. Legend in (a) applies to all subplots.}\label{fig:post_flow}
\end{figure}

Flow simulation is now performed on the 400 posterior models found using ESMDA with the transfer-learning-based surrogate model. These simulation results are compared to surrogate model predictions, in terms of P$_{10}$, P$_{50}$, P$_{90}$ flow responses, in Fig.~\ref{fig:post_flow}. Note that prior results are not shown here, and the $y$-axis ranges differ from those in Fig.~\ref{fig:post_prior_flow}. There is slightly more error here than in Fig.~\ref{fig:post_prior_flow}, though the overall agreement is still quite satisfactory. 

\begin{figure}[!htb]
\centering
\includegraphics[width = 0.5\textwidth]{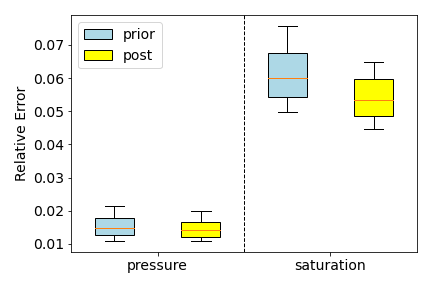}
\caption{Relative errors for prior and posterior pressure and saturation predictions for surrogate model trained with multifidelity data. Errors are for 400 realizations and are relative to HF simulation results.}\label{fig:error_post}
\end{figure}

Finally, in Fig.~\ref{fig:error_post}, we show relative errors for pressure and saturation predictions for prior and posterior transfer-learning-based surrogate results. These results are of interest because they quantify global accuracy, in contrast to accuracy in (local) well rate quantities, as shown in Fig.~\ref{fig:post_flow}. The blue boxes in Fig.~\ref{fig:error_post} display the relative errors (compared to HF simulation) for pressure and saturation predictions for 400 prior samples, while the yellow boxes show these errors for the posterior results (box quantities are as described for Fig.~\ref{fig:err_multi_p}). The errors in the prior models correspond to the histogram results in Fig.~\ref{fig:error_p_sat}. The consistency between the prior and posterior errors indicates that the surrogate model retains accuracy during the data assimilation process.

\clearpage

\section{Concluding Remarks} \label{sec:conclusion}
In this work, a transfer-learning-based surrogate model that uses multifidelity training data was developed to approximate state variables and well rates in two-phase subsurface flow. The underlying network for this implementation is an existing 3D recurrent residual U-Net, which has been used previously for oil-water and CO$_2$ storage problems. The surrogate model is trained with both high and low-fidelity numerical simulation data. Most of the network parameters are first estimated from LF data, while output-layer training and network fine-tuning are accomplished with HF data. In the examples, we used 2500 LF simulations and 200 HF simulations. The network input in all cases is the high-fidelity (fine-scale) geomodel. The LF simulations are themselves performed using coarsened geomodels constructed from the original fine-scale geological description using a global flow-based transmissibility upscaling procedure. The multifidelity approach provided about a 90\% savings in training computations and a 40\% savings in (GPU) training time compared to using only HF simulation data.

The transfer-learning-based surrogate model was applied for two-phase oil-water flow in 3D channelized systems, with eight wells operating under bottom-hole pressure control. The surrogate model was evaluated over a test set of 400 HF models. A reference surrogate model, in which all training runs were performed at high fidelity (the same 2500 data samples were used), was also constructed. The transfer-learning-based surrogate model was shown to provide accurate predictions for dynamic pressure and saturation fields, with errors very near those obtained by the reference surrogate model. Specifically, the median relative errors for pressure and saturation with the transfer-learning-based surrogate were 1.5\% and 5.5\%, compared to 1.1\% and 3.5\% for the reference surrogate. This represents much greater accuracy than that achieved in the LF numerical simulations, where the median saturation error was 19.4\%. Close agreement in P$_{10}$, P$_{50}$ and P$_{90}$ well rates between the transfer-learning-based surrogate predictions and HF simulations was also demonstrated. 

The surrogate model was then combined with ESMDA for use in history matching. The geological models were parameterized using the 3D CNN-PCA representation, and flow predictions were generated using the transfer-learning-based surrogate model. The overall procedure provided clear uncertainty reduction, and comparisons of flow predictions for posterior models demonstrated reasonable agreement between surrogate and numerical simulation results for P$_{10}$, P$_{50}$ and P$_{90}$ well rate quantities.

There are a number of topics for future research in this area. In this work, we used simulation data from only two fidelity levels. It may be possible to improve the performance of the surrogate model, and/or reduce computational demands, through use of data at several fidelity levels. The transfer-learning-based framework can be extended to treat larger and more complicated models. This includes coupled flow and geomechanics, which is important in CO$_2$ storage settings. It will also be of interest to incorporate physical loss in the training process. This could act to further reduce the number of high-fidelity training simulations required. Finally, the general framework could be extended to treat problems involving variable well controls and well locations, which would enable its use for a wide range of optimization problems.


\section*{Acknowledgements}
\noindent We are grateful to the Stanford Smart Fields Consortium for partial funding of this work. We thank Meng Tang for providing geological models, parameterization code and recurrent R-U-Net code, and Dylan Crain for providing the upscaling code used in this study.

\bibliographystyle{elsarticle-num-names} 
\bibliography{reference}

\begin{thebibliography}{31}
\expandafter\ifx\csname natexlab\endcsname\relax\def\natexlab#1{#1}\fi
\providecommand{\url}[1]{\texttt{#1}}
\providecommand{\href}[2]{#2}
\providecommand{\path}[1]{#1}
\providecommand{\DOIprefix}{doi:}
\providecommand{\ArXivprefix}{arXiv:}
\providecommand{\URLprefix}{URL: }
\providecommand{\Pubmedprefix}{pmid:}
\providecommand{\doi}[1]{\href{http://dx.doi.org/#1}{\path{#1}}}
\providecommand{\Pubmed}[1]{\href{pmid:#1}{\path{#1}}}
\providecommand{\bibinfo}[2]{#2}
\ifx\xfnm\relax \def\xfnm[#1]{\unskip,\space#1}\fi
\bibitem[{Tang et~al.(2020)Tang, Liu, and Durlofsky}]{tang2020deep}
\bibinfo{author}{M.~Tang}, \bibinfo{author}{Y.~Liu}, \bibinfo{author}{L.~J.
  Durlofsky},
\newblock \bibinfo{title}{A deep-learning-based surrogate model for data
  assimilation in dynamic subsurface flow problems},
\newblock \bibinfo{journal}{Journal of Computational Physics}
  \bibinfo{volume}{413} (\bibinfo{year}{2020}) \bibinfo{pages}{109456}.
\bibitem[{Tang et~al.(2021{\natexlab{a}})Tang, Liu, and
  Durlofsky}]{tang2021deep}
\bibinfo{author}{M.~Tang}, \bibinfo{author}{Y.~Liu}, \bibinfo{author}{L.~J.
  Durlofsky},
\newblock \bibinfo{title}{Deep-learning-based surrogate flow modeling and
  geological parameterization for data assimilation in 3{D} subsurface flow},
\newblock \bibinfo{journal}{Computer Methods in Applied Mechanics and
  Engineering} \bibinfo{volume}{376} (\bibinfo{year}{2021}{\natexlab{a}})
  \bibinfo{pages}{113636}.
\bibitem[{Tang et~al.(2021{\natexlab{b}})Tang, Liu, and
  Durlofsky}]{tang2021history}
\bibinfo{author}{M.~Tang}, \bibinfo{author}{Y.~Liu}, \bibinfo{author}{L.~J.
  Durlofsky},
\newblock \bibinfo{title}{History matching complex {3D} systems using
  deep-learning-based surrogate flow modeling and {CNN-PCA} geological
  parameterization},
\newblock in: \bibinfo{booktitle}{SPE Reservoir Simulation Conference},
  \bibinfo{organization}{OnePetro}, \bibinfo{year}{2021}{\natexlab{b}}.
\bibitem[{Tang et~al.(2021{\natexlab{c}})Tang, Ju, and
  Durlofsky}]{tang2021deep_co2}
\bibinfo{author}{M.~Tang}, \bibinfo{author}{X.~Ju}, \bibinfo{author}{L.~J.
  Durlofsky},
\newblock \bibinfo{title}{Deep-learning-based coupled flow-geomechanics
  surrogate model for {CO$_2$} sequestration},
\newblock \bibinfo{journal}{arXiv preprint arXiv:2105.01334}
  (\bibinfo{year}{2021}{\natexlab{c}}).
\bibitem[{Raissi et~al.(2019)Raissi, Perdikaris, and
  Karniadakis}]{raissi2019physics}
\bibinfo{author}{M.~Raissi}, \bibinfo{author}{P.~Perdikaris},
  \bibinfo{author}{G.~E. Karniadakis},
\newblock \bibinfo{title}{Physics-informed neural networks: {A} deep learning
  framework for solving forward and inverse problems involving nonlinear
  partial differential equations},
\newblock \bibinfo{journal}{Journal of Computational Physics}
  \bibinfo{volume}{378} (\bibinfo{year}{2019}) \bibinfo{pages}{686--707}.
\bibitem[{Wang et~al.(2020)Wang, Zhang, Chang, and Li}]{wang2020deep}
\bibinfo{author}{N.~Wang}, \bibinfo{author}{D.~Zhang},
  \bibinfo{author}{H.~Chang}, \bibinfo{author}{H.~Li},
\newblock \bibinfo{title}{Deep learning of subsurface flow via theory-guided
  neural network},
\newblock \bibinfo{journal}{Journal of Hydrology} \bibinfo{volume}{584}
  (\bibinfo{year}{2020}) \bibinfo{pages}{124700}.
\bibitem[{Wang et~al.(2021{\natexlab{a}})Wang, Chang, and
  Zhang}]{wang2021efficient}
\bibinfo{author}{N.~Wang}, \bibinfo{author}{H.~Chang},
  \bibinfo{author}{D.~Zhang},
\newblock \bibinfo{title}{Efficient uncertainty quantification for dynamic
  subsurface flow with surrogate by theory-guided neural network},
\newblock \bibinfo{journal}{Computer Methods in Applied Mechanics and
  Engineering} \bibinfo{volume}{373} (\bibinfo{year}{2021}{\natexlab{a}})
  \bibinfo{pages}{113492}.
\bibitem[{Wang et~al.(2021{\natexlab{b}})Wang, Chang, and
  Zhang}]{wang2021theory}
\bibinfo{author}{N.~Wang}, \bibinfo{author}{H.~Chang},
  \bibinfo{author}{D.~Zhang},
\newblock \bibinfo{title}{Theory-guided auto-encoder for surrogate construction
  and inverse modeling},
\newblock \bibinfo{journal}{Computer Methods in Applied Mechanics and
  Engineering} \bibinfo{volume}{385} (\bibinfo{year}{2021}{\natexlab{b}})
  \bibinfo{pages}{114037}.
\bibitem[{He et~al.(2020)He, Barajas-Solano, Tartakovsky, and
  Tartakovsky}]{he2020physics}
\bibinfo{author}{Q.~He}, \bibinfo{author}{D.~Barajas-Solano},
  \bibinfo{author}{G.~Tartakovsky}, \bibinfo{author}{A.~M. Tartakovsky},
\newblock \bibinfo{title}{Physics-informed neural networks for multiphysics
  data assimilation with application to subsurface transport},
\newblock \bibinfo{journal}{Advances in Water Resources} \bibinfo{volume}{141}
  (\bibinfo{year}{2020}) \bibinfo{pages}{103610}.
\bibitem[{Tartakovsky et~al.(2020)Tartakovsky, Marrero, Perdikaris,
  Tartakovsky, and Barajas-Solano}]{tartakovsky2020physics}
\bibinfo{author}{A.~M. Tartakovsky}, \bibinfo{author}{C.~O. Marrero},
  \bibinfo{author}{P.~Perdikaris}, \bibinfo{author}{G.~D. Tartakovsky},
  \bibinfo{author}{D.~Barajas-Solano},
\newblock \bibinfo{title}{Physics-informed deep neural networks for learning
  parameters and constitutive relationships in subsurface flow problems},
\newblock \bibinfo{journal}{Water Resources Research} \bibinfo{volume}{56}
  (\bibinfo{year}{2020}) \bibinfo{pages}{e2019WR026731}.
\bibitem[{Tartakovsky et~al.(2021)Tartakovsky, Barajas-Solano, and
  He}]{tartakovsky2021physics}
\bibinfo{author}{A.~M. Tartakovsky}, \bibinfo{author}{D.~A. Barajas-Solano},
  \bibinfo{author}{Q.~He},
\newblock \bibinfo{title}{Physics-informed machine learning with conditional
  {K}arhunen-{L}oeve expansions},
\newblock \bibinfo{journal}{Journal of Computational Physics}
  \bibinfo{volume}{426} (\bibinfo{year}{2021}) \bibinfo{pages}{109904}.
\bibitem[{Zhu and Zabaras(2018)}]{zhu2018bayesian}
\bibinfo{author}{Y.~Zhu}, \bibinfo{author}{N.~Zabaras},
\newblock \bibinfo{title}{Bayesian deep convolutional encoder--decoder networks
  for surrogate modeling and uncertainty quantification},
\newblock \bibinfo{journal}{Journal of Computational Physics}
  \bibinfo{volume}{366} (\bibinfo{year}{2018}) \bibinfo{pages}{415--447}.
\bibitem[{Mo et~al.(2019{\natexlab{a}})Mo, Zhu, Zabaras, Shi, and
  Wu}]{mo2019deep}
\bibinfo{author}{S.~Mo}, \bibinfo{author}{Y.~Zhu},
  \bibinfo{author}{N.~Zabaras}, \bibinfo{author}{X.~Shi},
  \bibinfo{author}{J.~Wu},
\newblock \bibinfo{title}{Deep convolutional encoder-decoder networks for
  uncertainty quantification of dynamic multiphase flow in heterogeneous
  media},
\newblock \bibinfo{journal}{Water Resources Research} \bibinfo{volume}{55}
  (\bibinfo{year}{2019}{\natexlab{a}}) \bibinfo{pages}{703--728}.
\bibitem[{Mo et~al.(2019{\natexlab{b}})Mo, Zabaras, Shi, and
  Wu}]{mo2019deep_auto}
\bibinfo{author}{S.~Mo}, \bibinfo{author}{N.~Zabaras},
  \bibinfo{author}{X.~Shi}, \bibinfo{author}{J.~Wu},
\newblock \bibinfo{title}{Deep autoregressive neural networks for
  high-dimensional inverse problems in groundwater contaminant source
  identification},
\newblock \bibinfo{journal}{Water Resources Research} \bibinfo{volume}{55}
  (\bibinfo{year}{2019}{\natexlab{b}}) \bibinfo{pages}{3856--3881}.
\bibitem[{Mo et~al.(2020)Mo, Zabaras, Shi, and Wu}]{mo2020integration}
\bibinfo{author}{S.~Mo}, \bibinfo{author}{N.~Zabaras},
  \bibinfo{author}{X.~Shi}, \bibinfo{author}{J.~Wu},
\newblock \bibinfo{title}{Integration of adversarial autoencoders with residual
  dense convolutional networks for estimation of non-{G}aussian hydraulic
  conductivities},
\newblock \bibinfo{journal}{Water Resources Research} \bibinfo{volume}{56}
  (\bibinfo{year}{2020}) \bibinfo{pages}{e2019WR026082}.
\bibitem[{Wen et~al.(2021)Wen, Tang, and Benson}]{wen2021towards}
\bibinfo{author}{G.~Wen}, \bibinfo{author}{M.~Tang}, \bibinfo{author}{S.~M.
  Benson},
\newblock \bibinfo{title}{Towards a predictor for {CO$_2$} plume migration
  using deep neural networks},
\newblock \bibinfo{journal}{International Journal of Greenhouse Gas Control}
  \bibinfo{volume}{105} (\bibinfo{year}{2021}) \bibinfo{pages}{103223}.
\bibitem[{Geneva and Zabaras(2020)}]{geneva2020multi}
\bibinfo{author}{N.~Geneva}, \bibinfo{author}{N.~Zabaras},
\newblock \bibinfo{title}{Multi-fidelity generative deep learning turbulent
  flows},
\newblock \bibinfo{journal}{arXiv preprint arXiv:2006.04731}
  (\bibinfo{year}{2020}).
\bibitem[{Meng and Karniadakis(2020)}]{meng2020composite}
\bibinfo{author}{X.~Meng}, \bibinfo{author}{G.~E. Karniadakis},
\newblock \bibinfo{title}{A composite neural network that learns from
  multi-fidelity data: Application to function approximation and inverse {PDE}
  problems},
\newblock \bibinfo{journal}{Journal of Computational Physics}
  \bibinfo{volume}{401} (\bibinfo{year}{2020}) \bibinfo{pages}{109020}.
\bibitem[{De et~al.(2020)De, Britton, Reynolds, Skinner, Jansen, and
  Doostan}]{de2020transfer}
\bibinfo{author}{S.~De}, \bibinfo{author}{J.~Britton},
  \bibinfo{author}{M.~Reynolds}, \bibinfo{author}{R.~Skinner},
  \bibinfo{author}{K.~Jansen}, \bibinfo{author}{A.~Doostan},
\newblock \bibinfo{title}{On transfer learning of neural networks using
  bi-fidelity data for uncertainty propagation},
\newblock \bibinfo{journal}{International Journal for Uncertainty
  Quantification} \bibinfo{volume}{10} (\bibinfo{year}{2020})
  \bibinfo{pages}{543--573}.
\bibitem[{Song and Tartakovsky(2022)}]{song2021transfer}
\bibinfo{author}{D.~H. Song}, \bibinfo{author}{D.~M. Tartakovsky},
\newblock \bibinfo{title}{Transfer learning on multi-fidelity data},
\newblock \bibinfo{journal}{Journal of Machine Learning for Modeling and
  Computing} \bibinfo{volume}{2} (\bibinfo{year}{2022})
  \bibinfo{pages}{31--47}.
\bibitem[{Zhou(2012)}]{zhou2012parallel}
\bibinfo{author}{Y.~Zhou}, \bibinfo{title}{Parallel {G}eneral-purpose
  {R}eservoir {S}imulation with {C}oupled {R}eservoir {M}odels and
  {M}ultisegment {W}ells}, Ph.D. thesis, Stanford University,
  \bibinfo{year}{2012}.
\bibitem[{Peaceman(1983)}]{peaceman1983interpretation}
\bibinfo{author}{D.~W. Peaceman},
\newblock \bibinfo{title}{Interpretation of well-block pressures in numerical
  reservoir simulation with nonsquare grid blocks and anisotropic
  permeability},
\newblock \bibinfo{journal}{SPE Journal} \bibinfo{volume}{23}
  (\bibinfo{year}{1983}) \bibinfo{pages}{531--543}.
\bibitem[{Zhang et~al.(2008)Zhang, Pickup, and Christie}]{zhang2008new}
\bibinfo{author}{P.~Zhang}, \bibinfo{author}{G.~E. Pickup},
  \bibinfo{author}{M.~A. Christie},
\newblock \bibinfo{title}{A new practical method for upscaling in highly
  heterogeneous reservoir models},
\newblock \bibinfo{journal}{SPE Journal} \bibinfo{volume}{13}
  (\bibinfo{year}{2008}) \bibinfo{pages}{68--76}.
\bibitem[{Chen et~al.(2008)Chen, Mallison, and Durlofsky}]{chen2008nonlinear}
\bibinfo{author}{Y.~Chen}, \bibinfo{author}{B.~T. Mallison},
  \bibinfo{author}{L.~J. Durlofsky},
\newblock \bibinfo{title}{Nonlinear two-point flux approximation for modeling
  full-tensor effects in subsurface flow simulations},
\newblock \bibinfo{journal}{Computational Geosciences} \bibinfo{volume}{12}
  (\bibinfo{year}{2008}) \bibinfo{pages}{317--335}.
\bibitem[{Crain(2020)}]{crain2020multifidelity}
\bibinfo{author}{D.~Crain},
\newblock \bibinfo{title}{Extended {F}ramework for {M}ultifidelity
  {U}ncertainty {Q}uantification in {S}ubsurface {F}low {S}ystems},
\newblock \bibinfo{organization}{Master's thesis, Stanford University},
  \bibinfo{year}{2020}.
\bibitem[{Jiang and Durlofsky(2021)}]{jiang2021treatment}
\bibinfo{author}{S.~Jiang}, \bibinfo{author}{L.~J. Durlofsky},
\newblock \bibinfo{title}{Treatment of model error in subsurface flow history
  matching using a data-space method},
\newblock \bibinfo{journal}{Journal of Hydrology} \bibinfo{volume}{603}
  (\bibinfo{year}{2021}) \bibinfo{pages}{127063}.
\bibitem[{Liu and Durlofsky(2021)}]{liu20213d}
\bibinfo{author}{Y.~Liu}, \bibinfo{author}{L.~J. Durlofsky},
\newblock \bibinfo{title}{{3D CNN-PCA}: A deep-learning-based parameterization
  for complex geomodels},
\newblock \bibinfo{journal}{Computers \& Geosciences} \bibinfo{volume}{148}
  (\bibinfo{year}{2021}) \bibinfo{pages}{104676}.
\bibitem[{Kingma and Ba(2014)}]{kingma2014adam}
\bibinfo{author}{D.~P. Kingma}, \bibinfo{author}{J.~Ba},
\newblock \bibinfo{title}{Adam: A method for stochastic optimization},
\newblock \bibinfo{journal}{arXiv preprint arXiv:1412.6980}
  (\bibinfo{year}{2014}).
\bibitem[{Shirangi and Durlofsky(2016)}]{shirangi2016general}
\bibinfo{author}{M.~G. Shirangi}, \bibinfo{author}{L.~J. Durlofsky},
\newblock \bibinfo{title}{A general method to select representative models for
  decision making and optimization under uncertainty},
\newblock \bibinfo{journal}{Computers \& Geosciences} \bibinfo{volume}{96}
  (\bibinfo{year}{2016}) \bibinfo{pages}{109--123}.
\bibitem[{Emerick and Reynolds(2013{\natexlab{a}})}]{emerick2013ensemble}
\bibinfo{author}{A.~A. Emerick}, \bibinfo{author}{A.~C. Reynolds},
\newblock \bibinfo{title}{Ensemble smoother with multiple data assimilation},
\newblock \bibinfo{journal}{Computers \& Geosciences} \bibinfo{volume}{55}
  (\bibinfo{year}{2013}{\natexlab{a}}) \bibinfo{pages}{3--15}.
\bibitem[{Emerick and Reynolds(2013{\natexlab{b}})}]{emerick2013investigation}
\bibinfo{author}{A.~A. Emerick}, \bibinfo{author}{A.~C. Reynolds},
\newblock \bibinfo{title}{Investigation of the sampling performance of
  ensemble-based methods with a simple reservoir model},
\newblock \bibinfo{journal}{Computational Geosciences} \bibinfo{volume}{17}
  (\bibinfo{year}{2013}{\natexlab{b}}) \bibinfo{pages}{325--350}.

\end{thebibliography}

\end{document}